\documentclass[10pt,journal,compsoc]{IEEEtran}

\newif\iftr
\newif\ifcnf

\trfalse
\cnftrue
 
\trtrue
\cnffalse

\newif\ifnohl
\nohlfalse

\newif\ifnonb
\nonbtrue

\newif\ifbiblat
\biblatfalse

\newif\ifsq
\newif\ifsqCAP
\newif\ifsqVS
\newif\ifsqEN
\newif\ifsqTIT

\usepackage{xcolor}
\definecolor{darkgrey}{RGB}{70,70,70}
\definecolor{darkgreen}{RGB}{0,130,0}
\definecolor{lightgrey}{RGB}{200,200,200}
\definecolor{lyellow}{RGB}{255,255,100}
\definecolor{llyellow}{RGB}{250,250,180}
\definecolor{lgreen}{RGB}{144,238,144}
\definecolor{raphael_comments}{RGB}{13, 145, 24}

\usepackage[customcolors]{hf-tikz}
\hfsetbordercolor{white}
\hfsetfillcolor{vlgray}

\definecolor{vlgray}{rgb}{0.77 0.77 0.77}
\definecolor{ablack}{rgb}{0.2 0.2 0.2}
\definecolor{vllgray}{rgb}{0.9 0.9 0.9}
\definecolor{bblue}{rgb}{0.7 0.7 0.99}

\usepackage{colortbl}

\newcommand{\trtxt}[1]{\textcolor{darkgreen}{#1}}

\newcommand{\vspaceSQ}[1]{\ifsqVS\vspace{#1}\fi}
\newcommand{\enlargeSQ}[1]{\ifsqEN\enlargethispage{\baselineskip}\fi}

\newcommand{\ignore}[1]{}

\sqtrue
\sqCAPfalse
\sqENtrue
\sqVStrue
\sqTITtrue

\sqfalse
\sqCAPfalse
\sqENfalse
\sqVSfalse
\sqTITfalse

\usepackage{etex}
\usepackage{balance}
\usepackage{epstopdf}
\usepackage{placeins}

\usepackage{graphicx}
\usepackage{float}
\usepackage{dblfloatfix}
\usepackage{multirow}
\usepackage{rotating}
\usepackage{makecell}
\usepackage{tabulary}
\usepackage{parcolumns}
\usepackage{tikz}
\usetikzlibrary{tikzmark}

\usepackage{xpatch}
\expandafter\xpatchcmd
\csname pgfk@/tikz/every picture/.@cmd\endcsname
{\thepage}{\arabic{page}}{}{}

\tikzstyle{comment} = [draw, fill=blue!70, text=white, text width=3cm, minimum height=1cm, rounded corners, align=left, font=\scriptsize]
\tikzstyle{background_alg} = [draw, fill=blue!20, opacity=0.4, inner sep=4pt, rounded corners=2pt]

\usetikzlibrary{shapes}
\usetikzlibrary{plotmarks}
\usetikzlibrary{calc, fit}

\usepackage{enumitem}

\usepackage{amsthm}
\usepackage{amsmath,amssymb,amsfonts}
\usepackage{mathtools,mathrsfs}

\newtheorem{theorem}{Theorem}[section]

\newtheorem{proposition}[theorem]{Proposition}

\newtheorem{example}[theorem]{Example}
\newtheorem{definition}[theorem]{Definition}

\usepackage{soul}
\usepackage{fontawesome}
\usepackage{pifont}
\usepackage{textcomp}
\usepackage{booktabs}
\usepackage{url}
\usepackage{pbox}
\usepackage[normalem]{ulem}
\usepackage[10pt]{moresize}

\ifsqCAP
\usepackage[font={normalfont, scriptsize}]{caption}
\usepackage[font={normalfont, scriptsize}]{subcaption}
\else
\usepackage[font={normalfont, scriptsize}]{caption}
\usepackage[font={normalfont, scriptsize}]{subcaption}
\fi

\ifsqTIT
\usepackage[compact]{titlesec}
\titlespacing*{\section}{0pt}{4pt}{3pt}
\titlespacing*{\subsection}{0pt}{3pt}{2pt}
\titlespacing*{\subsubsection}{0pt}{2pt}{1pt}
\fi

\usepackage{inconsolata}
\usepackage{listings}

\ifsq
\else
\fi

\newcommand{\maciej}[1]{\textcolor{blue}{[Maciej: #1]}}

\newcommand{\lukas}[1]{\textcolor{orange}{[Lukas: #1]}}

\newcommand{\florian}[1]{\textcolor{purple}{[Florian: #1]}}

\definecolor{hlL}{rgb}{0.8 0.8 0.99}

\newcounter{highlight}

\newcounter{hlLR}

\newcounter{hlLIR}

\newcounter{hlLIIR}

\newcounter{Ahighlight}

\usepackage{scalerel,stackengine}
\stackMath
\newcommand\rwh[1]{%
\savestack{\tmpbox}{\stretchto{%
  \scaleto{%
        \scalerel*[\widthof{\ensuremath{#1}}]{\kern-.6pt\bigwedge\kern-.6pt}%
                  {\rule[-\textheight/2]{1ex}{\textheight}}
                              }{\textheight}%
}{0.5ex}}%
\stackon[1pt]{#1}{\tmpbox}%
}

\usepackage[hang,flushmargin]{footmisc}

\DeclarePairedDelimiter\abs{\lvert}{\rvert}
\DeclarePairedDelimiter\norm{\lVert}{\rVert}

\renewcommand{\epsilon}{\ensuremath\varepsilon}

\renewcommand{\phi}{\ensuremath{\varphi}}

\newcommand{\fRB}[1]{\left(#1\right)}

\usepackage{physics}

\newcolumntype{y}{>{}l}

\usepackage{hhline}

\newif\ifuncompressed
\uncompressedtrue

\usepackage{bbm}
\usepackage{comment}
\usepackage[hidelinks]{hyperref}
\usepackage{mathtools} 
\usepackage[arrow, matrix]{xy}
\usepackage{siunitx}
\usepackage{tikz-cd}
\usepackage{xfrac}
\usepackage{pdfpages}
\usepackage{url}
\usepackage{afterpage}
\usepackage{verbatim}
\usepackage{bbm}
\usepackage{braket}
\usepackage{comment}
\usepackage{graphicx}
\usepackage{amsfonts}
\usepackage{textcomp}
\usepackage[acronym]{glossaries}

\ifcsname comment\endcsname%
    \renewcommand{\comment}[1]{\ignorespaces}
\else%
    \newcommand{\comment}[1]{\ignorespaces}
\fi

\newif\ifFINAL
\FINALtrue

\ifFINAL
\renewcommand{\maciej}[1]{}
\renewcommand{\florian}[1]{}
\fi

\IEEEaftertitletext{\vspace{-2\baselineskip}}

\makeglossaries

\newacronym[\glslongpluralkey={message passing neural networks}]{mpnn}{MPNN}{message-passing neural network}
\newacronym{mp}{MP}{message passing}
\newacronym{gnn}{GNN}{graph neural network}
\newacronym{gcn}{GCN}{graph convolutional network}
\newacronym{gat}{GAT}{graph attention network}
\newacronym{gin}{GIN}{graph isomorphism network}
\newacronym{bamp}{BAMP}{boundary-adjacency message-passing}
\newacronym{grl}{GRL}{graph representation learning}
\newacronym{gdm}{GDM}{graph data model}
\newacronym{hogdm}{HoGDM}{higher-order graph data model}
\newacronym{hog}{HoG}{higher-order graph}
\newacronym{subgrl}{SubGRL}{subgraph representations
learning}
\newacronym[\glslongpluralkey={higher-order GRLs}]{hogrl}{HoGRL}{higher-order GRL}
\newacronym{liwip}{LiWiP}{Lift-Wire-Process}
\newacronym{wltest}{WL-test}{Weisfeiler-Lehman test}
\newacronym{cin}{CIN}{cell isomorphism networks}
\newacronym{wl}{WL}{Weisfeiler-Lehman}
\newacronym{fwl}{FWL}{folklore-Weisfeiler-Lehman}
\newacronym{imp}{IMP}{incidence message passing}
\newacronym{damp}{DAMP}{down-adjacency message passing}
\newacronym{gi}{GI}{graph isomorphism }
\newacronym{ign}{IGN}{invariant graph network} 
\newacronym{gsn}{GSN}{graph substructure network} 
\newacronym{mpsn}{MPSN}{message passing simplicial network} 
\newacronym{ngnn}{NGNN}{nested graph neural network}
\newacronym{cwn}{CWN}{CW-network} 
\newacronym{lstm}{LSTM}{long short term memory network} 
\newacronym{han}{HAN}{heterogeneous graph attention network}
\newacronym{hgt}{HGT}{heterogeneous graph transformer}
\newacronym{het}{HET}{heterogeneous convolution} 
\newacronym{sin}{SIN}{simplicial isomorphism network} 
\newacronym{ml}{ML}{machine learning}
\newacronym{mlp}{MLP}{multi-layer perceptron}

\renewcommand{\marginpar}[1]{}

\let\oldcolorbox\colorbox
\renewcommand{\colorbox}[2]{\oldcolorbox{white}{#2}}

\begin{document}

\title{Demystifying Higher-Order Graph \\ Neural Networks}

\author{Maciej Besta$^{1\dagger}$\thanks{$^\dagger$Corresponding author}, Florian Scheidl$^1$, Lukas Gianinazzi$^1$, Grzegorz Kwasniewski$^1$,\\ Shachar Klaiman$^2$, Jürgen Müller$^2$, Torsten Hoefler$^1$\\
\vspace{0.5em}{\small $^1$ETH Zurich \quad $^2$BASF SE}}

\IEEEtitleabstractindextext{%
\begin{abstract}
Higher-order graph neural networks (HOGNNs) and the related architectures from Topological Deep Learning are an important class of GNN
models that harness polyadic relations between vertices beyond plain
edges. They have been used to eliminate issues such as
over-smoothing or over-squashing, to significantly enhance the accuracy of
GNN predictions, to improve the expressiveness of GNN architectures, and for numerous other goals.
A plethora of HOGNN models have been introduced, and they come with diverse neural architectures, and
even with different notions of what the ``higher-order'' means.
This richness makes it very challenging to appropriately analyze and compare HOGNN models,
and to decide in what scenario to use specific ones.
To alleviate this, we first design an in-depth taxonomy and a blueprint for HOGNNs.
This facilitates designing models that maximize performance.
Then, we use our taxonomy to analyze and compare the available HOGNN models.
The outcomes of our analysis are synthesized in a set of insights that
help to select the most beneficial GNN model in a given scenario, and a
comprehensive list of challenges and opportunities for further research
into more powerful HOGNNs.
\vspace{-0.5em}
\end{abstract}


}

\maketitle

\IEEEdisplaynontitleabstractindextext
\IEEEpeerreviewmaketitle

\newcommand{\dimp}{\Leftrightarrow}
\newcommand{\rimp}{\Rightarrow}
\newcommand{\limp}{\Leftarrow}
\newcommand{\defi}{$:\Leftrightarrow$}
\newcommand{\nec}{"$\limp$"}
\newcommand{\suf}{"$\rimp$"}
\newcommand{\xor}{\oplus}
\newcommand{\dset}[2]{\left\{  \, #1 \, \middle| \, #2  \right\}	}
\newcommand{\tuple}[1]{\left(#1\right)}
\newcommand{\listing}[3]{ {#1}_{#2}, \dots , {#1}_{#3} }
\newcommand{\conj}[1]{\overline{#1}}
\newcommand{\cc}[1]{\overline{#1}}
\newcommand{\pot}[1]{2^{{#1}}}
\newcommand{\union}[2]{\bigcup\limits_{#1}{#2}}
\newcommand{\inter}[2]{\bigcap\limits_{#1}{#2}}
\newcommand{\sub}{\subseteq}
\newcommand{\nsub}{\not \subseteq}
\newcommand{\super}{\supseteq}
\newcommand{\bs}{\ensuremath{\backslash}}
\newcommand{\sm}{\ensuremath{\setminus}}
\newcommand{\du}{\sqcup} 
\newcommand{\Du}{\bigsqcup} 
\newcommand{\Card}[1]{Card({#1})}
\newcommand{\smzero}{\ensuremath{\setminus}\{0\}}
\newcommand{\impcom}[1]{\ensuremath{\stackrel{\text{#1}}{\implies}}}
\newcommand{\eqcom}[1]{\ensuremath{\stackrel{\text{#1}}{=}}}
\newcommand{\subcom}[1]{\ensuremath{\stackrel{\text{#1}}{\subseteq}}}
\newcommand{\supcom}[1]{\ensuremath{\stackrel{\text{#1}}{\supseteq}}}
\newcommand{\com}[2]{\stackrel{{#1}}{#2}}
\newcommand{\nth}{$n^{th}$}
\newcommand{\thh}{$^{th}$}
\newcommand{\linc}{$"\subseteq":$}
\newcommand{\rinc}{$"\supseteq":$}
\newcommand{\foa}[3]{\ensuremath{\forall {#1}\leq {#2} \leq {#3}}}
\newcommand{\mat}[3]{M_{{#1}\times{#2}}({#3})}
\newcommand{\lcup}[2]{\bigcup\limits_{#1}^{#2}}
\newcommand{\ldiscup}[2]{\bigsqcup\limits_{#1}^{#2}}
\newcommand{\lcap}[2]{\bigcap\limits_{#1}^{#2}}
\newcommand{\calM}{\mathcal{M}}
\newcommand{\calN}{\mathcal{N}}
\newcommand{\modtime}{*}
\newcommand{\calL}{\mathcal{L}}
\newcommand{\lsum}[2]{\sum\limits_{#1}^{#2}}
\newcommand{\p}[2]{\prod\limits_{#1}{#2} }
\newcommand{\lprod}[2]{\prod\limits_{#1}^{#2}}
\newcommand{\litimes}[2]{\bigtimes\limits_{#1}^{#2}}%
\newcommand{\modulo}[2]{{\raisebox{.2em}{$#1$}\left/\raisebox{-.2em}{$#2$}\right.}}
\newcommand{\modulol}[2]{{\raisebox{-.2em}{$#2$}\left \ensuremath{\backslash} \raisebox{.2em}{$#1$}\right}}
\newcommand{\gspan}[1]{\langle {#1}\rangle}
\newcommand{\ngspan}[1]{\langle \langle{#1}\rangle \rangle}
\newcommand{\sign}{sign}
\newcommand{\zmzm}[1]{(\modulo{\ZE}{{#1}\ZE})}
\newcommand{\zmzmu}[1]{(\modulo{\ZE}{#1}\ZE)^\times}
\newcommand{\units}{^\times}
\newcommand{\inner}[1]{{#1}^{\circ}}
\newcommand{\closed}[1]{\; closed \;in \;{#1}}
\newcommand{\Tfour}{T4}
\newcommand{\normal}{normal}
\newcommand{\cpone}[2]{\prod\limits_{#1}({#2})}
\newcommand{\inj}{\xhookrightarrow}
\newcommand{\isoS}[2]{\underset{({#1})}{\overset{{#2}}{\cong}}}
\newcommand{\iso}[1]{\xrightarrow[({#1})]{\cong}}
\newcommand{\inc}[1]{\underset{({#1})}{\leqslant}}
\newcommand{\inv}{^{-1}}
\renewcommand{\gg}[2]{\interval{#1}{#2}}
\newcommand{\oo}[2]{\interval[open]{#1}{#2}}
\newcommand{\og}[2]{\interval[open left]{#1}{#2}}
\newcommand{\go}[2]{\interval[open right]{#1}{#2}}
\newcommand{\limes}[3]{\lim\limits_{#1 \rightarrow #2}{#3}}
\newcommand{\rlimes}[3]{\lim\limits_{#1 \searrow #2}{#3}}
\newcommand{\llimes}[3]{\lim\limits_{#1\nearrow #2}{#3}}
\newcommand{\limsu}[3]{\limsup\limits_{#1\rightarrow #2} #3}
\newcommand{\limin}[3]{\liminf\limits_{ #1 \rightarrow #2 } #3 }
\newcommand{\func}[3]{ #1 : #2 \to #3}
\newcommand{\su}[3]{\sum_{#1}^{#2}{#3} }
\newcommand{\pro}[3]{\prod_{#1}^{#2}{\left(#3\right)} }
\newcommand{\floor}[1]{\lfloor #1 \rfloor}
\renewcommand{\norm}[1]{\left\lVert#1\right\rVert}
\newcommand{\inp}[1]{\left\langle#1\right\rangle }
\newcommand{\iprod}[2]{\inp{#1,#2}}
\renewcommand{\abs}[1]{\left\lvert {#1} \right\rvert}
\newcommand{\BF}{BF}
\newcommand{\SF}{SF}
\newcommand{\pmap}{\pi}
\newcommand{\dint}[4]{\int_{#1}^{#2}\! #3 \mathrm{d}{#4}}
\newcommand{\uint}[2]{\int {#1} \, \mathrm{d}{#2}}
\newcommand{\der}[2]{\frac{\dif {#1}}{ \dif {#2} }}
\newcommand{\pnder}[3]{ \frac{\partial^{#1} {#2} }{ \partial {#3}^{#1} }}
\newcommand{\pder}[2]{ \frac{\partial {#1} }{ \partial {#2}}}
\newcommand{\nder}[3]{ \frac{\dif^{#1} {#2} }{ \dif {#3}^{#1} } }
\newcommand{\matn}[2]{M_{#1 \times #2}}
\newcommand{\M}{M_{m\times n}(\KE)}
\newcommand{\Mn}{M_{n\times n}(\KE)}
\newcommand{\koord}[2]{\left[#1\right]_{#2}}
\newcommand{\cvec}[2]{ \left[{#1}\right]_\basis{#2} }
\newcommand{\cvech}[2]{ \left[{#1}\right]_\basis{#2}^\ast }
\newcommand{\abb}[3]{\left[{#1}\right]_{#2}^{#3}}		
\newcommand{\bfabb}[2]{\psi_{#1}(#2)}  
\newcommand{\tmat}[3]{ \left[ {#1} \right]_\basis{#2}^\basis{#3} }
\newcommand{\bmat}[2]{  \psi_{\basis{#2}}(#1)  }
\newcommand{\basis}[1]{ \mathcal{#1} }
\newcommand{\matr}[1]{
    \begin{pmatrix}
        #1
    \end{pmatrix}
}
\newcommand{\id}{\mathbbm{1}} 
\newcommand{\spur}{spur}
\newcommand{\signature}{\text{sign}}
\newcommand{\Gl}{G\ell}
\newcommand{\arr}[1]{
    \begin{matrix*}
        #1
    \end{matrix*}
}
\newcommand{\augmatr}[2]{ \left( \arr{#1} \; \middle| \; \arr{#2} \right) }
\newcommand{\row}[2]{#1_{(#2)}}
\newcommand{\col}[2]{#1^{(#2)}}
\newcommand{\spa}[1]{span\, #1}
\newcommand{\qs}[2]{#1/#2}
\newcommand{\indS}[1]{[{#1}]}
\newcommand{\nat}{\mathbb{Z}_{>0}}
\newcommand{\intg}[1]{\mathbb{Z}_{\geq{#1}}}
\newcommand{\RE}{\mathbb{R}}
\newcommand{\RI}{\mathbb{R}_\infty}
\newcommand{\QE}{\mathbb{Q}}
\newcommand{\CE}{\mathbb{C}}
\newcommand{\NE}{\mathbb{N}}
\newcommand{\NO}{\mathbb{N}_0}
\newcommand{\KE}{\mathbb{K}}
\newcommand{\ZE}{\mathbb{Z}}
\newcommand{\PE}{\mathbb{P}}
\newcommand{\EE}{\mathbb{E}}
\newcommand{\TE}{\mathbb{T}}
\newcommand{\SE}{\mathbb{S}}
\newcommand{\Rgeq}{\RE_{\geq 0}}
\newcommand{\Rleq}{\RE_{\leq 0}}
\newcommand{\Rg}{\RE_{\geq 0}}
\newcommand{\Rl}{\RE_{\leq 0}}
\newcommand{\Zgeq}{\ZE_{\geq 0}}
\newcommand{\Zleq}{\ZE_{\leq 0}}
\newcommand{\Zg}{\ZE_{\geq 0}}
\newcommand{\Zl}{\ZE_{\leq 0}}
\newcommand{\PC}{\mathcal{P}}
\newcommand{\BC}{\mathcal{B}}
\newcommand{\EC}{\mathcal{E}}
\newcommand{\QC}{\mathcal{Q}}
\newcommand{\RC}{\mathcal{R}}
\newcommand{\LC}{\mathcal{L}}
\newcommand{\FC}{\mathcal{F}}
\newcommand{\TC}{\mathcal{T}}
\newcommand{\SC}{\mathcal{S}}
\newcommand{\WC}{\mathcal{W}}
\newcommand{\yF}{\mathbf{y}}
\newcommand{\rF}{\mathbf{r}}
\newcommand{\vF}{\mathbf{v}}
\newcommand{\FF}{\mathbf{F}}
\newcommand{\sF}{\mathbf{s}}
\newcommand{\zF}{\mathbf{z}}
\newcommand{\gF}{\mathbf{g}}
\newcommand{\iF}{\mathbf{i}}
\newcommand{\ZK}{\mathfrak{Z}}
\newcommand{\REK}{\mathfrak{RE}}
\newcommand{\IMK}{IM}
\newcommand{\falls}{\text{ falls }}
\newcommand{\ci}{i} 
\newcommand{\zsh}[1]{Z({#1})}
\newcommand{\pathc}[1]{P({#1})}
\newcommand{\obda}{$w.l.o.g.$}


\newcommand{\lDu}[2]{\biguplus\limits_{#1}^{#2}} 

\newcommand{\RP}[1]{\mathbb{R}\mathbb{P}^{#1}}
\newcommand{\ball}[2]{B_{#1}(#2)}
\newcommand{\ballSub}[3]{B_{#1}^{#3}(#2)}
\newcommand{\REb}[1]{\mathbb{R}_{>{#1}}}

\newcommand{\homop}[3]{{#1}\stackrel{{#3}}{\simeq}{#2}} 
\newcommand{\homo}[3]{{#1}\stackrel{{#3}}{\approx}{#2}} 
\newcommand{\homoe}[3]{{#1}\stackrel{{#3}}{\approxeq}{#2}}

\newcommand{\fundtimes}{\bullet}
\newcommand{\concat}{\ast}
\newcommand{\const}[1]{[c_{#1}]}
\newcommand{\constp}[1]{c_{#1}}
\newcommand{\fundp}[1]{\pi_1({#1})}
\newcommand{\cont}[2]{C({#1},{#2})}
\newcommand{\maps}[2]{F({#1},{#2})}

\newcommand{\nisoS}[2]{\underset{({#1})}{\overset{{#2}}{\ncong}}}
\newcommand{\niso}[1]{\xrightarrow[({#1})]{\ncong}}

\newcommand{\indgrcomp}[1]{\cdot_{#1}}
\newcommand{\fptimes}{\star}
\newcommand{\fp}[1]{\circledast_{#1}}
\newcommand{\Lebn}{\eta}

\newcommand{\Hom}[3]{Hom_{(#1)}(#2,#3)}
\newcommand{\lplus}[2]{\bigoplus\limits_{#1}^{#2}}
\newcommand{\ltens}[2]{\bigotimes\limits_{#1}^{#2}}
\newcommand{\Tens}[3]{#2\otimes_{#1}{#3}}
\newcommand{\tens}[3]{#2\otimes_{#1}{#3}}
\newcommand{\Ten}[1]{\otimes_{#1}}
\newcommand{\ten}[1]{\otimes_{#1}}

\newcommand{\otspam}{\mapsfrom}
\newcommand{\local}[2]{#1[\frac{1}{#2}]}
\newcommand{\tocom}[1]{\stackrel{{#1}}{\longrightarrow}}

\newcommand{\spans}[1]{\langle #1 \rangle}
\newcommand{\scalarp}[2]{\langle #1,#2  \rangle}

\newcommand{\germs}[2]{\mathcal{F}_{#1}{#2}}
\newcommand{\tang}[2]{\mathcal{T}_{#1}{#2}}

\newcommand{\surj}{\twoheadrightarrow}
\newcommand{\ldirsum}[2]{\bigoplus\limits_{#1}^{#2}}

\renewcommand{\tangent}[2]{\frac{\partial}{\partial {#1}}|_{#2}}
\newcommand{\tangentVF}[1]{\frac{\partial}{\partial {#1}}}

\newcommand{\Rnneg}{\mathbb{R}_{\geqslant 0}}
\newcommand{\Rpos}{\mathbb{R}_{>0}}
\newcommand{\distprod}{\star}
\newcommand{\entr}[2]{H_{#1}^{#2}}
\newcommand{\sphere}[2]{\mathbb{S}^{#1}_{#2}}

\newcommand{\To}{\mathbb{T}}

\newcommand{\lp}[2]{L_\mu^{#1}(#2)}
\newcommand{\bowdin}[3]{D_{#2}^{#3}(#1)}
\newcommand{\ulocen}[2]{\overline{h_\mu}(#1,#2)}
\newcommand{\llocen}[2]{\underline{h_\mu}(#1,#2)}
\newcommand{\pdynen}[2]{h_\mu(#1,#2)}
\newcommand{\dynen}[1]{h_\mu(#1)}
\newcommand{\converge}{\com{n\to \infty}{\longrightarrow}}
\newcommand{\hb}{\rightarrowtail}

\newcommand{\renyi}[2]{H_{#1,#2}}

\newenvironment{comp_item}{\begin{itemize}[topsep=0pt,partopsep=0pt, label=$\bullet$]
                               \setlength{\itemsep}{0pt}
                               \setlength{\parskip}{0pt}
                               \setlength{\parsep}{0pt}
                               }{\end{itemize}}
\newenvironment{comp_enum}{\begin{enumerate}[topsep=0pt,partopsep=0pt]
                               \setlength{\itemsep}{0pt}
                               \setlength{\parskip}{0pt}
                               \setlength{\parsep}{0pt}
                               }{\end{enumerate}}

\newenvironment{circ_item}[1][\faCircleO]{\begin{itemize}[leftmargin=*,label=#1, topsep=0pt,partopsep=0pt]
                                              \setlength{\itemsep}{0.25pt}
                                              \setlength{\parskip}{0.25pt}
                                              \setlength{\parsep}{0.25pt}

                                              }{\end{itemize}}

\newcommand{\transpose}{^{\mathsf{T}}}
\newcommand{\adjoint}{^{\ast}}
\newcommand{\multiset}[1]{\{\{#1\}\}}

\newcommand{\incidence}{\mathcal{I}}
\newcommand{\coincidence}{\mathcal{CI}}
\newcommand{\metahetgraph}{multyhop network}
\newcommand{\Metahetgraph}{Multyhop network}
\newcommand{\metaterm}[1][v]{\mathfrak{M}_t(#1)}
\newcommand{\simplex}{node-set}
\newcommand{\simplices}{node-sets}

\iftr
\else
\fi

\vspaceSQ{-1em}
\section{Introduction}
\label{sec:intro}
\vspaceSQ{-0.5em}

Graph neural networks (GNNs)~\cite{wu2020comprehensive, zhou2020graph, zhang_deep_2022,
  chami2020machine, hamiltonGraphRepresentationLearning, jin2019hypergraph, bronstein2017geometric} are a powerful class of deep learning (DL) models
for classification and regression over interconnected graph datasets. They have
  been used to study human interactions, analyze protein structures, design
  chemical compounds, discover drugs, identify intruder machines, model
relationships between words, find efficient transportation routes, 
and many others~\cite{achantaSLICSuperpixels2010, schaubRandomWalksSimplicial2020, zhaoAttentionBasedGraphNeural2020, yuSpatioTemporalGraphConvolutional2018,
yaoDeepMultiViewSpatialTemporal2018, kimHypergraphAttentionNetworks2020, wangKnowledgeGraphEmbedding2020, yingGraphConvolutionalNeural2018, roddenberryHodgeNetGraphNeural2019, mengSubgraphPatternNeural, tekin_measuring_2017, zhang2019hyper, bordesQuestionAnsweringSubgraph2014, gianinazzi2021learning}.

Many successful GNN models have been proposed, for example Graph Convolution
Network (GCN)~\cite{kipf_semi-supervised_2017}, Graph Attention Network
(GAT)~\cite{velickovic_graph_2018}, Graph Isomorphism Network
(GIN)~\cite{xu2018powerful}, or Message-Passing (MP) Neural
Networks~\cite{gilmer2017neural}.
These GNN models are designed for ``plain graphs'' where relations are only
\emph{dyadic}, i.e., only defined for vertex
\emph{pairs}~\cite{hamilton_representation_2018, wu2020comprehensive,
zhang_deep_2022}. While being simple, such pairwise relations can be
insufficient to adequately capture relationships encoded in
data~\cite{battiston2020networks, battiston_physics_2021}. 
For example, consider the following two cases of co-authorship relations: (a)
three authors work together (as a group of three) on a single paper, and (b)
every two of these authors work as a pair on separate
papers~\cite{yang2022efficient}. These cases cannot be distinguished when using
a plain graph, because they are both modeled as a clique over three vertices.
%
%
%
Other such examples can be found in social networks (e.g., a group of friends
forms a \emph{polyadic} relation~\cite{alvarez-rodriguez_evolutionary_2021}),
in pharmaceutical interaction networks (e.g., multi-drug
interactions~\cite{tekin_measuring_2017}), in
neuroscience~\cite{giusti_twos_2016} or
ecology~\cite{grilli_higher-order_2017}.
To capture such relationships, going beyond pairwise interactions is necessary.

\emph{Higher-order graph data models (HOGDMs)} address this by explicitly
encoding polyadic interactions into the graph data model (GDM). HOGDMs have
been intensely studied, and many classes of such data models were proposed,
including hypergraphs (HGs), simplicial complexes (SCs), or cell complexes
(CCs)~\cite{bick_what_2022}.
Simultaneously, in recent years, there has been an increasing interest in
\emph{higher-order graph representation learning (HOGRL)}, an important part of graph representation learning (GRL), with many
\emph{higher-order graph neural network (HOGNN)} models being proposed. These
models have gained wide recognition as they have been proven to be
fundamentally more powerful than GNN models defined on plain
graphs~\cite{xu2018powerful}, for example in terms of what graphs they can distinguish between.

Many HOGNNs have been introduced, but the term ``higher-order'' has been used
in so many different settings, that it is not clear how to reason about, let
alone compare, different HOGNNs.
A potential HOGNN model can be based on any of the available HOGDMs (HGs, SCs,
CCs, etc.) and it can harness any of the available GNN mechanisms (convolution,
attention, MP, etc.). Moreover, many works introduce HO into plain graphs,
by considering convolution over ``higher-order neighboring vertices'' or graph
motifs. Some models even use hierarchical nested graph data models and use convolution,
attention, or MP, on such nested graphs.
All these aspects results in a {very} large space of potential HOGNNs,
hindering the understanding of fundamental principles behind these models, differences
between models, and their pros and cons.

To address these issues, we analyze different aspects of HOGNNs
and derive a taxonomy, in which we formalize and define new classes of
graph data models, relations between them, and the corresponding classes of HOGNNs (\textbf{contribution~\#1}). 
The taxonomy comes with an accompanying blueprint recipe for generating new HOGNNs (\textbf{contribution~\#2}).
We use our taxonomy to study over 100 HOGNN related schemes (\textbf{contribution~\#3}) and we discuss their expressiveness, time complexities, and applications.
%
%
Our work will help to design more powerful future GNNs.

\subsection{Scope of this Work vs.~Related Surveys \& Analyses}

We focus specifically on HOGNNs based on the MP paradigm.
We exclude works related to early non-GNN higher-order learning, such as the
work by Schölkopf et al.~\cite{zhou2006learning}. We also do not focus on
spectral graph learning beyond what is related to HOGNNs and
MP~\cite{schaub2021signal}.

Our work complements existing surveys.
%
%
These works analyze HOGDMs in the context of complex physical
processes~\cite{bick_what_2022, battiston2020networks, torres2021and} and
signal processing~\cite{schaub2021signal}.
Recent works on topological deep learning~\cite{hajijtopological, 
papillon2023architectures, papamarkou2024position} covers a very broad set of topics related to
topological neural networks.
Another work~\cite{antelmi2023survey} focuses on broad representation learning
for hypergraphs.
Some other results provide blueprints for a limited class of HOGNNs, for example
Node-based Subgraph GNNs~\cite{frasca2022understanding} or others~\cite{huang2021unignn, chien2021you}.
We complement these papers by providing a taxonomy of a broad set of HOGNNs together with an accompanying general  blueprint for devising new HOGNNs. 


\iftr
There are also many related works on GNNs, but they do not focus on
higher order. They include general surveys~\cite{bronstein2021geometric,
wu2020comprehensive, zhou2020graph, zhang_deep_2022, chami2020machine,
  hamiltonGraphRepresentationLearning, bronstein2017geometric, zhang2019graph}, works on
the spatial--spectral dichotomy~\cite{chen2021bridging, balcilar2020bridging},
the expressive power of GNNs~\cite{sato2020survey}, heterogeneous
graphs~\cite{yang2020heterogeneous, xie2021survey}, analyzes of GNNs for
specific applications (knowledge graph completion~\cite{arora2020survey},
traffic forecasting~\cite{jiang2021graph, tedjopurnomo2020survey}, symbolic
computing~\cite{lamb2020graph}, recommender systems~\cite{wu2020graph}, text
classification~\cite{huang2019text}, action recognition~\cite{ahmad2021graph}),
explainability of GNNs~\cite{yuan2020explainability}, and the SW/HW
co-design~\cite{abadal2021computing}.
\fi


\section{Background, Naming, Notation}

We first establish consistent naming and notation. Table~\ref{tab:gdms} lists all less common acronyms.

\subsection{Plain Graph Data Model \& Basic Notation}
\label{sec:graphs}

The fundamental graph data model is a tuple $G=(\mathcal{V},\mathcal{E})$; $\mathcal{V}$ is a set of vertices
(nodes) and $\mathcal{E}$ is a set of edges; $|\mathcal{V}|=n$ and $|\mathcal{E}|=m$.
We will be referring to it as a \emph{plain graph (PG)} 
%
%
%
or, when it is clear from the context, as \emph{graph}. This model, by default,
does not incorporate any explicit higher-order structure information.
If edges model directed relations, we use a directed graph $G = (\mathcal{V},\mathcal{E})$ in which
the edges are a subset of ordered vertex pairs $\mathcal{E}\sub \mathcal{V}\times \mathcal{V}$.
$\mathcal{N}(v)$ denotes the set of vertices adjacent to vertex (node)~$v$,
$d_v$ is $v$'s degree, and $d$ is the maximum degree in $G$.
The {adjacency matrix} $\textbf{A} \in \{0,1\}^{n \times n}$ of a graph
determines the connectivity of vertices: $\textbf{A}(i,j) = 1 \Leftrightarrow
(i,j) \in \mathcal{E}$.

We call two graphs $G_1=(\mathcal{V}_1,\mathcal{E}_1), G_2=(\mathcal{V}_2,\mathcal{E}_2)$ \emph{isomorphic} if there
exists a bijection $\varphi\colon \mathcal{V}_1\to \mathcal{V}_2$ that preserves edges, i.e.,
$\{\varphi(u),\varphi(v)\}\in \mathcal{E}_2 \dimp \{u,v\}\in \mathcal{E}_1$, for all $u,v\in \mathcal{V}_1$.
An isomorphism of directed graphs requires $(\varphi(u),\varphi(v))\in \mathcal{E}_2
\dimp (u,v) \in \mathcal{E}_1$ for all $u,v\in \mathcal{V}_1$.

The {input, output, and latent} feature vector of a vertex~$i$ are denoted
with, respectively, $\textbf{x}_i, \textbf{y}_i, \textbf{h}_i \in
\mathbb{R}^{k}$, $k$ is the number of features\footnote{\scriptsize For
clarity, we use the same symbol for the number of input, output, and latent
features; this does not impact the generality of any insights in this work.}.
These vectors can be grouped in matrices, denoted respectively as $\textbf{X},
\textbf{Y}, \textbf{H} \in \mathbb{R}^{n \times k}$.
If needed, we use the iteration index $(l)$ to denote the latent features in a
given iteration (e.g., a GNN layer)~$l$ (e.g., $\textbf{h}^{(l)}_i$,
$\textbf{H}^{(l)}$).

We denote multisets with double brackets $\multiset{\cdot}$. 
%
%
%
For a set $\mathcal{V}$, we
denote its power set by $2^\mathcal{V}=\{W:\ W\sub \mathcal{V}\}$.
%
%
For a non-negative integer
$n\in \nat$ let $[n]=\{1,\dots,n\}$. $A\cong B$ is an isomorphism between $A$
and $B$. $\mathbbm{1}$ is the indicator function, i.e., $\mathbbm{1}_{p}=1$ if
$p$ is true and $\mathbbm{1}_{p}=0$ otherwise.

\iftr 
\subsection{Graph Representation Learning}

In general \textbf{representation learning (RL)}, one aims to find
transformations which extract useful information from raw input
data~\cite{bengio_representation_2013}. Denote the input space with
$\mathcal{S}$ and the feature space with $\mathcal{R}$. Then, a
\emph{representation} is a parametric function $\mathcal{F}_{\Theta}\colon
\mathcal{S}\to \mathcal{R}$ with parameters $\Theta$. For an element $s\in
\mathcal{S}$, we call $\mathcal{F}_{\Theta}(s)$ its \emph{feature
representation} or simply the \emph{representation} of $s$.  Subsequently, one
can use the transformed inputs $\mathcal{F}_{\Theta}(s)$ to perform downstream
ML tasks.
 
Since raw input data often contains information irrelevant to the ML
problem or does not have the desired format, it is preprocessed first. In the
past, the standard approach was \emph{feature engineering}. Based on domain
knowledge, practitioners would design transformations which map the raw input
to vectors containing relevant information, constituting the input for an ML
method.  Representation learning is a step towards automating feature
engineering. Instead of choosing which features a transformation should
extract, the practitioner chooses a hypothesis space
$\{\mathcal{F}_{\Theta}\}_{\Theta}$ and designs an algorithm which infers the
parameters $\Theta$ of the representation $\mathcal{F}_{\Theta}$, which is used
to compute the features. Representation learning has successfully been applied
to multiple {ml} subdomains, most prominently in computer
vision~\cite{voulodimos_deep_2018}, speech
recognition~\cite{malik_automatic_2021}, natural language
processing~\cite{chowdhary_natural_2020} and recommender
systems~\cite{zhang_deep_2019}.

One of the main challenges in designing RL methods is finding transformations
which appropriately capture the symmetries in the
data~\cite{bronstein2021geometric}. For example, an image representation
should not depend on planar translations of the image. This consideration has
led to the design of convolutional neural network (CNN)
architectures~\cite{li_survey_2020}, which respect the translational invariance
symmetry. However, while images are stored as a regular grid of pixels,
networked data often exhibit a more complex structure.

\textbf{Graph representation learning
(GRL)}~\cite{hamilton_representation_2018} refers to a collection of ML
techniques, in which one learns a new feature function $\textbf{h}\colon \mathcal{V} \cup
\mathcal{E} \cup \{G\} \to \RE^{d_{\text{out}}}$ for a graph $G=(\mathcal{V},\mathcal{E})$ with
initial features $\textbf{x}\colon \mathcal{V} \cup \mathcal{E} \cup \{G\} \to
\RE^{d_{\text{in}}}$. The inferred features $\textbf{h}$ are used for
downstream ML tasks - for example, node prediction or graph prediction.
In this work, we focus on two fundamental GRL techniques: Graph Neural Networks
(GNNs) and Random Walks (RWs).
\fi

\subsection{Graph Neural Networks (GNNs)}
\label{sec:gnns-summary}

Each vertex and often each edge are associated with
\emph{input feature vectors} that carry task-related information.
For example, if nodes and edges model papers and citations, then a node input
feature vertex could be a one-hot encoding of the presence of
words in the abstract.
When developing a GNN model, one specifies how to transform the input, i.e.,
the graph structure~$\textbf{A}$ and the input features~$\textbf{X}$, into the
output feature matrix~$\textbf{Y}$ (unless specified otherwise,
$\textbf{X}$ models vertex features). 
In this process, intermediate \emph{hidden latent vectors} are often created.
One updates these hidden features \emph{iteratively}, usually more than once.
A single iteration is called a \emph{GNN layer}.
Finally, \emph{output feature vectors} are used in \emph{downstream ML tasks},
such as node classification.

A single \emph{GNN layer} consists of a \emph{graph-related} operation (usually
sparse), an operation related to \emph{traditional neural networks} (usually
dense), and a non-linear activation (e.g., ReLU~\cite{kipf_semi-supervised_2017}) and/or
normalization.
An example sparse operation is graph convolution~\cite{kipf_semi-supervised_2017} in
which each vertex~$v$ generates a new feature vector by summing and transforming the
features of its neighbors.
Example dense operations are MLPs or
linear transformations.

\iftr
\subsubsection{Local vs.~Global Formulations}
\label{sec:gnns-loc-glo}
\fi

\iftr
Many GNN models are specified with a so called \textbf{local formulation}.
Here, to obtain the latent feature vector~$\textbf{h}_i$ of a given node~$i$ in
a given GNN layer, one aggregates the feature vectors of the neighbors~$\mathcal{N}(i)$
of~$i$ using a \emph{permutation invariant} aggregator~$\bigoplus$, such as sum
or max. 
\else
To obtain the latent feature vector~$\textbf{h}_i$ of a given node~$i$ in
a given GNN layer, one aggregates the feature vectors of the neighbors~$\mathcal{N}(i)$
of~$i$ using a \emph{permutation invariant} aggregator~$\bigoplus$, such as sum
or max. 
\fi
In the process, feature vectors of the neighbors of~$i$ may be
transformed by a function~$\psi$. Finally, the aggregator outcome is usually
further modified with another function~$\phi$. In summary, one obtains a
feature vector~$\textbf{h}^{(l+1)}_i$ of a vertex~$i$ in the next GNN layer
$l+1$ as 

\vspace{-2em}
\small
\begin{gather}
\textbf{h}^{(l+1)}_i = \phi \left( \textbf{h}^{(l)}_i, \bigoplus_{j \in \mathcal{N}(i)} \psi\left(\textbf{h}^{(l)}_i, \textbf{h}^{(l)}_j \right) \right) \label{eqn:mpnn}
\end{gather}
\normalsize

Different forms of $\psi$ are a basis of three GNN classes: \emph{Convolutional
GNNs} (C-GNNs), \emph{Attentional GNNs} (A-GNNs), and \emph{Message-Passing
GNNs} (MP-GNNs).
$\psi$ returns a product of $\textbf{h}_j^{(l)}$ with -- respectively -- a
fixed scalar coefficient (C-GNNs), a learnable function that returns a scalar
coefficient (A-GNNs), and a learnable function that returns a vector
coefficient (MP-GNNs)~\cite{bronstein2021geometric}.
%
%
%
For example, in the seminal GCN model~\cite{kipf_semi-supervised_2017},
$\textbf{h}_i^{(l+1)} = ReLU \fRB{ \textbf{W}^{(l)} \times \fRB{ \sum_{j \in
\mathcal{\widehat{N}}(i)} \frac{1}{\sqrt{(d_i + 1)(d_j + 1)}} \textbf{h}_j^{(l)} }}$.
Here, $\bigoplus$ sums $\mathcal{N}(i) \cup \{i\} \equiv \mathcal{\widehat{N}}(i)$, $\psi$ returns
a product of each neighbor~$j$'s feature vector with a scalar~$1 / \sqrt{d_i
d_j}$, and $\phi$ is a linear projection followed by $ReLU$.''

\iftr
Some GNN models also have \textbf{global linear algebraic} formulations, in
which one employs operations on whole matrices $\textbf{X}$, $\textbf{H}$,
$\textbf{A}$, and others.
For example, in the GCN model, $\textbf{H}^{(l+1)} = ReLU(\widehat{\textbf{A}}
\textbf{H}^{(l)} \textbf{W}^{(l)})$, where $\widehat{\textbf{A}}$ is the
\emph{normalized adjacency matrix with self loops}.
\fi

\iftr
A standard chain rule is used to obtain mathematical formulations for
the backward pass of respective GNN models~\cite{tripathy2020reducing}.
One can prove that the backward formulations can be
constructed using the same linear algebra kernels (sparse matrix--dense matrix product,
sampled dense--dense product) as the forward pass~\cite{wang2019deep}.
\fi

\section{Why Higher-Order GNNs?}
\label{sec:motivate}

We now discuss motivation \& applications of HOGNNs.

\subsection{Greater Expressive Power}

Perhaps the strongest motivation is to overcome the limited expressivity of traditional GNNs, which are typically at most as powerful as the 1-dimensional Weisfeiler–Lehman test (1-WL). A 1-WL GNN cannot distinguish certain non-isomorphic graphs (e.g. regular graphs or those differing by higher-order structures like cycles vs.~chords) because it only aggregates tree-like neighborhoods. HOGNNs substantially boost expressivity by encoding higher-order neighborhoods and relationships. Theoretically, many HOGNNs are provably more powerful than 1-WL~\cite{morris_weisfeiler_2021}. For instance, k-node tuple networks (k-GNNs) can match the graph discrimination power of the $k$-WL test for any $k\ge 2$~\cite{morris_weisfeiler_2021}.
\iftr
Even at lower complexity, subgraph-based GNNs (like those augmenting features with cycles or cliques) are strictly more expressive than 1-WL and not less powerful than 3-WL~\cite{bevilacqua_equivariant_2022}. Cell and simplicial complex networks lift the graph to a higher-dimensional combinatorial object, thereby capturing structures like loops, cavities, and multi-node motifs in their message-passing – these models have been shown to detect graph properties and patterns that plain GNNs miss. In practice, this translates to better performance on tasks where configuration of more than two nodes matters. For example, a cell complex GNN can understand that a triangle of nodes forming a cycle is a fundamentally different ``signal'' than three nodes connected in a tree shape, because the triangle can be modeled as a 2D cell~\cite{bodnar_weisfeiler_2022}. Classic GNNs treat both as just three edges. HOGNNs thus break the symmetry that 1-WL fails to break, allowing the network to learn functions that depend on higher-order patterns (such as graph cycles, cliques, or chordal structures).
\fi
This improved expressivity is crucial for graph classification, isomorphism testing, and any task where global structure is not a simple sum of pairwise relations.

\iftr
For example,
Figure~\ref{fig:WL_limits} shows two non-isomorphic graphs $G_1,G_2$.  Colors
indicate node features and, for each node, we portray its one-hop neighborhood
in the adjacent surface enclosed by dashed lines.

\begin{figure}[h]
    \centering
        \includegraphics[width=1.0\linewidth]{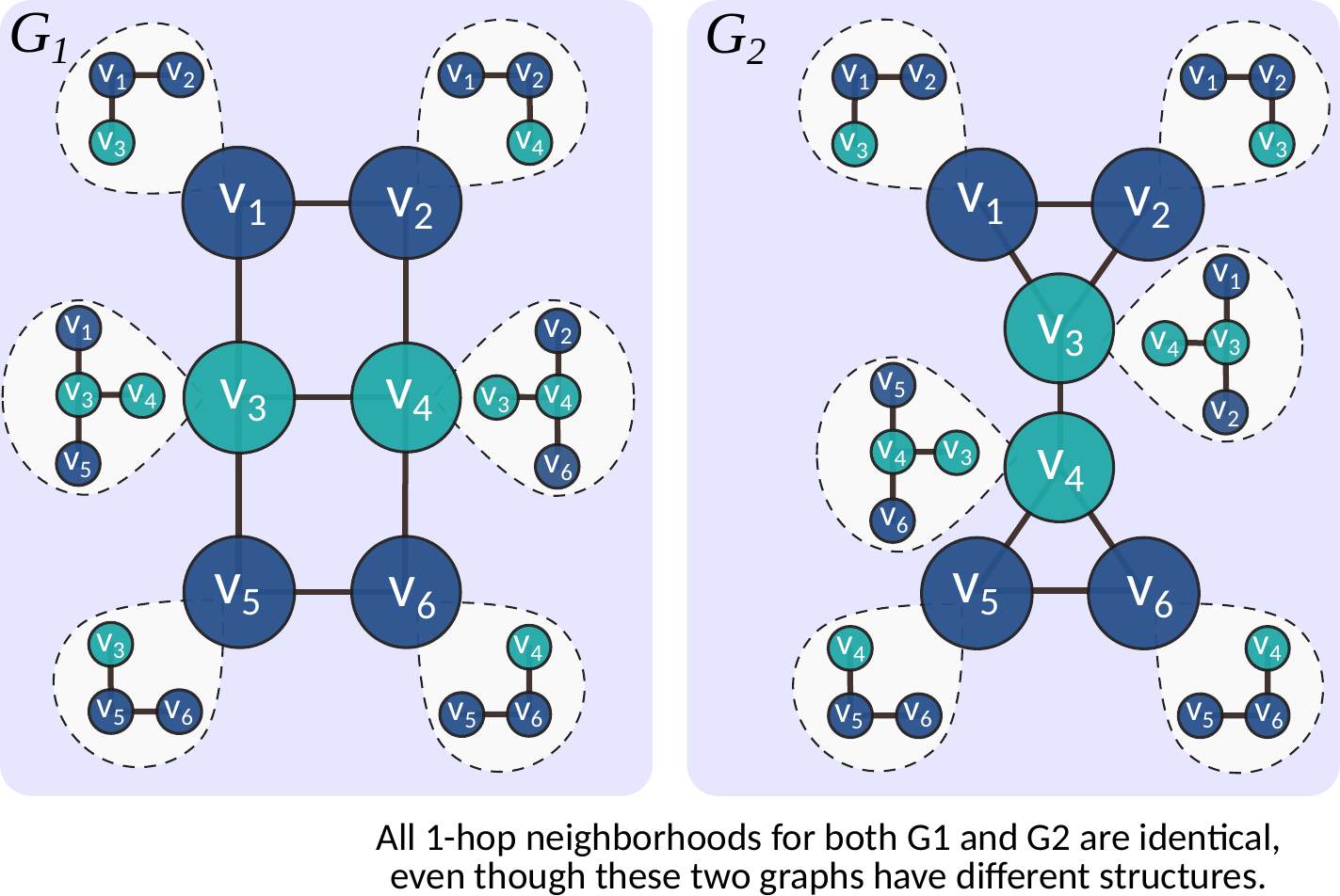}
    \caption{Computational structure of aggregations for a given vertex in simple 1-hop GNNs (it includes 1-hop neighbors and a given vertex). Graphs $G_1$ and $G_2$ are non-isomorphic but cannot be distinguished by a simple GCN.}
    \vspace{-1.5em}
    \label{fig:WL_limits}
\end{figure}

In the best case, a node $v\in \mathcal{V}$ retains all features of the nodes in its one-hop
neighbourhood $\mathcal{N}(v)$ in a single iteration. Since a permutation-invariant
aggregation $\bigoplus$ is applied to incoming messages
$\psi(\textbf{x}_v,\textbf{x}_u)$, a node only sees the \emph{set} of transformed
features $\{\psi(\textbf{x}_v,\textbf{x}_u): u\in \mathcal{N}(v)\}$ and cannot
associate them with specific nodes. Thus, the most information about features
that can be obtained through one iteration is $\textbf{x}_v^{\text{new}}
= (\textbf{x}_v, \{\textbf{x}_u: u\in \mathcal{N}(v)\})$.
Note that for the graphs $G_1, G_2$, the one-hop neighborhoods of the corresponding vertices look the same. The same will hold after each further MP step (only the messages change).
Thus, the readout function will map these graphs to the same value. However,
the graphs are not isomorphic. For example, $G_{2}$ contains two triangles,
while $G_1$ contains none. 
This problem has sparked an increased interest in GRL designs that incorporate
\emph{polyadic} relationships, aka higher-order structures, into their learning
architecture.

\fi


\subsection{Overcoming Over-Smoothing}

When GNNs are stacked deep, node embeddings tend to become indistinguishable (converging to similar values), a problem known as \textit{over-smoothing}~\cite{zhang2024beyond, giraldo2023trade, nguyen2023revisiting}. This happens because repeated neighbor averaging drives representations toward a common subspace~\cite{arroyo2025vanishing}. Higher-order models can delay or avoid this effect~\cite{yadatioversquashing, tahademystifying, einizade2025continuous}. Firstly, HOGNNs often do not require as many layers to capture global context: a single higher-order message-passing step can capture what would take many two-node hops. Thus, the model can remain shallow and expressive. Secondly, the richer structures in HOGDMs provide additional degrees of freedom in propagation.
\iftr
Messages can be passed on hyperedges or through k-simplex connections that inject new information at each layer, preventing the collapse into overly homogeneous embeddings. 
\fi
Here, because HOGNNs incorporate more structure than just the graph adjacency, they have been shown to maintain heterogeneity in node features even with multiple propagation steps.
\iftr
Empirically, many HOGNN variants have been noted to preserve differentiation of nodes better than their graph-GNN counterparts.
\fi

\subsection{Overcoming Over-Squashing}

\textit{Over-squashing}~\cite{nguyen2023revisiting} is a phenomenon where information from distant parts of a graph is ``squeezed'' into a fixed-size embedding, causing long-range dependencies to vanish. In standard GNNs, a node's representation becomes insensitive to far-away nodes because many hops of aggregation compress an exponential neighborhood into one vector. HOGNNs combat this by directly modeling higher-order interactions, which effectively shorten path lengths and broaden the communication bandwidth~\cite{yadatioversquashing}.
\iftr
For example, a hypergraph model can connect a group of nodes in one hop via a hyperedge instead of requiring multiple pairwise hops; similarly, a simplicial or cell complex can propagate messages through higher-dimensional structures (faces, cells) that span multiple nodes. By introducing these macro-scale connections, HOGNNs allow information to flow less diffusely: important global signals can reach a node in fewer steps, relieving the pressure on any single aggregation step's capacity.
\fi

\section{Taxonomy \& Blueprint}

%
In general, when analyzing or constructing a HOGNN architecture, one must consider the details of the harnessed \textbf{graph data model (GDM)} and the details of the \textbf{neural architecture} that harnesses a given GDM.
We first discuss the HO aspects of the GDM (Section~\ref{sec:gdm-framework}) and later those of the GNN architecture based on that GDM (Section~\ref{sec:hognn-framework}). 
We then describe how prescribing these aspects forms a blueprint for new HOGNN architectures (Section~\ref{sec:hognn-blueprint}).
%
%

\begin{figure*}[hbtp]
\vspace{-1em}
    \centering
    \includegraphics[width=0.96\textwidth]{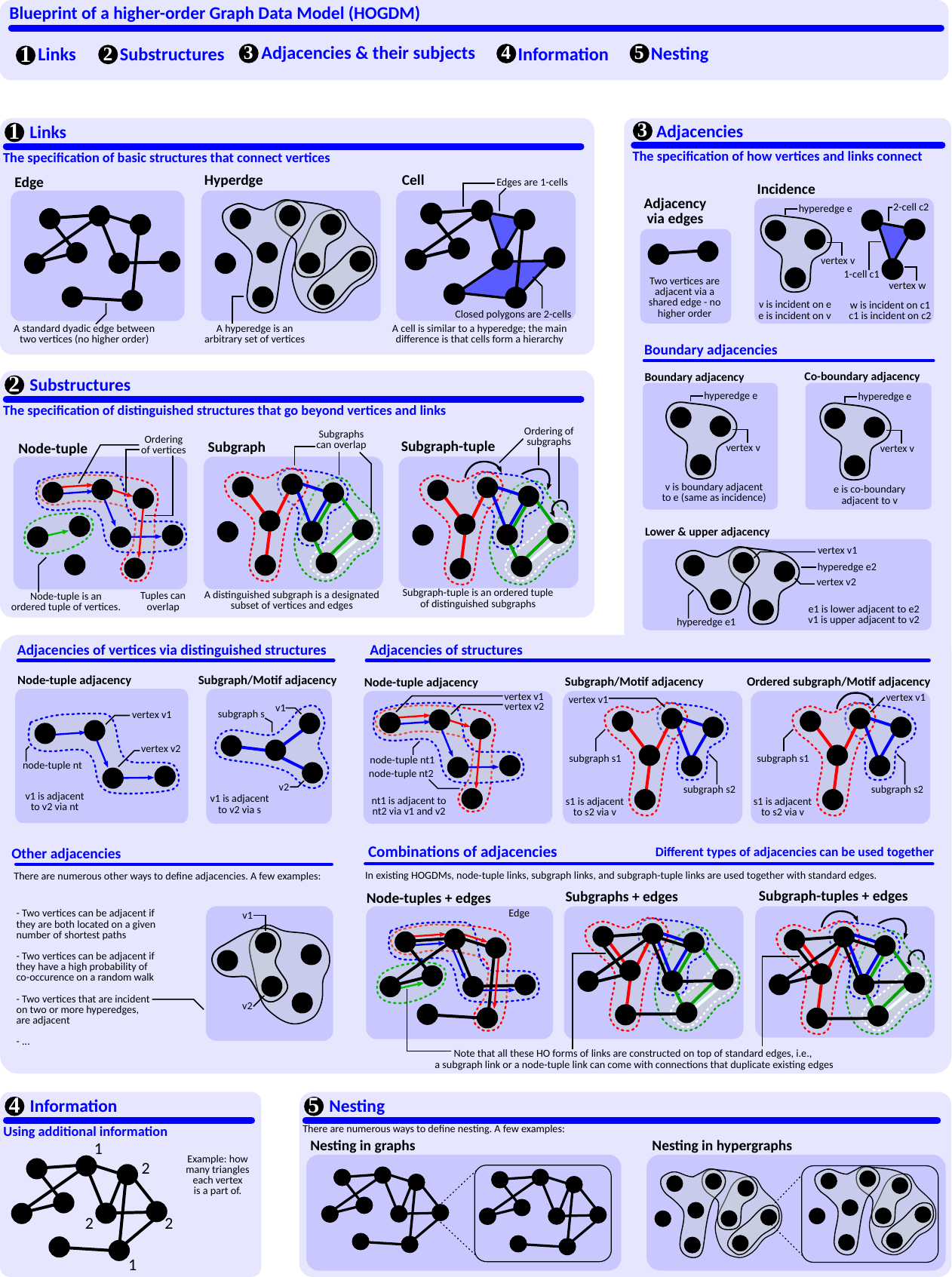}
    \caption{A blueprint of a higher-order graph data model (HOGDM).}
        \vspace{-1em}
    \label{fig:hogdm-blueprint}
\end{figure*}

\begin{figure*}[hbtp]
\vspace{-1em}
    \centering
    \includegraphics[width=0.96\textwidth]{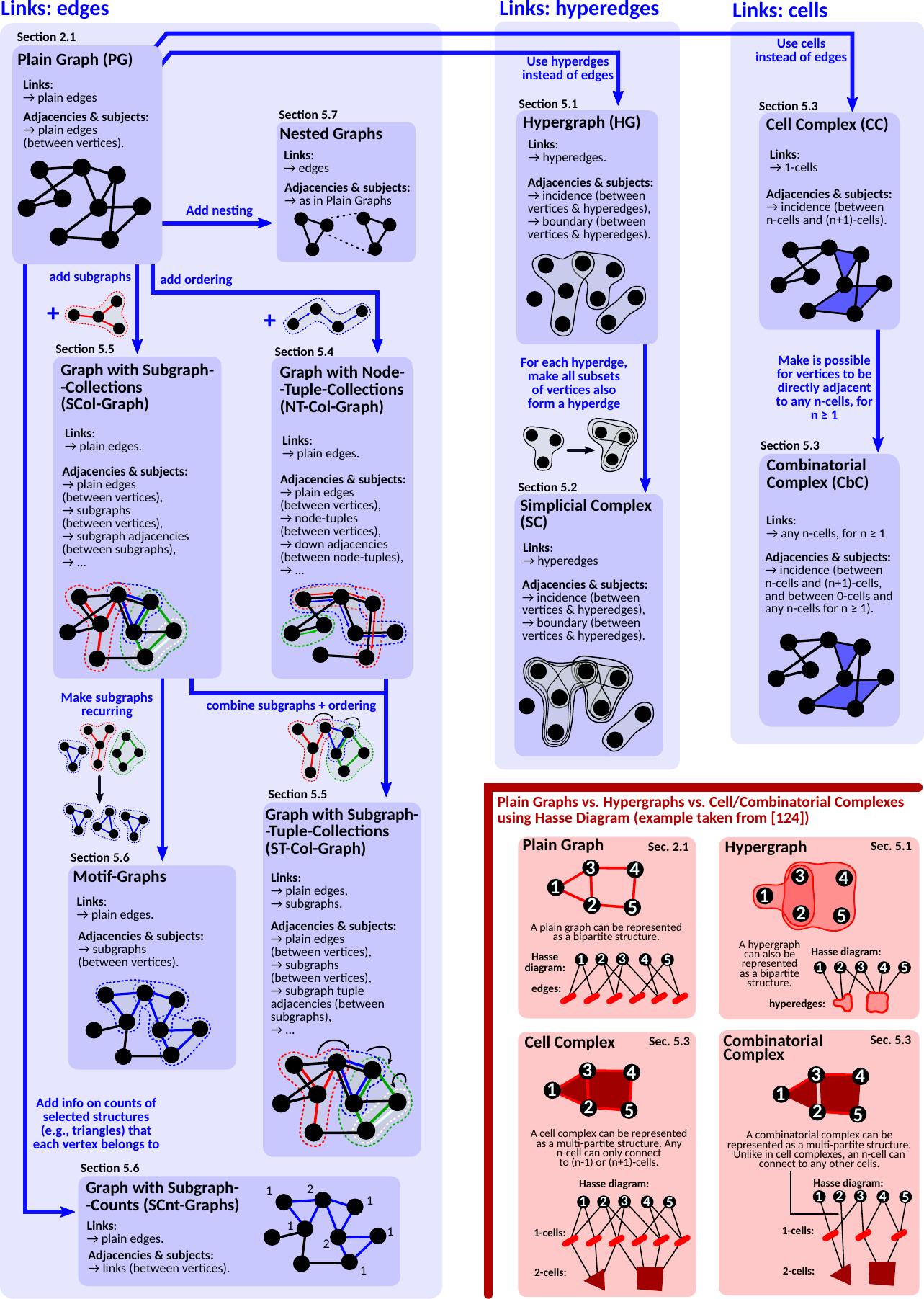}
    \caption{A taxonomy of higher-order graph data models (HOGDMs).}
        \vspace{-1em}
    \label{fig:hogdms}
\end{figure*}

\subsection{Higher Order in Graph Data Models}
\label{sec:gdm-framework}

Figure~\ref{fig:hogdm-blueprint} illustrate the blueprint of HOGDMs. The taxonomy of HOGDMs is illustrated in Figure~\ref{fig:hogdms} and in Table~\ref{tab:gdms}.

The first fundamental element of any HOGDM is a specification of \textbf{links between vertices}. For example, in PGs, a link is an ordered tuple of two vertices (in directed PGs) or a set of two vertices (in undirected PGs). In hypergraphs, a link (called a hyperedge) is a set of \textit{arbitrarily many} vertices. In cell complexes, one uses cells which form a hierarchy of connections that can be pictured with Hasse diagrams (see the red part of Figure~\ref{fig:hogdms}); see extensive past works~\cite{hajijtopological, Hajij2023CombinatorialCB} for details on Hasse diagrams.

Another fundamental GDM element is the \textbf{adjacency}: the specification of how vertices, links, and potentially other parts of a given GDM connect. The adjacency always specifies \textbf{what is connected} (e.g., vertices, subgraphs) and \textbf{how it is connected} (e.g., vertices are adjacent when they share an edge). In PGs, adjacency is fully determined by links. However, one can make it richer. For example, in a GDM that we refer to as ``Motif-Graph'' and detail in Section~\ref{sec:graph_motifs}, two vertices can be adjacent if they both belong to one triangle (or another specified subgraph). Other examples of node adjacency notions are based on shortest paths or the probability of co-occurrences in a random walk. In hypergraphs, an example of adjacency between two nodes can also be based on the number of hyperedges these two nodes are incident on.

The third fundamental HOGDM element is the specification of potential \textbf{distinguished substructures} among vertices, links, and adjacencies. For example, in Subgraph GNNs (detailed in Section~\ref{sec:graph_with_subgraphs}), one uses message-passing over graphs with distinguished collections of vertices and edges. Such structures are usually specified by appropriately extending a given GDM definition; in the above example of Subgraph GNNs, a definition of specific vertex or edge collections. These structures are often assigned feature vectors, which are updated in each GNN layer similarly to those of vertices and edges.

One can also introduce HO by associating \textbf{additional information} with vertices, links, and distinguished substructures. For example, one could enhance each vertex with the number of triangles it is part of.

Finally, while an individual vertex usually models a fundamental entity (as in plain graphs or in hypergraphs), it can also model a \textbf{nested graph}, as in nested GNNs. For example, when modelling molecules and their interactions, the higher-level graph represents molecules as vertices and their interactions as (hyper-)edges. Moreover, each molecule is represented as a plain graph in which atoms form vertices and atomic bonds form edges.

\iftr
We will discuss in more detail how all these HO aspects are harnessed by different existing HOGDMs in Section~\ref{chap:graph_data_models}.
\fi


\begin{table*}[t]
\centering
\footnotesize
\scriptsize
\setlength{\tabcolsep}{4pt}
\renewcommand{\arraystretch}{0.6}
\begin{tabular}{@{}lllllll@{}}
\toprule
\multicolumn{2}{c}{{\textbf{GDM name, example reference, and abbreviation}}} & \makecell{\textbf{Used links} \\ \textbf{between vertices}} & \makecell{\textbf{Harnessed} \\ \textbf{adjacency notions}} & \makecell{\textbf{Distinguished} \\ \textbf{substructures}} & \makecell{\textbf{Additional} \\ \textbf{information}} & \textbf{Nesting} \\
%
%
%
\midrule
Plain graph & PG & edge & incidence & --- & --- & --- \\
\midrule
Hypergraph~\cite{arya2020hypersage} & HG & hyperedge & Incidence, boundary & --- & --- & --- \\
Nested Hypergraph~\cite{yadati2020neural} $\bigstar$ & Nested-HG & hyperedge & Incidence, boundary & --- & --- & Yes \\
Simplicial Complex~\cite{bodnar_weisfeiler_2021} & SC & hyperedge/cell & Incidence, boundary & simplices & --- & --- \\
Cell Complex~\cite{bodnar_weisfeiler_2022} & CC & cell & Incidence, boundary & --- & --- & --- \\
Combinatorial Complex~\cite{hajijtopological} & CbC & cell & Incidence, boundary & --- & --- & --- \\
\midrule
%
%
%
Graph with Node-Tuple-Collections~\cite{morris_weisfeiler_2021} $\bigstar$ & NT-Col-Graph & edge, node-tuple & edge, tuple adjacency & node tuples & --- & --- \\
Graph with Subgraph-Collections~\cite{xu2018powerful} $\bigstar$ & SCol-Graph & edge, subgraph & edge, subgraph adjacency & subgraphs & --- & --- \\
Graph with Subgraph-Tuple-Collections~\cite{qian2022ordered} $\bigstar$ & ST-Col-Graph & edge, subgraph-tuple & subgraph-tuple adjacency & subgraph-tuples & --- & --- \\
Graph with Motifs~\cite{rossi_hone_2018} $\bigstar$ & Motif-Graph & edge & motif adjacency & motif & --- & --- \\
Graph with Subgraph-Counts~\cite{bouritsas_improving_2021} $\bigstar$ & SCnt-Graph & edge & edge & --- & subgraph count & --- \\
Nested Graph~\cite{wang_gognn_2020} $\bigstar$ & Nested-Graph & edge & edge & --- & --- & Yes \\
\bottomrule
\end{tabular}
\vspaceSQ{-0.5em}
\caption{\textbf{Comparison of considered higher-order graph data models (HOGDMs) with respect to the taxonomy introduced in Section~\ref{sec:gdm-framework}}. 
``$\bigstar$'' indicates a graph data model formally stated in this work.
%
%
%
}
%
%
\label{tab:gdms}
\vspaceSQ{-1.5em}
\end{table*}

\subsection{Higher Order in GNN Architectures}
\label{sec:hognn-framework}

The central part of specifying an HOGNN architecture is determining the \textbf{message-passing (MP)} channels that will be used in GNN layers. We refer to this step as \textbf{wiring}.
Formally, wiring creates a set of tuples $W = \{(x,y)\}$ where $x, y$ can be any parts of the used GDM, such as vertices, links, and any substructures. These tuples form MP channels that are then used to exchange information in each GNN layer.
The exchanged data are feature vectors; thus, $x$ and $y$ from each element of $W$ have feature vectors.

An important aspect of an imposed wiring is its \textbf{flavor}. Directly extending the notion of flavor in MP over plain graphs~\cite{bronstein2021geometric}, we distinguish \textit{convolutional}, \textit{attentional}, and \textit{general message-passing} flavors. However, while in the GNNs over plain graphs it was straightforward to define these flavors based on the different forms of $\psi$, in HOGNNs, it becomes more complicated, because model formulations can be very complex. For example, in HOGNNs based on hypergraphs or simplicial complexes, exchanging a message between two vertices may involve generating multiple feature vectors assigned to intermediate steps within one GNN layer. For this, we use the following definition: an HOGNN model is convolutional, attentional, or general message-passing if -- respectively -- all the functions used by the model return fixed scalar coefficients, at least one function is learnable and all the functions return scalar coefficients, and at least one function is learnable and it returns a vector coefficient.
Note that these flavors can be used together. For example, in Motif-graphs, one could have convolutional message-passing along edge-based adjacencies and, in addition, attentional message-passing along adjacencies defined by being in a common triangle.

\iftr

Another relevant aspect of the harnessed wiring pattern, is whether it uses \textbf{multi-hop channels}. These are wiring channels that connect vertices or links which are multiple steps of adjacency away from each other. Such channels may be useful when dealing with issues such as oversmoothing.

A technical aspect of constructing a HOGNN is whether to use the \textbf{local or the global formulation} (or whether use a combination of both), when constructing the MP channels.
As we illustrate in Section~\ref{chap:from_gdm_to_repL}, most HOGNN architectures use either the local formulation, or a mixture of local and global formulations, when prescribing MP channels.
The local formulation is usually easier to develop, but the global formulation may result in lower running times of the GNN computation, as it makes it easier to take advantage of features such as vectorization~\cite{besta2023high}.
\fi

Finally, when building a HOGNN architecture, one must consider how to \textbf{transform the input dataset into, and from, an HO format}.
In many datasets, the data is stored as a plain graph~\cite{morris_tudataset_2020, hu_open_2020}. 
We refer to a transformation from the plain to the HO format as \emph{lifting}. We also denote a mapping from an HOGDM to a PG as \textit{lowering}.
When conducting a lifting, one usually does not want to lose (or change) any structural information. For example, one usually wants to preserve isomorphism properties.
Formally, we have

\begin{definition}[Graph data model lifting]\label{def:lifting}
Let $\mathfrak{G}$ be the class of graphs and $\mathfrak{K}$ a \acrlong{hogdm} equipped with a notion of isomorphism. A GDM \emph{lifting} from $\mathfrak{G}$ to $\mathfrak{K}$ is a map $f\colon \mathfrak{G} \to \mathfrak{K}$ that preserves isomorphisms. That is, for any $G_1,G_2\in \mathfrak{G}$, $G_1$ and $G_2$ are isomorphic if and only if $f(G_1)$ and $f(G_2)$ are isomorphic.
\end{definition}

In contrast, lowerings generally introduce a loss of information, as we will see in Section~\ref{sec:hg-pg}.

\begin{figure*}[t]
\vspaceSQ{-1em}
    \centering
    \includegraphics[width=1.0\textwidth]{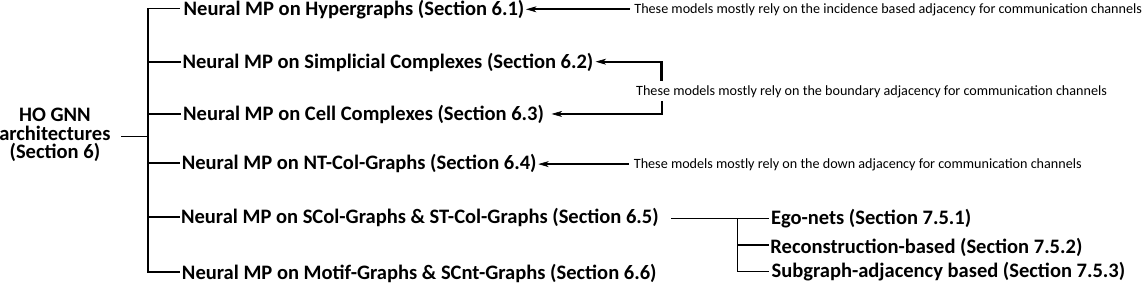}
    \caption{A taxonomy of higher-order GNN architectures.}
    \label{fig:hognns}
\end{figure*}

We discuss in more detail how all these HO aspects are harnessed by different existing HOGNNs in Section~\ref{chap:from_gdm_to_repL}.
Figure~\ref{fig:hognns} illustrates the taxonomy of HO architectures.
Further details on liftings between hypergraphs and related HOGDMs (simplicial complexes, cell complexes, and combinatorial complexes) as well as a framework implementing these liftings can be found in other works~\cite{telyatnikov2024topobenchmarkx, hajijtopological}.

\subsection{Blueprint \& Pipeline for HOGNNs}
\label{sec:hognn-blueprint}

\begin{figure}[h]
\vspaceSQ{-1em}
    \centering
    \includegraphics[width=\linewidth]{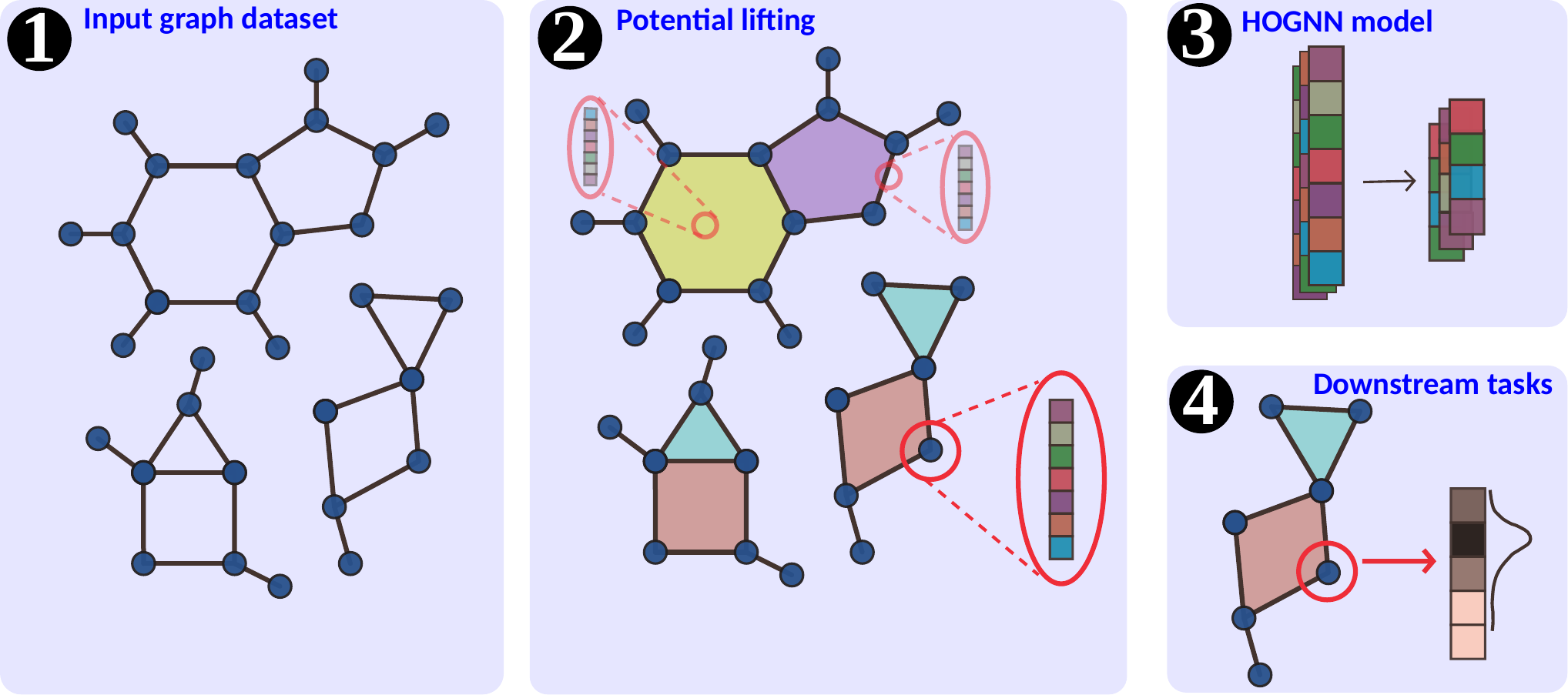}
    \caption{A typical HOGNN data pipeline. (1) The input dataset is a
    set of one or more graphs $\{G=(V,E,\textbf{x})\}$. (2) The input is
    lifted to a selected HOGDM (HOGDMs are detailed in Section~\ref{chap:graph_data_models}). Vectors in the red ovals indicate $\RE^k$-valued features. In the HOGDM, we commonly have features for higher-order
    structures, for example, edges and hyperedges.
        (3) The constructed HOGNN architecture transforms features (HOGNNs are detailed in Section~\ref{chap:from_gdm_to_repL}). (4)
        Final features are fed to downstream tasks, for example, node
        prediction.}
        \vspaceSQ{-1em}
    \label{fig:GRL_data_pipeline}
\end{figure}

In our blueprint for creating HOGNNs with desired properties, one first specifies a HOGDM by selecting the HO aspects described in Section~\ref{sec:gdm-framework}. This includes selecting a form of links, adjacencies, distinguished substructures, additional information, and nesting. Many of possible selections result in already existing HOGDMs, for example, if using hyperedges as links, one obtains a hypergraph as the GDM. Many other selections would result in novel GDMs, with potentially more powerful expressiveness properties.
Next, one specifies the details of the HO neural model, as discussed in Section~\ref{sec:hognn-framework}. This includes details of wiring and of how the feature vectors are transformed between GNN layers, and the specifics of harnessed lifting(s) and lowering(s).
A typical HOGNN pipeline is in Figure~\ref{fig:GRL_data_pipeline}.

\section{Higher-Order Graph Data Models} 
\label{chap:graph_data_models}

We now investigate GDMs used in HOGNNs. 
First, we analyze the existing established GDMs: \emph{hypergraphs (HGs)} (Section~\ref{sec:hypergraphs}) as well as their specialized variants, namely, \emph{simplicial complexes (SCs)} (Section~\ref{sec:simplicial-complexes}) and \emph{cell complexes (CCs)} (Section~\ref{sec:cell-complexes}).
Next, we introduce \textit{new} HOGDMs that formally capture data models used in various existing HOGNNs.
These are \emph{graphs equipped with node-tuple collections (NT-Col-Graphs)} (Section~\ref{sec:graph_with_node_tuples}), \emph{graphs equipped with subgraph collections (SCol-Graphs)} (Section~\ref{sec:graph_with_subgraphs}), \emph{graphs equipped with subgraph-tuple collections (ST-Col-Graphs)} (Section~\ref{sec:graph_with_subgraphs}), \emph{graphs equipped with motifs (Motif-Graphs)} (Section~\ref{sec:graph_motifs}), \emph{graphs equipped with subgraph counts (SCnt-Graphs)} (Section~\ref{sec:graph_motifs}), and \emph{Nested-GDMs} (Section~\ref{sec:nested-gdms}).
%
%


We summarize the considered GDMs in terms of the provided GDM blueprint from Section~\ref{sec:gdm-framework} in Table~\ref{tab:gdms}.
{In general}, HGs, SCs, and CCs introduce HO by harnessing different notions of links and adjacencies beyond plain dyadic interactions. NT-Col-Graphs, SCol-Graphs, and ST-Col-Graphs utilize certain distinguished structures. Motif-Graphs harness motif-based forms of adjacency. SCnt-Graphs count subgraphs as distinguished information. Finally, some models use nesting of graphs or hypergraphs within vertices.

In PGs, links are edges between \textit{two} vertices. In HGs and their specialized variants (SCs, CCs), HO is introduced by making edges being able to link \textit{more than two vertices}.
In NT-Col-graphs, SCol-graphs, and ST-Col-graphs, one introduces HO by distinguishing collections of substructures \textit{in addition to harnessing edges between vertices}.

\subsection{Hypergraphs} \label{sec:hypergraphs}

Hypergraphs generalize plain graphs by allowing arbitrary subsets of entities to form \emph{hyperedges}.

\begin{definition}\label{def:attributed_hypergraph}
    A \emph{HG} with features is a tuple $H=(\mathcal{V},\mathcal{E},\mathbf{x})$
    comprised of a set of nodes $\mathcal{V}$, hyperedges $\mathcal{E}\subset2^\mathcal{V}$, and
    features $\mathbf{x}\colon \mathcal{V}\cup \mathcal{E} \to \RE^k$.
\end{definition}

This enables modeling complex and diverse forms of polyadic relationships, used broadly in various domains and problems such as clustering or partitioning.

We call two HGs $\mathcal{H}_1 = (\mathcal{V}_1,\mathcal{E}_1,\mathbf{x}_1),
\mathcal{H}_2=(\mathcal{V}_2,\mathcal{E}_2,\mathbf{x}_2)$ \textbf{isomorphic}, if there is
a bijective node relabeling $\varphi\colon \mathcal{V}_1\to \mathcal{V}_2$ such that all hyperedges
and features are preserved, i.e., $\forall_{e\in 2^{\mathcal{V}_1}}\ e\in \mathcal{E}_1
\iff \{\varphi(v):v\in e\}\in \mathcal{E}_2$, $\forall_{v \in \mathcal{V}_1}\
\mathbf{x}_1(v) = \mathbf{x}_2(\varphi(v))$ and $\forall_{e\in \mathcal{E}_1}\
\mathbf{x}_1(e) = \mathbf{x}_2(\{\varphi(v):v\in e\})$.

To simplify our notation, for any vertex $v$ in $\mathcal{V}$ we write $v \in \mathcal{H}$ and for every hyperedge $e$ in $\mathcal{E}$, we write $e \in \mathcal{H}$.
Moreover, for the precision of the following GDM concepts, we also define for each vertex $v\in \mathcal{V}$ a set $\hat v = \{v\}$, and for each hyperedge $e\in \mathcal{E}$ we define $\hat{e} \equiv e$.

\subsubsection{Higher-Order Forms of Adjacency}
\label{sec:hg-adj}

The HG definition gives rise to multiple ``higher-order'' flavors of
node and hyperedge \emph{adjacency}.
Two basic flavors are \emph{incidence} and \emph{boundary adjacency}.
They were originally defined for cell complexes (CCs)~\cite{hajij_cell_2021};
we adapt them for HGs.

For a vertex $v$ and hyperedge $e$, if $v \in e$, then $v$ is \emph{incident} on $e$ and $e$ is \emph{incident} on $v$. The \emph{degree} of a vertex or hyperedge is the number of hyperedges or vertices incident on it, respectively.
%

%


\begin{definition}\label{def:hypergraph:incidence_boundary_adjacency}
Let $\mathcal{H}=(\mathcal{V},\mathcal{E})$ be a HG, and $b,c\in
\mathcal{V} \cup \mathcal{E}$. We call $b$ \emph{boundary-adjacent} on $c$ and write $b\prec c$
if $\hat{b}\subsetneq \hat{c}$.
\end{definition}

Boundary adjacency gives rise to four closely related forms of adjacency between
\simplices. For $c\in \mathcal{V} \cup \mathcal{E}$ we define
{\footnotesize\begin{align*}
    \mathcal{B}^{}(c)&=\{b\in \mathcal{V} \cup \mathcal{E}: b\prec c\} &\text{boundary adjacencies,
    } \\
    \mathcal{C}^{}(c)&=\{d\in \mathcal{V} \cup \mathcal{E}: c\prec d\} &\text{co-boundary
    adjacencies,}\\
    \mathcal{N}_{\downarrow}(c)&=\{b\in \mathcal{V} \cup \mathcal{E}:
    \exists \tau: \tau\prec b, \tau\prec c\} &\text{lower adjacencies}, \\
    \mathcal{N}_{\uparrow}(c)&=\{d\in \mathcal{V} \cup \mathcal{E}:
    \exists \tau: c\prec \tau, d\prec \tau\} &\text{upper adjacencies}.
\end{align*}}


These notions of adjacency are sometimes referred to using different names.
For example, the HyperSAGE model~\cite{arya2020hypersage} introduces the
notions of ``intra'' and ``inter-edge neighborhoods''. These are --
respectively -- co-boundary and upper adjacencies of a given vertex.

Different forms of adjacency in HGs can be modeled with plain graph
incidence within plain graphs. First, boundary adjacency and regular incidence are equivalent. When
a node is connected to an edge, the edge is a co-boundary adjacency of the
node. When two edges share a node, they are lower-adjacent. Standard adjacency
between nodes in a graph corresponds to the upper adjacency.

\subsubsection{Lowering Hypergraphs}
\label{sec:hg-pg}


Lowering a HG prescribes how to map a given HG into a plain graph
%
%
First, while in graphs nodes are related by sharing a single edge, in HGs,
nodes can potentially share multiple hyperedges. This relation gives rise to a
canonical \textbf{mapping of HGs to edge-weighted graphs}.
Here, we assign a weight $w_{uv}$ to any pair of nodes $u,v\in \mathcal{V}$ given by the
number of hyperedges they share. This mapping has been used to learn
node representations by first applying it to the HG and then employing
a GNN designed for edge-weighted graphs~\cite{bai_hypergraph_2021,
feng_hypergraph_2019}.

An HG $\mathcal{H}=(\mathcal{V},\mathcal{E})$ can also be \textbf{mapped to a
bipartite graph} $G_H = (\tilde{\mathcal{V}}=\mathcal{V} \cup \mathcal{E},\tilde{\mathcal{E}})$, in which a
node-hyperedge pair $\{u,e\}$ for $u\in \mathcal{V}, e\in \mathcal{E}$ belongs to
$\tilde{E}$ if $u\in e$~\cite{heydari_message_2022}.
%
features of incident nodes~\cite{heydari_message_2022}. 

Another scheme is \textbf{clique expansion} ~\cite{arya2020hypersage, pu2012hypergraph}, in which one
converts each hyperedge $e$ to a clique by adding edges between any two vertices
in $e$.
%

Some of the above-mentioned lowerings involve information loss in general. For example, clique expansion
applied to hyperedges $e_1 = \{v_1, v_2\}$ and $e_2 = \{v_1, v_2, v_3\}$ would
erase information about the presence of two distinctive hyperedges, instead
resulting in a plain clique connecting vertices $v_1, v_2,
v_3$~\cite{arya2020hypersage}.
Simultaneously, one can recover the initial HF from its bipartite representation~\cite{gailhard2025hygene}.


\begin{figure}[h]
    \centering
    \includegraphics[width=0.7\linewidth]{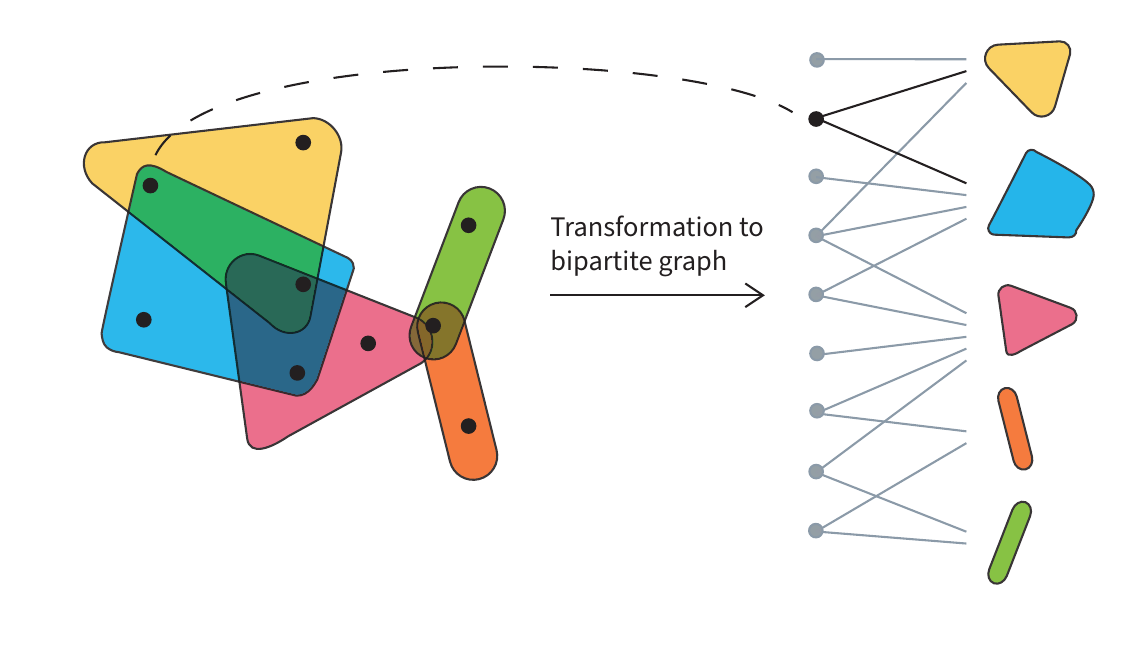}
    \caption{Two ways of describing a hypergraph:\ (1) a set of entities $\mathcal{V}$
    with subsets forming hyperedges (left), (2) a bipartite graph with nodes
    given by entities and subsets, and edges between them in case of incidence
	(right). \maciej{Florian, could you extend it with other forms of
	"Lowering" - from Section~\ref{sec:hg-pg}? I'd make it a nice
	two-column figure} \florian{Yes, we could look into that, let's discuss the details before we start drawing :)}} \label{fig:hypergraph_bipartite}
\end{figure}

These different HG lowerings have vastly different storage requirements. With edge-weighted graphs, the representation takes $O(|\mathcal{V}| + |\mathcal{E}|)$ space. Let $k$ be the largest degree of any edge in the HG. Then, the representation as a bipartite graph or using the star expansion takes $O(|\mathcal{V}| + \sum_{e\in \mathcal{E}} |e| ) \subseteq O(|\mathcal{V}| + k|\mathcal{E}|k) \subseteq  O(|\mathcal{V}| (1 + |\mathcal{E}|))$. The clique expansion takes $O(k|\mathcal{V}|)$ space.

The HG definition is very generic, and several variations that make it more
restrictive have been proposed. Two subclasses of HGs are particularly popular:
simplicial complexes (SCs) and cell complexes (CCs).

\subsection{Simplicial Complexes}
\label{sec:simplicial-complexes}

SCs restrict the HG definition to ensure stronger assumptions on the polyadic
relationships. Specifically, in SCs, vertices are also connected with hyperedges; however, if a given set of vertices is connected with a hyperedge, then \emph{all possible} subsets of these vertices also need to form a hyperedge.
%
%
This assumption is useful in, e.g., social networks, where subsets
of friend groups often also form such groups.

\begin{definition}\label{def:simplicial_complex}
%
A \emph{simplicial complex (SC)} with features $\mathcal{H}=(\mathcal{V},\mathcal{E},
\mathbf{x})$ is a HG in which for any hyperedge $e\in \mathcal{E}$, any
 nonempty subset $d\subset e$ corresponds to a vertex or hyperedge $d=\hat{c}$ for some $c\in\mathcal{H}$; as with
HGs, we have $\mathbf{x}\colon \mathcal{V}\cup \mathcal{E} \to \RE^k$.
\end{definition}

All forms of generalized \emph{adjacencies} (Section~\ref{sec:hg-adj}) are directly
applicable also to SCs.
Any plain graph satisfies the definition of an SC since the
endpoints of each edge are nodes in the graph. 

We also need further notions, used in GNNs based on SCs.
For $p\in \ZE_{\geqslant 0}$, \emph{$p$-\simplices} in an HG
$\mathcal{H}$ are all sets of nodes composed of $p+1$ distinct nodes: 
$\mathcal{H}^{p}:=\{\{v_0,\dots,v_{p}\}\in\mathcal{H}:|\{v_0,\dots,v_{p}\}|=p+1\}$.
In an SC, the $0$-\simplices ~correspond to nodes, $1$-\simplices
~to edges, $2$-\simplices ~to triangles. We call the maximal $p\in \nat$ such that
$\mathcal{H}$ contains a $p$-\simplex ~the \emph{dimension} of 
$\mathcal{H}$. Generalizing the notion of edge direction in simple
graphs, \simplices ~can have an orientation $\mathfrak{o}\colon \mathcal{H}\to
\{\pm 1\}$. For an arbitrary fixed order $\mathcal{V}=(v_1, \dots, v_n)$ of the nodes, we
say a \simplex ~$(v_{i_0}, \dots, v_{i_p})$ is positively oriented if
$(v_{i_0}, \dots, v_{i_p})$ appears in a positive permutation of
$(v_1, \dots, v_n)$ and negatively oriented otherwise. Orientation in 
SCs is incorporated into certain simplicial GNN methods by changing the
sign of messages~\cite{bodnar_weisfeiler_2021, goh_simplicial_2022}.

\subsubsection{Lifting PGs to SCs}

A commonly used technique to transform a PG into an SC is to form a hyperedge from each clique in $G$.

\begin{example}\label{ex:clique_complex_lifting}
For a simple graph $G=(\mathcal{V},\mathcal{E})$ and a maximal clique size $k$, we define a
\emph{clique complex (CqC) lifting} to be a transformation, in which we construct a
simplicial complex $\mathcal{H}=(\mathcal{V}, \tilde{\mathcal{E}})$ by setting
$\tilde{\mathcal{E}}=\{\{v_0,\dots,v_l\}: l\in [k-1], \{v_0, \dots,v_l\} \text{ are
fully connected in $G$}\}$.
\end{example}

\noindent
The resulting HG is indeed an SC because each hyperedge is defined by a fully
connected subset of nodes in $G$. Thus, any subcollection of these
nodes is also fully connected.

\begin{proposition}\label{prop:clique-complex-lifting}
CqC lifting preserves isomorphisms.
\end{proposition}

\iftr
When a simple graph has features, there are different ways to assign features
to the lifted CqC (for example, by setting the feature of the clique \simplex ~to
the average or sum of node features in the clique).
\fi

Note that all forms of representing HGs with PGs 
(Section~\ref{sec:hg-pg}) directly apply to SCs as well.

\subsection{Cell Complexes}
\label{sec:cell-complexes}

\begin{figure}[h]
    \centering
    \newcommand{\boundaryscaling}{0.48}
    \begin{subfigure}[b]{\boundaryscaling\linewidth}
        \includegraphics[width=\linewidth]{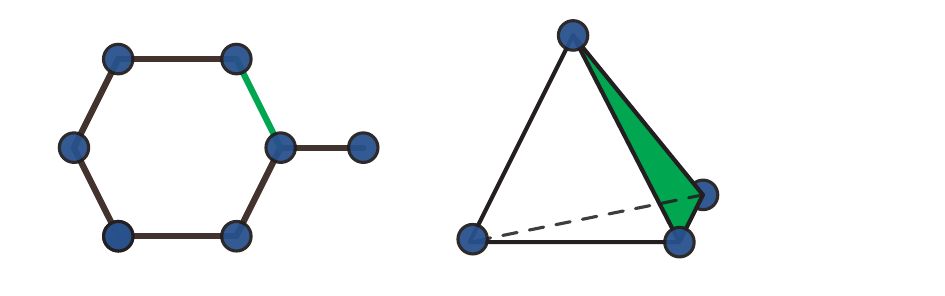}
        \caption{\textbf{Hypergraphs and cells}. Cell complex $\mathcal{C}$ with
            $1$-cell
            $c$ (left),
            simplicial complex $\mathcal{S}$ with
            $2$-cell $s$ (right).}
        \label{subfig:hyperedges_of_focus}
    \end{subfigure}
    ~
    \begin{subfigure}[b]{\boundaryscaling\linewidth}
        \includegraphics[width=\linewidth]{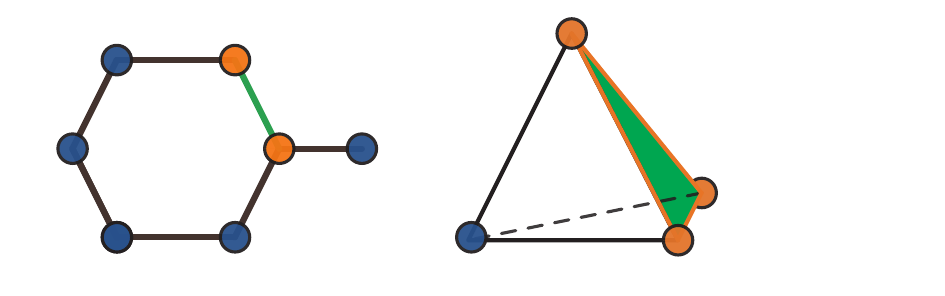}
        \caption{\textbf{Incidences}. The $1$-cell $c$ has incident nodes, $2$-cell
            $s$ has incident nodes and edges.}
        \label{subfig:incidences}
    \end{subfigure}
    \begin{subfigure}[b]{\boundaryscaling\linewidth}
        \includegraphics[width=\linewidth]{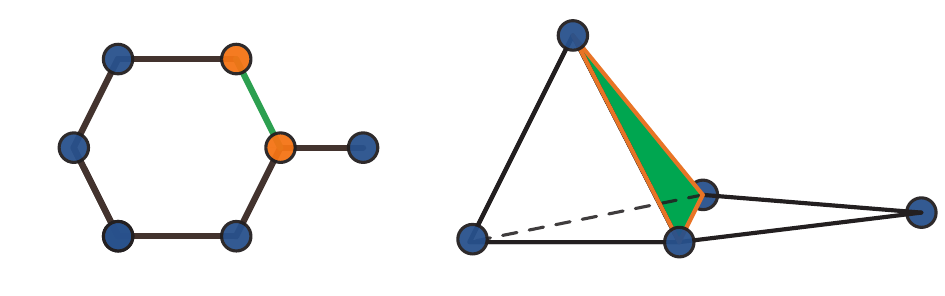}
        \caption{\textbf{Boundary adjacencies}. Incident nodes form boundaries of
            $c$, incident edges the boundaries of $s$.}
        \label{subfig:boundary_adjacency}
    \end{subfigure}
    ~
    \begin{subfigure}[b]{\boundaryscaling\linewidth}
        \includegraphics[width=\linewidth]{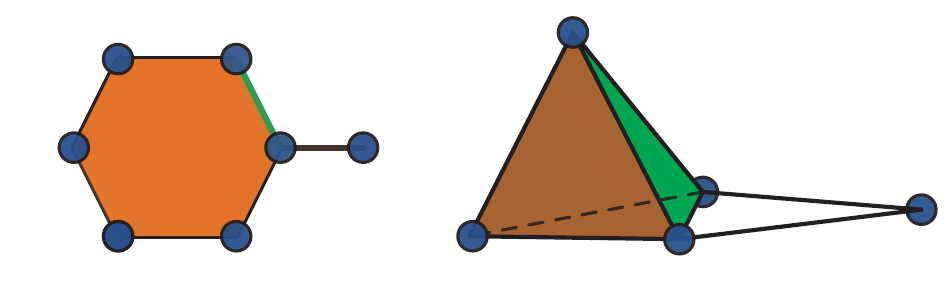}
        \caption{\textbf{Co-boundary adjacencies}. The $1$-cell $c$ has a
        cycle, the $2$-cell $s$ a $3$-dimensional \simplex as co-boundary.}
        \label{subfig:co-boundary_adjacency}
    \end{subfigure}
    \begin{subfigure}[b]{\boundaryscaling\linewidth}
        \includegraphics[width=\linewidth]{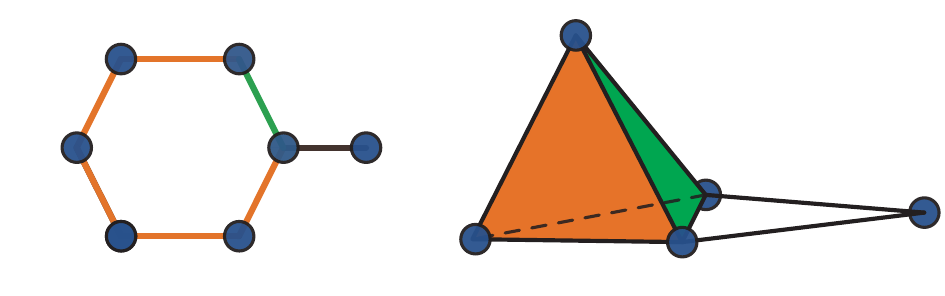}
        \caption{\textbf{Upper adjacencies}. For $c$, these are the edges with the
        cycle as co-boundary, for $s$ the $2$-cells with the $3$D \simplex as co-boundary.}
        \label{subfig:upper_adjacency}
    \end{subfigure}
    ~
    \begin{subfigure}[b]{\boundaryscaling\linewidth}
        \includegraphics[width=\linewidth]{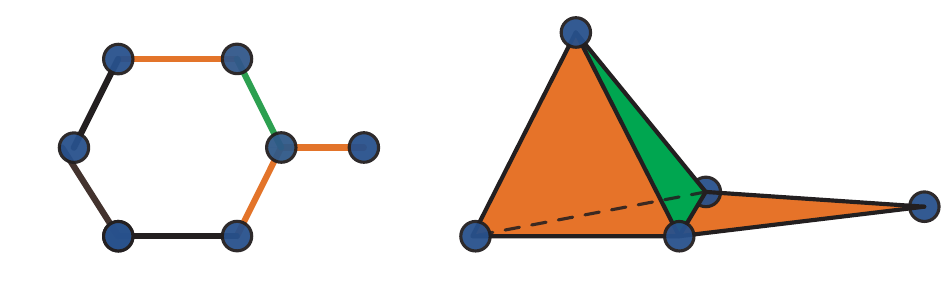}
        \caption{\textbf{Lower adjacencies}. For $c$, these are edges which share
        nodes as boundaries, for $s$, $2$-cells which share edges with $s$ as
        boundaries.}
        \label{subfig:lower_adjacency}
    \end{subfigure}
    \caption{Adjacency notions related to boundary-adjacency in two hypergraphs. As
    higher-order graphs we consider a cell complex $\mathcal{C}$ with nodes as
        $0$-cells, edges as $1$-cells and a $2$-dimensional hyperedge corresponding to the
        cycle, and a $3$-dimensional simplicial complex $\mathcal{S}$.
        Subfigure~\ref{subfig:hyperedges_of_focus} highlights a $1$-cell $c \in
        \mathcal{C}$, and a $2$-cell $s\in \mathcal{S}$, respectively, in green.
        Subfigures~\ref{subfig:incidences}-\ref{subfig:upper_adjacency} display the
        adjacencies of $c$ and $s$ in orange (in
        Subfigure~\ref{subfig:co-boundary_adjacency} the co-boundary of $s$ is
        three-dimensional therefore coloured brown to distinguish it from
        two-dimension faces.)}
    \label{fig:hypergraph_adjacencies}
\end{figure}

\begin{figure}[h]
\vspace{-1.5em}
    \centering
    \includegraphics[width=0.6\linewidth]{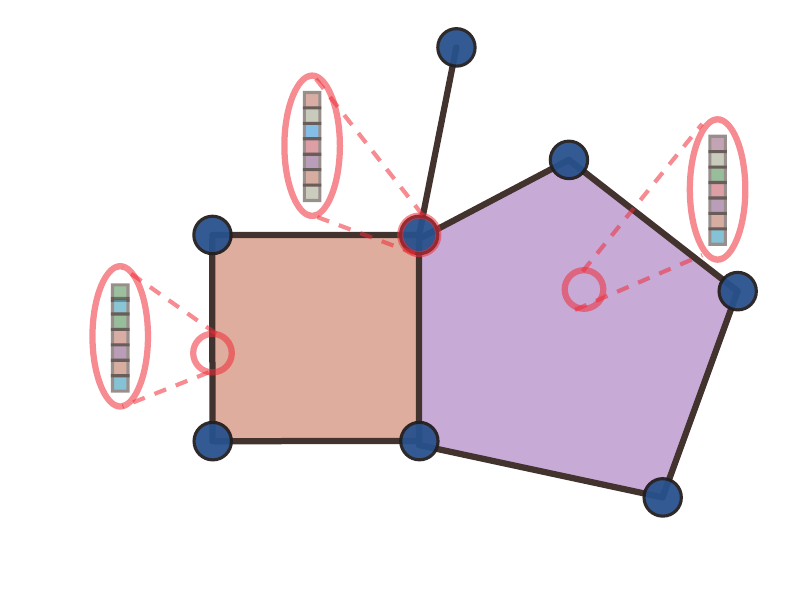}
    \caption{Example cell complex $\mathcal{H}$ with features. Nodes correspond to $0$-cells, edges to $1$-cells and induced cycles to $2$-cells. The vertical arrays visualise the features of the highlighted cells.}
    \label{fig:cell_complex_example}
\end{figure}

The combinatorical constraints of SCs, as described in the previous section, may result in a very large number of cells in order to represent different HO structures. Consider, for example, the molecules portrayed in Figures~\ref{fig:cell_complex_example} and \ref{fig:hypergraph_adjacencies}. In order to capture the full effect of the molecular rings (cycles) we would need to go well beyond the commonly used 2-simplices. Furthermore, the strict requirement that subsets of hyperedges necessarily form a hyperedge as well can be difficult to satisfy in different complex systems. For example, some pairs of substances only react in the presence of a
catalyst. This relationship can be captured by a hypergraph with a hyperedge connecting
three substances but lacking edges between the pairs which do not interact.
Similarly, in drug treatments, certain multi-drug interactions may only appear
in the presence of more than two drugs~\cite{tekin_measuring_2017}. 

The introduction of cell complexes (CCs) came to address the above limitations by creating a hierarchy which is constructed by attaching the boundaries of $n$-dimensional spheres to certain $(n-1)$-cells in the complex \cite{bodnar_weisfeiler_2022}. This generalizes and removes the strict dependency of the hierarchy with the number of vertices as is the case in SCs. Specifically, we can now construct cell complexes by using vertices (0-cells), edges (1-cells), and surfaces (2-cells), which can already cover the most common applications. The formal definition of regular CCs is as follows:

\begin{definition}[\cite{bodnar_weisfeiler_2022}]\label{def:reg_cell_complex}
A real \emph{regular CC} is a higher-dimensional real subset $X\subset \RE^d$ with a finite partition
$P=\{ X_\sigma \}_{\sigma \in P_X}$ of $X$ into so-called \emph{cells} $X_\sigma$ of $X$, such that 
\small
\begin{enumerate}[leftmargin=1.0em]
\item (\textbf{Boundary-subset condition}:) For any pair of cells $X_\sigma, X_\tau$, we have $X_\tau \cap
\overline{X_\sigma} \neq \emptyset\ \text{iff}\ X_\tau \subseteq
\overline{X_\sigma}$. This condition enforces a poset structure of the subspaces of a
space, i.e., $\tau \leq \sigma\ \text{iff}\ X_\tau \subseteq
\overline{X_\sigma}$.
\item (\textbf{Homeomorphism condition}) Every cell is homeomorphic to $\mathbb{R}^n$ for some $n$.
\item (\textbf{Restriction condition}): For every $\sigma \in P_X$ there is a homeomorphism $\phi$ of a closed
ball in $\mathbb{R}^{n_\sigma}$ to $\overline{X_\sigma}$ such that the
restriction of $\phi$ to the interior of this ball is a homeomorphism onto
$X_\sigma$,
\end{enumerate}
\normalsize
\end{definition}

\noindent
where $\overline{X_\sigma}$ is the \emph{closure} of a cell $X_\sigma$, i.e., all points in $X_\sigma$ together with all limit points of $X_\sigma$.
A \emph{homeomorphism} from a given partition $p\in {P}$ to a given domain $\RE^{d_p}$ is a continuous map that transforms $p$ to $\RE^{d_p}$ (and its inverse is also continuous). 
For full details, see Mendelson's Introduction to
Topology~\cite{mendelson_introduction_2012}.

The cells in a CC form a multipartite structure where $n$-cells are connected to $(n+1)$-cells, which are in turn connected to $(n+2)$-cells, and so on, see Fig.~1 in~\cite{Hajij2023CombinatorialCB} for a depiction of this multipartite structure of a CC. This leads to a multipartite structure where vertices are only connected to $1$-cells directly. This is in contrast to the SC where vertices are fully connected to all other cells.
Different forms of adjacency in CCs, HGs, and Scs, are shown in Figure~\ref{fig:hypergraph_adjacencies}.

As in SCs, a CC inherits the notion of isomorphism as well as the generalized
forms of adjacency between \simplices, as specified for HGs.

%
A cell's dimension is the dimension of the space it is homeomorphic to. We call
cells of dimension $p$ the $p$-\emph{cells} of the complex. The
\emph{dimension} of a CC is the maximum dimension of its constituents. Note
that the dimension of a cell $c=\{v_{i_0},\dots,v_{i_p}\}$ need not be related
to the number of nodes it comprises. For example, $2$-cells associated with
cycles of length $k\geqslant 4$ have dimension $2$ but comprise $k$ nodes.
Using the nomenclature from before, $p$-\simplices ~need not be $p$-cells.
\emph{As a matter of fact, an SC is a CC for which every $p$-\simplex ~has
dimension $p$ for $p\in \ZE_{\geqslant 0}$.}

In contrast to SCs, CCs can encode $p$-dimensional substructures containing
more than $p+1$ vertices. Thus, they are more flexible and better suited for
learning on certain datasets, for example, molecular
datasets~\cite{bodnar_weisfeiler_2022}. Similarly to SCs, representations for
CCs are learned through MP along boundary adjacency (see
Definition~\ref{def:hypergraph:incidence_boundary_adjacency}) and related
adjacency types~\cite{hajij_cell_2021}.

Consider the HG $\mathcal{H}$ portrayed in
Figure~\ref{fig:cell_complex_example} with nodes marked in blue, edges in black
and higher-order hyperedges in orange and purple, respectively.
%
%
Note that the
hyperedges are associated with cycles in the underlying graph. If we consider
the HG as a topological space, we observe that a \simplex ~touches another if and
only if there is an incidence relation between them. Moreover, each \simplex ~$c$
is homeomorphic to $\RE^{d_c}$ for some $d_c$ (i.e., there is a continuous map
that transforms $c$ to $\RE^{d_c}$, see~\cite{mendelson_introduction_2012}).
These are two of the defining properties of CCs.

\subsubsection{Lifting PGs to CCs}

Next, we relate PGs to CCs via lifting transformations. For this, we extend the
notion of HG to multi-HG $\mathcal{H}$ by allowing some hyperedges to occur
multiple times (possibly with different features). We define the
$p$-\emph{skeleton} of a CC $\mathcal{H}$ to be all cells in $\mathcal{H}$ with
dimension at most $p$, denoted $\mathcal{H}^{\leqslant p}$. We call a lifting
map $f$ from PGs to CCs \emph{1-skeleton-preserving} if for any graph $G$, the
$1$-skeleton of $f(G)$ (identified as a multi-graph composed of nodes
$\mathcal{V}=\{0\text{-cells}\}$ and multiedges $E=\multiset{1\text{-cells}}$) is
isomorphic to $G$. In particular, any lifting that attaches multiple $1$-cells
to a pair of nodes is not 1-skeleton-preserving. While the only
1-skeleton-preserving lifting from PGs to SCs is the CqC lifting~\cite[page
5]{bodnar_weisfeiler_2022}, there are multiple 1-skeleton-preserving liftings
of PGs to CCs. We will focus on liftings that map cycles in a graph to
$2$-cells. We distinguish between two types of cycles.

\begin{definition}\label{def:cycle}
    A \emph{cycle} in a graph $G=(\mathcal{V},\mathcal{E})$ is a collection of nodes
    $v_{i_0},\dots,v_{i_p}$ that are connected by a walk $v_{i_0}\to v_{i_1}\to
    \dots\to v_{i_p}\to v_{i_0}$. An \emph{induced cycle} is a cycle that does
    not contain any proper sub-cycles.
\end{definition}

When lifting cycles to $2$-cells, we will specify the maximum length of induced
cycles $k_{\text{ind-cycle}}$ and regular cycles $k_{\text{cycle}}$ to consider.
Since any induced cycle is also a cycle, we will have $k_{\text{cycle}}\leqslant
k_{\text{ind-cyc}}$.

\begin{definition}\label{def:cell_lifting}
    Consider a graph $G=(\mathcal{V},\mathcal{E})$, $k_{\text{cl}}$, $k_{\text{ind-cycle}}$,
    $k_{\text{cycle}}\in \nat$. We define a $ (k_{\text{cl}},
    k_{\text{ind-cycle}},k_{\text{cycle}})$-cell lifting by attaching a
    $p$-cell to every clique of size $p+1\leqslant k_{\text{cl}}$, a $2$-cell
    to any induced cycle of length at most $k_{\text{ind-cycle}}$ and a
    $2$-cell to any non-induced cycle of length at most $k_{\text{cycle}}$.
\end{definition}

\begin{proposition}\label{prop:cellular_lifting}
    The cell lifting defined above is 1-skeleton-preserving and
    isomorphism-preserving.
\end{proposition}

%

\subsection{\hspace{-0.5em}\mbox{Graphs with Node-Tuple Collections}} \label{sec:graph_with_node_tuples}

Hypergraphs and their specialized variants, simplicial and cell complexes, are
restricted, because hyperedges are \emph{subsets} of nodes and are thus
\emph{fundamentally unordered}. Moreover, they cannot contain one node
multiple times, i.e., they are not multisets. However, as we will discuss in
detail in Section~\ref{chapter:HoGRL_expressiveness}, harnessing order of nodes
or multiset properties can enable distinguishing non-isomorphic graphs that
would otherwise be indistinguishable by WL schemes.

We now formally introduce \emph{graphs with
node-tuple-collections} (\emph{NT-Col-graphs})
that address the above issues by using ordered node-tuples instead of
hyperedges to encode higher-order structures.
While tuples of nodes have already been used in HOGNNs~\cite{morris_weisfeiler_2021}, {we are the first to formally define NT-col-graphs as a separate data model, as a part of our blueprint, in order to facilitate developing HOGNNs.}

\begin{definition}\label{def:node_tuple_collection}
    A \emph{node-tuple-collection} $\mathcal{C}=(\mathcal{V},\mathcal{E},\mathfrak{T},\mathbf{x})$ is a
    graph $G=(\mathcal{V},\mathcal{E})$ together with a collection of node tuples $\mathfrak{T}\sub
    \cup_{k=2}^{k_{\max}}\mathcal{V}^k$, possibly endowed with features $\mathbf{x}\colon \mathcal{V} \cup
    \mathcal{E} \cup \mathfrak{T} \to \RE^d$.
\end{definition}

\begin{figure}[h!]
    \centering
    \includegraphics[width=0.6\linewidth]{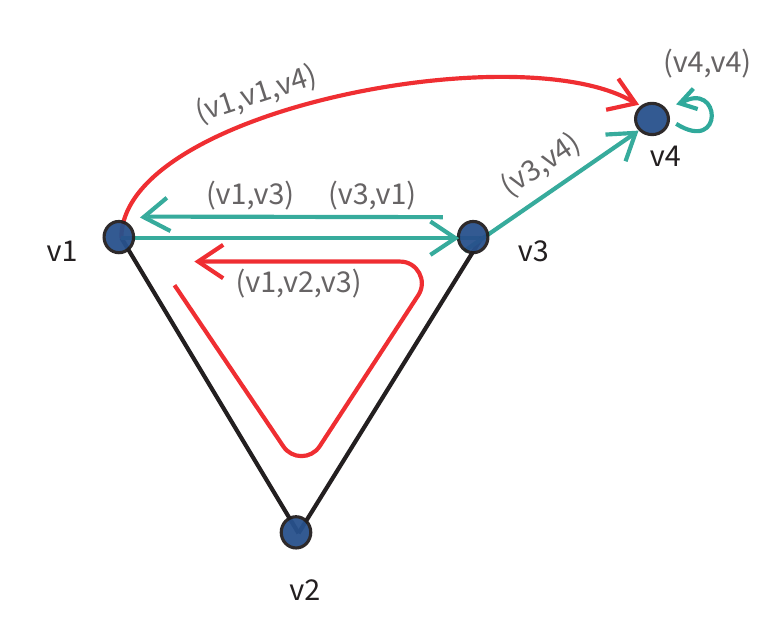}
    \caption{Example of a node-tuple-collection $\mathcal{C}$. Nodes in the underlying graph are depicted in dark blue and node-tuples as directed edges or paths.}
    \label{fig:node_tuple_example}
\end{figure}

We denote one $k$-tuple of nodes as
$\textbf{v}=(v_{i_1},\dots,v_{i_k}) \in \mathcal{V}^k$.
Note that one may use node-tuples of length two, which possibly do \emph{not}
appear as edges in the underlying plain graph $G$ (in particular, if the graph
is undirected). 
%
%
An example node-tuple-collection is depicted in
Figure~\ref{fig:node_tuple_example}.

One can extend the notion of adjacency in different ways to accommodate node-tuple collections.
For example, Morris et al.~\cite{morris_weisfeiler_2021} in their $k$-GNN architecture introduce \textbf{down adjacency} between node-tuples that is similar to lower-adjacencies:

\begin{definition}[Down adjacency]\label{def:node_tuple:down_down}
    For an NT-col-graph $\mathcal{C}=(G,\mathfrak{T})$, a \emph{down-adjacency} is defined as
    \begin{align*}
        \mathcal{N}_{\Downarrow}(\textbf{v}) &= \{\textbf{w}\in
        \mathcal{V}^k:\exists! j\in \{1,\dots,k\}: w_j\neq v_j\}.
    \end{align*}
\end{definition}

However, since the number of down-adjacencies scales exponentially in $k$, the authors also propose an MP
architecture in which only messages from \textbf{local down adjacencies} are processed. Formally, we have

\begin{definition}[Local down adjacency]\label{def:node_tuple:down_local_down}
    For an NT-col-graph $\mathcal{C}=(G,\mathfrak{T})$, we define
    a \emph{local down-adjacency} to be
    \begin{align*}
        \mathcal{N}_{\text{loc},\Downarrow}(\textbf{v}) &= \{\textbf{u}\in
        \mathcal{V}^k: \exists! j: v_j\neq u_j, \{v_j,
        u_j\}\in
        E \}.
    \end{align*}
\end{definition}

\iftr
\subsubsection{Lifting Graphs to NT-Col-Graphs}

\maciej{This is badly defined, we would need reasonable lifting...}

%
Before we define a lifting from a graph to an NT-col-graph, we introduce an
encoding for node-tuples which captures their relation in the underlying graph
$G$.

\begin{definition}\label{def:isomorphism_type}
    For a graph $G=(\mathcal{V},\mathcal{E},\mathbf{x})$, the \emph{isomorphism type} of a
    node-tuple $\textbf{v}\in \mathcal{V}^k$, denoted $\textsc{Iso}(\textbf{v},G)
    \in\RE^{2\binom{k}{2}+k\cdot d}$ is defined as follows. The first
    $\binom{k}{2}$ entries capture equality
    relations $\mathbbm{1}_{v_i=v_j}$, the next $\binom{k}{2}$ entries capture adjacency
    information $\mathbbm{1}_{\{v_{i},v_j\}\in E}$ and the final $k\cdot d$ entries
    capture the features $(\mathbf{x}_{v_i})_{v_i\in \textbf{v}}$ associated
    with the nodes in the tuple.
    \maciej{This def in unclear, i.e., it's not clear how these entries actually
    look like - please define explicitly} \florian{Sure, I'll think about how we can put this consisely. Just to clarify:
for the first $\binom{k}{2}$ entries of a node-tuple's isomorphism type, we
consider the corresponding pairs of nodes appearing in the node tuple and put a $1$ in the entry if they are identical
and $0$ otherwise. For the next $\binom{k}{2}$ entries,
we put a $1$ if the corresponding pair of nodes are connected in the underlying graphs and $0$ otherwise.
    I assume the feature entries are clear?}
\maciej{Is this formally defined in some ref?} \florian{Yes, the isomorphism type is an established concept, e.g.,
    you should find it in Morris' work.}
\end{definition}

\begin{definition}\label{def:isomorphism_type_lifting}
    Given a graph $G=(\mathcal{V},E,\mathbf{x})$ and $k_{\max}\in \nat$,
    we define the \emph{isomorphism-type (iso-type) lifting} to be an NT-col-graph 
    $\mathcal{C}=(\mathcal{V},E,\mathfrak{T},\textbf{z})$ comprised of all node-tuples
    $\mathfrak{T}=\cup_{k=2}^{k_{\max}}\mathcal{V}^k$ of length at most $k_{\max}$, and
features $\textbf{z}(\{G\}):=\mathbf{x}(\{G\})$, $\textbf{z}(c):=\mathbf{x}(c)$
for $c\in \mathcal{V}\cup E$ and $\textbf{z}(\textbf{v})=\textsc{IsoT}(\textbf{v},G)
    \in\RE^{2\binom{k}{2}+k\cdot d}$ for all node-tuples
$\textbf{v}\in \mathfrak{T}$.
\maciej{Is this formally defined in some ref?} \florian{Only the isomorphism type, not the lifting as we define it above.}
\end{definition}

\noindent
Next, we define isomorphism between NT-col-graphs and
show that an iso-type lifting preserves isomorphisms.
\maciej{Is this formally defined in some ref?} \florian{No, we introduce this in order to later establish that our
notion of isomorphism is preserved by the lifting.}

\begin{definition}\label{def:isomorphism_node_tuple_collections}
    Two NT-col-graphs $\mathcal{C}_1=(\mathcal{V}_1,E_1,\mathfrak{T}_1,
    \mathbf{x}_1),
    \mathcal{C}_2=(\mathcal{V}_2, E_2,\mathfrak{T}_2,\mathbf{x}_2)$ are \emph{isomorphic} if
    \begin{enumerate}
        \item the underlying
        graphs are
        isomorphic,
        \item the node relabelling function
        $\varphi\colon
        \mathcal{V}_1\to \mathcal{V}_2$ from the graph isomorphism
        satisfies $(v_{i_{1}},\dots,v_{i_k})\in \mathfrak{T}_1$ if and only if  $
        (\varphi(v_{i_1}),\dots,\varphi(v_{i_k}))\in \mathfrak{T}_2$ for any
        node-tuple $(v_{i_1},\dots,v_{i_k})\in \mathcal{V}^{k}$ for any $k\in \nat$,
        \item \label{def:item_node_tuple_iso} features are preserved
        $\mathbf{x}_2(\varphi({c}))=\mathbf{x}_1({c})$ for any $c\in \mathcal{V}\cup E$, $\mathbf{x}_2(\varphi(\textbf{v}))=\mathbf{x}_1(\textbf{v})$ for any $\textbf{v}\in \mathcal{C}_1$, and $\mathbf{x}_2(\{\mathcal{C_2}\})=\mathbf{x}_1(\{\mathcal{C}_1\})$, where the node-relabelling function is defined component-wise for edges and node-tuples.
    \end{enumerate}
\end{definition}

\begin{proposition}\label{prop:isomorphism_type_lifting}
    The isomorphism-type lifting from
    Definition~\ref{def:isomorphism_type_lifting} preserves isomorphisms.
\end{proposition}

\subsubsection{Lifting Hypergraphs to NT-Col-Graphs}

\maciej{Is this formally defined in some ref?} \florian{No. And I'm not sure it's of interest for the larger community. What do you think?
I added it to the thesis to give a more complete picture of the structure relations.}
Hypergraphs can also be encoded as 
NT-col-graphs. 
The encoding is similar to mapping simple graphs to directed
graphs, which adds two directed edges for every edge occurring in the input
graph. This transformation formally establishes another step in the taxonomy of
HoGRL.

\begin{proposition}\label{prop:hypergraph_to_node_tuple_collection}
    Given an HG $\mathcal{H}=(\mathcal{V},\mathcal{E},\mathbf{x})$, we choose an
    arbitrary ordering of the nodes $\mathcal{V}=(v_1,\dots,v_n)$ and define a lifted
NT-col-graph $(G,\mathfrak{T},\textbf{z})$: the underlying
graph is given by $G=(\mathcal{V},E)$, where the edge set is all hyperedges made up
of two nodes $E=\{e \in \mathcal{E}: |e|=2\}$, node-tuples are
        $\mathfrak{T}=\{(v_{i_1},\dots,v_{i_p}): \{v_{i_1},\dots,v_{i_p}\} \in \mathcal{E} \land
        i_{j-1}\leqslant i_j\}$,
and features $\mathbf{z}(v)=\mathbf{x}(v), v\in \mathcal{V}$,
$\mathbf{z}(e)=\mathbf{x}(e), e\in E$,
$\mathbf{z}((v_{i_1},\dots,v_{i_p}))=\mathbf{x} (\{v_{i_1},\dots,v_{i_p}\})$
for $\{v_{i_1},\dots,v_{i_p}\}\in \mathcal{E}$ and
  $\mathbf{z}(\{f(\mathcal{H})\})=\mathbf{x}(\{\mathcal{H}\})$. This lifting 
preserves isomorphism.
\end{proposition}

\fi

\subsection{Graphs with Subgraph[-Tuple] Collections}\label{sec:graph_with_subgraphs}

Next, we formally introduce a GDM implicitly used by several neural architectures where the
``first-class citizens'' are arbitrary \textit{subgraphs}.
In these schemes, plain graphs are enriched with a collection of subgraphs
selected with certain criteria specific to a given architecture. These subgraphs are 
not necessarily induced, and may even introduce new ``virtual'' edges
not present in the original graph dataset.
Such architectures capture substructures of an input graph to achieve expressivity
beyond the 1-\acrshort{wltest}~\cite{xu2018powerful}. Some achieve this by
learning subgraph representations~\cite{rongDropEdgeDeepGraph2020, pappDropGNNRandomDropouts2021,
thiede_autobahn_2022, kim_efficient_2022, bevilacqua_equivariant_2022,
zhao_stars_2022, zhang_nested_2021, sandfelder_ego-gnns_2021}, which they
subsequently transform to graph representations.  Others predict the properties
of subgraphs themselves. For example, in medical diagnostics, given a
collection of phenotypes from a rare disease database, the task is to predict
the category of disease that best fits this
phenotype~\cite{alsentzer_subgraph_2020}.

Inspired by these different use cases, we introduce a GDM called the \emph{graph with a subgraph-collection} (SCol-graph).

\begin{definition}\label{def:subgraph_collection} A \emph{graph with a
subgraph-collection (SCol-graph)} is a tuple $\mathcal{C}=(\mathcal{V},\mathcal{E},\mathcal{S},\mathbf{x})$
comprising a simple graph $G=(\mathcal{V},\mathcal{E})$, a collection of
    (possibly non-induced) subgraphs $\mathcal{S}$ with $\mathcal{V}_H\sub \mathcal{V},\mathcal{E}_H\sub
    \mathcal{V}_H\times \mathcal{V}_H$ for every $H=(\mathcal{V}_H,\mathcal{E}_H)\in\mathcal{S}$, and features
    $\mathbf{x}\colon \mathcal{V}\cup \mathcal{E} \cup\mathcal{S} \to \RE^d$.
\end{definition}

\begin{figure}[h!]
    \centering
    \includegraphics[width=0.5\linewidth]{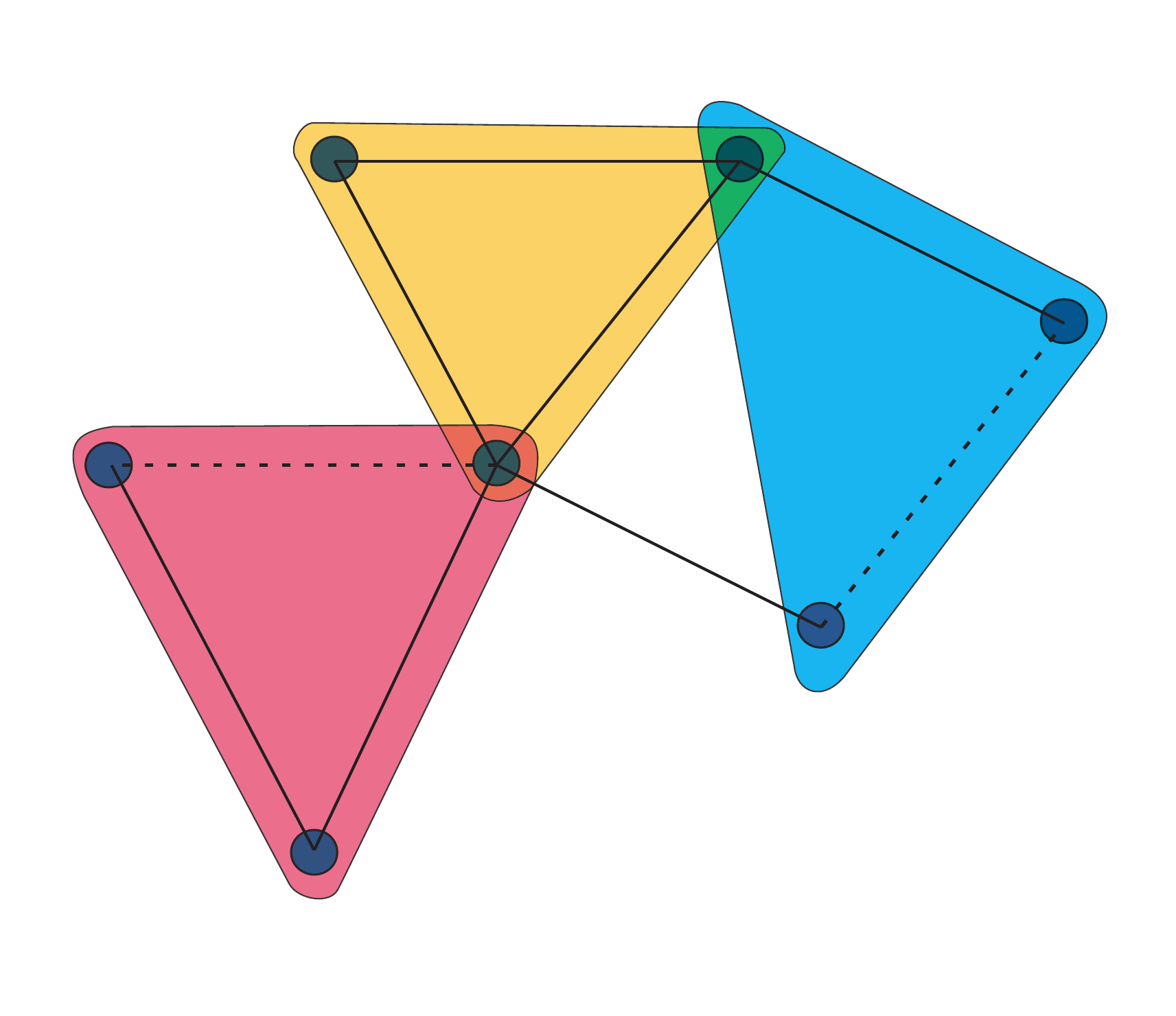}
    \caption{Example of a subgraph-collection $\mathcal{C}$ with subgraphs in red, yellow and blue. The edges in the underlying graph are displayed by continuous lines and the virtual edges which appear in subgraphs but not in the underlying graph, by dotted lines.}
    \label{fig:subgraph_example}
\end{figure}

To the best of our knowledge, notions of incidence and boundary-adjacency as in Def.~\ref{def:hypergraph:incidence_boundary_adjacency} have not been defined for SCol-graphs. Instead, the adjacency is usually very specific to a given scheme, we will refer collectively to it as the \textbf{subgraph adjacency}.
Some architectures associate subgraphs with single nodes, subsequently defining subgraph adjacency via adjacency of the
corresponding nodes~\cite{sandfelder_ego-gnns_2021, zhang_nested_2021}.
Another approach is to measure the number of nodes or edges that subgraphs
share and define adjacencies between them based on these
overlaps~\cite{thiede_autobahn_2022}. Others use even more complex notions of
adjacency to construct MP channels between subgraphs or their connected
components~\cite{alsentzer_subgraph_2020}.
%
%
Figure~\ref{fig:subgraph_example} shows an example of a subgraph-collection.

Similarly to NT-Col-graphs, one can also impose an ordering of subgraphs in SCol-graphs. This has been proposed in a recent work by Qian et al.~\cite{qian2022ordered}. We refer to the resulting underlying GDM as \textbf{graph with a subgraph-tuple collection (ST-Col-graph)}. Analogously to NT-Col-Graphs, one can then extend the notion of down adjacency into the realm of ST-Col-graphs.

\subsection{Motif-Graphs \& SCnt-Graphs}
\label{sec:graph_motifs}

In PGs, links are edges between \textit{two} vertices. In HGs and their specialized variants (SCs, CCs), HO is introduced by making edges being able to link \textit{more than two vertices}, as well as extending adjacency notions to links.
In NT-Col-graphs, SCol-graphs, and ST-Col-graphs, one introduces HO by distinguishing collections of substructures \textit{in addition to harnessing edges between vertices}.
Now, we describe another way of introducing HO into GDMs: \textit{defining adjacencies using substructures consisting of dyadic interactions}. These substructures are called \textit{motifs}~\cite{besta2022motif, benson2016higher}, for example triangles~\cite{strausz2022asynchronous}. The key formal notions here are a {motif} and the underlying concepts of a \emph{graphlet} and \emph{orbit}.

\begin{definition}\label{def:orbit}
    Let $G=(\mathcal{V},\mathcal{E})$ be a graph. A \emph{graphlet} in $G$ is a subgraph $H=
    (\mathcal{V}_H,\mathcal{E}_H)$ such that $\mathcal{V}_H\subset \mathcal{V}, \mathcal{E}_H \subset \{1 .. \binom{\mathcal{V}_H}{2} \} \cap \mathcal{E}$. A
    graphlet automorphism is a permutation of the nodes in $\mathcal{V}_H$ that preserves
    edge relations. A set of nodes $\mathcal{V}_H$ in a graphlet $H$ define an
    \emph{orbit} if the action of the automorphism group is transitive, that
    is, for any $u,v\in \mathcal{V}_H$ there exists an automorphism $\phi$ on $\mathcal{V}_H$ such
    that $\phi(u)=v$.  Finally, a \emph{motif} is identified with the
    isomorphism equivalence class of a graphlet or an orbit. 
\end{definition}

We will refer to a GDM that is based on ``motif-driven'' adjacencies as \textit{Motif-Graphs}. We distinguish two classes of such GDMs.
%

In one class, one uses motifs to define ``motif induced'' neighborhoods, in which - intuitively - the neighbors of a given vertex are determined based on whether they share an edge that is included in a given selected motif. For example, in HONE~\cite{rossi_hone_2018}, one first chooses a collection of motifs $\mathcal{S}$ and builds a new edge-weighted graph $W_M$ for every motif $M\in \mathcal{S}$. Given an input graph $G$, the weight of an edge $e\in W_M$ is the number of subgraphs of $G$ which contain $e$ and which are isomorphic to $M$. The motif-weighted graphs $W_M$ for $M\in \mathcal{S}$ can then be used to learn node or graph representations~\cite{rossi_hone_2018, lee2019graph}.
The same forms of motif-based adjacency are used to design MotifNet, a GNN model that seamlessly works with directed graphs~\cite{monti2018motifnet}.
%
%
This simple notion of motif-based adjacency can also have more complex forms. For example, HONE also proposes a ``higher-order form'' of this motif adjacency: a more abstract concept where adjacency is not just about direct node-to-node connections but involves the role of nodes across different instances of the same motif or across different motifs. For example, two nodes might be considered adjacent in this higher-order sense if they frequently participate in similar motifs, even if they are not directly connected.

In another class of Motif-Graphs, the key idea is to extend the features of vertices with information on \emph{how many selected motifs each vertex belongs to}.
This idea has been explored for triangles and for more general subgraphs, and it has been shown to improve a GRL method's ability to distinguish non-isomorphic graphs~\cite{bouritsas_improving_2021}.


\subsection{Nested GDMs}
\label{sec:nested-gdms}

Finally, some HOGDMs incorporate \textit{nesting}. In such a GDM, a distinguished part of a graph (or a hypergraph) is modeled as a ``higher-order vertex'' with some inner structure that forms a graph itself. 
This GDM has been implicitly used by different GNN architectures, including Graph of Graphs Neural Network~\cite{wang_gognn_2020}, Graph-in-Graph Network~\cite{liu2022graph}, hierarchical graphs~\cite{liSemiSupervisedGraphClassification2019}, and Neural Message Passing with Recursive Hypergraphs~\cite{yadati2020neural}.
The details of such a formulation (e.g., whether or not to connect such higher-order vertices explicitly to plain vertices) depends on a specific architecture.
A motivating example use case for such a GDM is a network of interacting proteins: while an individual protein has a graph structure, one can also model interactions between different proteins as a graph.

\section{Higher-Order GNN Architectures}
\label{chap:from_gdm_to_repL}

We now analyze HOGNN architectures that harness different HOGDMs.
%
%
Most such architectures have a structure similar to the MP framework (see Sec.~\ref{sec:gnns-summary}), except that, instead of only updating the features of nodes, they also update the features of higher-order structures. For example, a subgraph or a hyperdge can have their own feature vectors.
%

We first discuss HOGNN architectures that -- as their underlying GDM -- harness HGs (Section~\ref{sec:hognns-hgs}), SCs (Section~\ref{subsec:hypergraph:simplicial_complex}), CCs (Section~\ref{subsec:hypergraph:cell_complex}), NT-Col-Graphs (Section~\ref{sec:hognns-ntcol}), SCol-Graphs and ST-Col-Graphs (Section~\ref{sec:subgraph_repL}), and Motif-Graphs as well as SCnt-Graphs (Section~\ref{sec:hognns-motifs}). We then discuss other orthogonal aspects of HOGNNs such as MP flavors (Section~\ref{sec:flavors}) or nesting (Section~\ref{sec:hognns-nested}).
%

For more insightful discussions, in the following section we also present in more detail representative HOGNN architectures from different classes of schemes.

Table~\ref{tab:overview_higher_order_methods} compares selected representative HOGNN models.
We illustrate how these models are expressed in their respective MP blueprints. We select one model from each MP flavor (convolutional, attentional, general) and from a corresponding GDM. 

%

\begin{table*}[hbtp]
    \scriptsize
    \footnotesize
    \renewcommand{\arraystretch}{2.1}
    \centering
    \begin{tabular}{clccccc}
    \toprule
        {\textbf{HoGDM}}                         & \textbf{Wiring}                   & \textbf{Flavour}                  & \textbf{Obj}                                     & \textbf{Example update equations}  & \textbf{Eq.\ ref} & \textbf{Examples}  \\ \cline{1-7}
        {\multirow{3}{*}{hypergraph}}   & \multirow{3}{*}{IMP}  & \multirow{3}{*}{gen}  &                 $v\in \mathcal{V}$                                & $\textbf{x}^{\text{new}}_v = \phi_\mathcal{V}\left(            \textbf{x}_v,            \bigoplus_{f\in            \mathcal{E}: v\sub f}{\textbf{m}_f}            \right)$.                           &  \multirow{3}{*}{~\cite{heydari_message_2022}}               & \\
        \multicolumn{1}{l}{}                              &                       &                       &                                        & $\textbf{m}_e = \phi_{\mathcal{E}}\left( \textbf{x}_e,            \bigotimes_{u\in \mathcal{V}: u\sub            e}\psi_\mathcal{V}\left(\textbf{x}_{u}\right)\right).$ & &                \\
        \multicolumn{1}{l}{}                              &                       &                       & $e\in \mathcal{E}$                      & $\textbf{x}^{\text{new}}_e =            \phi_{\mathcal{E}}\left( \textbf{x}_e,            \bigodot_{u\in \mathcal{V}: u\sub            e} \psi_\mathcal{V}\left(\textbf{x}_{u}\right)\right).$       &     &                \\ \cline{1-7}
        \multirow{3}{*}{hypergraph}         & \multirow{3}{*}{IMP}  & \multirow{3}{*}{att}  &           $v\in \mathcal{V}$                                & $\textbf{x}^{\text{new}}_v = \tfrac{1}{\sqrt{\deg(u)}}            \tfrac{1}{|f\in \mathcal{E}: v\sub f|}            \sum_{f\in            \mathcal{E}: v\sub f}a(\textbf{x}_v,\textbf{x}_f)\textbf{m}_f\Theta$.                                                                                                                                                 &     \multirow{3}{*}{~\cite{bai_hypergraph_2021}}      &       \\ \cline{5-5}
        &                       &                       &                       & $\textbf{m}_e = \omega_{e}\tfrac{1}{|u\in \mathcal{V}:            u\sub            e|}\sum_{u\in \mathcal{V}: u\sub            e}\tfrac{1}{\sqrt{\deg(u)}}a(\textbf{x}_e,\textbf{x}_v)\textbf{x}_{u}            .$                                                                                                      &               &  \\ \cline{5-5}
        &                       &                       &                                         & $a(\textbf{x}_v,\textbf{x}_f) = \tfrac{\exp(\sigma            (\operatorname{sim}(\textbf{x}_v\Theta,\textbf{x}_f\Theta)))            }{\sum_{e\in \mathcal{E}: v\sub e}\exp(\sigma            (\operatorname{sim}(\textbf{x}_v\Theta,\textbf{x}_e\Theta)))}$.    &    &               \\ \cline{1-7}
        \multirow{2}{*}{hypergraph}         & \multirow{2}{*}{IMP}  & \multirow{2}{*}{conv} &  $v\in \mathcal{V}$                                & $\textbf{x}^{\text{new}}_v = \tfrac{1}{\sqrt{\deg(u)}}            \tfrac{1}{|f\in \mathcal{E}: v\sub f|}\sum_{f\in            \mathcal{E}: v\sub f}\textbf{m}_f \Theta$.   &  \multirow{2}{*}{~\cite{feng_hypergraph_2019}}     &   \multirow{2}{*}{~\cite{bai_hypergraph_2021, heydari_message_2022}}        \\
        &                       &                       &                                                                                   & $\textbf{m}_e = \omega_{e}\tfrac{1}{|u\in \mathcal{V}: u\sub            e|}\sum_{u\in \mathcal{V}: u\sub            e}\tfrac{1}{\sqrt{\deg(u)}}\textbf{x}_{u}.$                                                                                                    &      &          \\ \cline{1-7}
        {\multirow{3}{*}{cell complex}} & \multirow{3}{*}{BAMP} & \multirow{3}{*}{gen}  &                                 $c\in \mathcal{H}$                      & $\textbf{x}_c^{\text{new}}            = \phi\left( \textbf{x}_c, m^{\mathcal{B}}_c,            m^{\mathcal{N}_{\uparrow}}_c\right)$                                                                                                                                                              &        \multirow{3}{*}{~\cite{bodnar_weisfeiler_2022}}                  &     \multirow{3}{*}{~\cite{hajij_cell_2021}}              \\ \cline{5-5}
        \multicolumn{1}{l}{}                              &                       &                       &              &    $m^{\mathcal{B}}_c = \bigoplus_{b\in            \mathcal{B}(c)}\psi_{\mathcal{B}}\left( \textbf{x}_c,\textbf{x}_b            \right)$                                                                                                 &     &                \\ \cline{5-5}
        \multicolumn{1}{l}{}                              &                       &                                                  &                      &        $m^{\mathcal{N}_{\uparrow}}_c = \bigoplus_{d\in            \mathcal{N}_{\uparrow}(c), \delta\in \mathcal{C}(c)\cap \mathcal{C}(d)            }\psi_{{\uparrow}}\left(            \textbf{x}_c,\textbf{x}_d, \textbf{x}_{\delta}\right)$                                                                                                      &                                          &                \\ \cline{1-7}
    {\multirow{4}{*}{cell complex}} & \multirow{4}{*}{BAMP} & \multirow{4}{*}{att}  &                   $e\in \mathcal{H}_1$  & $\textbf{x}_e^{\text{new}} = \alpha_p\left( \tilde{\mathbf{x}}_e \right) \cdot \tilde{\mathbf{x}}_e $ &  \multirow{4}{*}{~\cite{giusti2022cell}}  &                \\ \cline{5-5}
    \multicolumn{1}{l}{}            &                       &                       &                                         & $\tilde{\mathbf{x}}_e = \phi\left( \mathbf{x}_e, m^{\mathcal{N}_{\uparrow}}_c, m^{\mathcal{N}_{\downarrow}}_c\right)$ &       &          \\ \cline{5-5}
    \multicolumn{1}{l}{}            &                       &                       &                                         & $m^{\mathcal{N}_{\uparrow}}_c= \bigoplus_{f\in \mathcal{N}_{\uparrow}} \alpha_{\uparrow}(\mathbf{x}_e, \mathbf{x}_f)\psi_{\uparrow}\left( \mathbf{x}_e \right)$ &      &          \\ \cline{5-5}
    \multicolumn{1}{l}{}            &                       &                       &                                         & $m^{\mathcal{N}_{\downarrow}}_c= \bigoplus_{f\in \mathcal{N}_{\downarrow}} \alpha_{\downarrow}(\mathbf{x}_e, \mathbf{x}_f)\psi_{\downarrow}\left( \mathbf{x}_e \right)$   &      &          \\ \hline
        {\multirow{3}{*}{cell complex}} & \multirow{3}{*}{BAMP} & \multirow{3}{*}{conv}  &                          $c\in \mathcal{H}$                 & $\textbf{x}_c^{\text{new}} = \sigma\left(\sum_{d\in \mathcal{N}_{\uparrow}(c)}\beta_{cd}\mathbf{x}_d \Theta\right)$ &        \multirow{3}{*}{~\cite{hajij_cell_2021}}     &  \multirow{3}{*}{~\cite{bodnar_weisfeiler_2022}}    \\ \cline{5-5}
        \multicolumn{1}{l}{}                              &                       &                       &                                         &  $\beta_{cd} = \mathbbm{1}_{\{c=d\}}+\tfrac{|\mathcal{C}(c)\cap \mathcal{C}(d)|}{\sqrt[]{\deg^{\mathcal{N}_{\uparrow}}(c)\cdot \deg^{\mathcal{N}_{\uparrow}}(d)}}$                                                                                       &         &       \\ \cline{5-5}
        \multicolumn{1}{l}{}                              &                       &                       &                     &  $\deg^{\mathcal{N}_{\uparrow}}(c) = \sum_{d\in \mathcal{H}}|\mathcal{C}(c)\cap \mathcal{C}(d)| $                                                    &         &       \\ \cline{1-7}
        \multirow{5}{*}{simplicial complex}               & \multirow{5}{*}{BAMP} & \multirow{5}{*}{gen}  &         $c\in \mathcal{H}$                      & $\textbf{x}_c^{\text{new}}            = \phi\left( \textbf{x}_c, m^{\mathcal{B}}_c, m^{\mathcal{C}}_c,            m^{\mathcal{N}_{\uparrow}}_c, m^{\mathcal{N}_{\downarrow}}_c\right)$                                                                    &  \multirow{5}{*}{~\cite{bodnar_weisfeiler_2021}}       &        \\ \cline{5-5}
        &                       &                       &                                         & $m^{\mathcal{B}}_c = \bigoplus_{d\in            \mathcal{B}(c)}\psi_{\mathcal{B}}\left( \textbf{x}_c,\textbf{x}_d            \right)$                                                                                                                                                                                            &   &             \\ \cline{5-5}
        &                       &                       &                                         & $m^{\mathcal{C}}_c = \bigoplus_{d\in            \mathcal{C}(c)}\psi_{\mathcal{C}}\left(            \textbf{x}_c,\textbf{x}_d\right)$                                                                                                                                                                                          &   &             \\ \cline{5-5}
        &                       &                       &                                         & $m^{\mathcal{N}_{\uparrow}}_c = \bigoplus_{d\in            \mathcal{N}_{\uparrow}(c), \delta\in \mathcal{C}(c)\cap \mathcal{C}(d)            }\psi_{{\uparrow}}\left(            \textbf{x}_c,\textbf{x}_d, \textbf{x}_{\delta}\right)$                                                                                       &     &           \\ \cline{5-5}
        &                       &                       &                                         & $m^{\mathcal{N}_{\downarrow}}_c = \bigoplus_{d\in            \mathcal{N}_{\downarrow}(c), \delta\in \mathcal{C}(c)\cap \mathcal{C}(d)            }\psi_{{\downarrow}}\left(            \textbf{x}_c,\textbf{x}_d, \textbf{x}_{\delta}\right)$                                                                                 &     &           \\ \cline{1-7}
        \multirow{2}{*}{simplicial complex}               & \multirow{2}{*}{BAMP} & \multirow{2}{*}{att}  & $c\in \mathcal{H}$                      & $\textbf{x}_{c}^{\text{new}} = \phi\left( \sum_{d\in            \mathcal{N}_{\uparrow}(c)}            \alpha_{c,d}^{\mathsf{U}}\Theta_{\mathsf{U}} \textbf{x}_{d},            \sum_{d\in \mathcal{N}_{\downarrow}(c)}            \alpha_{c,d}^{\mathsf{L}}\Theta_{\mathsf{L}} \textbf{x}_{d}            \right)$              &  \multirow{2}{*}{~\cite{goh_simplicial_2022}}   &             \\  \cline{5-5}
        &                       &                       &                                         & $\alpha_{c,d}^{\mathsf{N}} = \tfrac{\exp(\sigma            (\operatorname{sim}(\textbf{x}_c\Theta_{\mathsf{N}},            \textbf{x}_d\Theta_{\mathsf{N}})))            }{\sum_{e\in \mathcal{N}}\exp(\sigma            (\operatorname{sim}(\textbf{x}_c\Theta_{\mathsf{N}},            \textbf{x}_d\Theta_{\mathsf{N}})))}$ &        &        \\ \cline{1-7}

    \multirow{2}{*}{simplicial complex}               & \multirow{2}{*}{BAMP} & \multirow{2}{*}{conv}  & $c\in \mathcal{H}$                      & $\textbf{x}_{c}^{\text{new}} = \sigma\left( \sum_{d\in \mathcal{N}_{\uparrow}(c)}\Theta_{\mathsf{U}} \textbf{x}_{d} + \sum_{d\in \mathcal{N}_{\downarrow}(c)} \Theta_{\mathsf{L}} \textbf{x}_{d} + \Theta_{\mathsf{S}} \textbf{x}_{c} \right)$              & \multirow{2}{*}{~\cite{yang_simplicial_2022}}   &      \multirow{2}{*}{~\cite{ebli_simplicial_2020, bunch_simplicial_2020}}        \\  \cline{5-5}
    &                       &                       &                                         &    \makecell{Other formulations are complex \\ and omitted due to space constraints} &   &        \\ \cline{1-7}
        {node-tuple-c}                  & DAMP                  & gen                  & $\textbf{v}\in            \mathfrak{T}$ & So far unexplored                                                       &         \multirow{1}{*}{-}       &  \\ \cline{1-7}
        {node-tuple-c}                  & DAMP                  & att                  & $\textbf{v}\in            \mathfrak{T}$ & So far unexplored                                                                           &      \multirow{1}{*}{-}                 &\\ \cline{1-7}
        {node-tuple-c}                  & DAMP                  & conv                  & $\textbf{v}\in            \mathfrak{T}$ & $\textbf{x}_\textbf{v}^{\text{new}} = \sigma\left(            \Theta_1\textbf{x}_{\textbf{v}}+\sum_{\textbf{u}\in            \mathcal{N}_{\downarrow}(\textbf{v})            }\Theta_2\textbf{x}_{\textbf{u}}\right)$                                                       &                 ~\cite{morris_weisfeiler_2021}                                 & ~\cite{morris2020weisfeiler} \\
        \bottomrule
    \end{tabular}\hspace*{-1cm}
    \caption{Selection of HOGNNs which can be expressed
    as MP architectures. We list the used \acrfull{hogdm}, the
    \acrlong{mp} wiring scheme (Wiring), the \acrlong{mp} flavour (Flavour), which
    objects have their features updated during \acrlong{mp} (Obj), and the update
    equation. In the update equations, $\bigoplus,\bigotimes, \bigodot$ refer to
    permutation-invariant aggregators, $\Theta$ are learnable matrices, $\psi,\phi$
        are learnable functions, $\text{sim}$ is a real-valued similarity measure for
        two vectors of equal size, for example, an inner product. 
    }
    \label{tab:overview_higher_order_methods}
\end{table*}

\renewcommand{\arraystretch}{1}

        \normalsize

\subsection{Neural MP on Hypergraphs (HGs)}
\label{sec:hognns-hgs}



In HOGNN models based on HGs, the dominant approach for wiring is to define message passing channels by incidence; we call such models \textbf{incidence based message passing} models or \textbf{IMP} for short.
Here, information flows between nodes and hyperedges. This architecture appears in multiple HG neural network models~\cite{feng_hypergraph_2019, bai_hypergraph_2021, heydari_message_2022} and was described in a general form by Heydari and Livi~\cite{heydari_message_2022}. 
Note that an alternative perspective on such models is that one can see an HG as a plain bipartite graph, as there are two node sets: HG nodes and HG hyperedges, and edges between them are defined by incidence; these edges are used as channels in MP.

Given an HG with features $\mathcal{H}=(\mathcal{V},\mathcal{E},\textbf{x})$, during one MP step, node and hyperedge features are updated as follows:

\ifsq\footnotesize\fi
\begin{align}
    \textbf{x}^{\text{new}}_e &= \phi_{\mathcal{E}}\left( \textbf{x}_e,
    \bigodot_{u\in \mathcal{V}: u\sub e}\underbrace{\psi_\mathcal{V}\left(\textbf{x}_{u}\right)}_{\text{message }u\to
    e}\right), \label{line:hgmp-ln-1} \\
    \textbf{m}_e &= \psi_{\mathcal{E}}\left( \textbf{x}_e,
    \bigotimes_{u\in \mathcal{V}: u\sub e}\underbrace{\psi_\mathcal{V}\left(\textbf{x}_{u}\right)}_{\text{message }u\to
    e}\right), \label{line:hgmp-ln-2} \\
    \textbf{x}^{\text{new}}_v &= \phi_\mathcal{V}\left( \textbf{x}_v, \bigoplus_{f\in
    \mathcal{E}: v\sub f}\underbrace{\textbf{m}_f}_{\text{message }f\to v}
    \right), \label{line:hgmp-ln-3}
\end{align}
\normalsize

\noindent
for $v\in \mathcal{V}$, $e\in \mathcal{E}$, learnable functions $\phi_\mathcal{V},\phi_{\mathcal{E}}, \psi_\mathcal{V},\psi_{\mathcal{E}}$, and possibly distinct permutation-invariant aggregators $\bigodot, \bigotimes, \bigoplus$.
Thus, messages first travel from nodes to incident hyperedges, and these hyperedges aggregate the messages to obtain new features $\textbf{x}^{\text{new}}_e$ for themselves, using $\bigodot$ (Eq~(\ref{line:hgmp-ln-1})). Next, each hyperedge builds a message (Eq.~(\ref{line:hgmp-ln-2})), possibly using a different aggregation scheme $\bigotimes$. This message is then sent to all incident nodes, and these messages can be further aggregated using a yet another scheme $\bigoplus$.
In contrast to MP on plain graphs, this architecture performs two levels of nested aggregations per MP step, since computing $\textbf{x}^{\text{new}}_v$ requires knowing $\textbf{m}_e$.
%

There are numerous IMP architectures that harness HGs, for example Hypergraph Convolution~\cite{bai_hypergraph_2021}, Hypergraph Neural Networks~\cite{feng_hypergraph_2019}, Hypergraph Attention~\cite{bai_hypergraph_2021}, HyperGCN~\cite{yadati2019hypergcn}, Hypergraph Networks with Hyperedge Neurons~\cite{dong2020hnhn}, Hyper-SAGNN~\cite{zhang2019hyper}, and others~\cite{ma2022learning, zhang2022hypergraph, song2023chgnn, wang2022equivariant, aponte2022hypergraph, arya2021adaptive, fu2022p, fu2019hplapgcn}. 
%

\subsection{Neural MP on Simplicial Complexes (SCs)}
\label{subsec:hypergraph:simplicial_complex}

A primary mode of wiring in HOGNN models based on SCs is through \textbf{boundary adjacencies} and related neighborhood notions.
In this mode, a vertex or a hyperedge can receive messages from its \mbox{boundary-,} co-boundary-, upper-, and lower adjacencies, or a subset thereof. We describe the MP architecture by Bodnar et al.~\cite{bodnar_weisfeiler_2021}, see also work by Hajij et al.~\cite{hajij2022simplicial} and others~\cite{isufi2022convolutional}.

Let $\mathcal{H}=(\mathcal{V},\mathcal{E},\textbf{x})$ be an SC. During one step of
\acrlong{mp}, node and simplex features are updated as follows:

\ifsq\footnotesize\fi
\begin{align*}
    m^{\mathcal{B}}_c &= \bigoplus_{b\in
    \mathcal{B}(c)}\psi_{\mathcal{B}}\left( \textbf{x}_c,\textbf{x}_b
    \right),\quad 
    m^{\mathcal{C}}_c = \bigoplus_{b\in
    \mathcal{C}(c)}\psi_{\mathcal{C}}\left( \textbf{x}_c,\textbf{x}_b
    \right), \\
    m^{\mathcal{N}_{\uparrow}}_c &= \bigoplus_{d\in
    \mathcal{N}_{\uparrow}(c), \delta\in \mathcal{C}(c)\cap \mathcal{C}(d)
    }\psi_{{\uparrow}}\left(
    \textbf{x}_c,\textbf{x}_d, \textbf{x}_{\delta}\right), \\
    m^{\mathcal{N}_{\downarrow}}_c &= \bigoplus_{d\in
    \mathcal{N}_{\downarrow}(c), \delta\in \mathcal{B}(c)\cap \mathcal{B}(d)
    }\psi_{{\downarrow}}\left(
    \textbf{x}_c,\textbf{x}_d, \textbf{x}_{\delta}\right), \\
    \textbf{x}_c^{\text{new}} &= \phi\left( \textbf{x}_c, m^{\mathcal{B}}_c,
    m^{\mathcal{C}}_c, m^{\mathcal{N}_{\uparrow}}_c, m^{\mathcal{N}_{\downarrow}}_c
    \right),
\end{align*}
\normalsize
for learnable maps $\phi,\psi_{\mathcal{B}},\psi_{\mathcal{C}},
\psi_{\uparrow},\psi_{\downarrow}$ and potentially different permutation-invariant aggregators $\bigoplus$.
We call models using this set of update equations boundary-adjacency based message passing models or \textbf{BAMP} for short.

In BAMP, compared to IMP, messages may have a smaller reach. For example, two edges
that neither share nodes nor co-boundaries will not exchange messages in
\gls{bamp}, while, in \gls{imp}, they will if they are incident to a common
hyperedge. 
%
Let us consider the effect that a node $u$ has on the feature update
of node $v$ in the two architectures. \maciej{should it be "hyperdges u, v"? otherwise, for nodes u,v the following IMP and BAMP are the same?} \florian{Why? Can't u and v belong to the same hyperedge but not be upper-adjacent?} In \gls{imp}, $u$ first sends its message
to all hyperedges it is incident on. These aggregate the messages they receive
from their incident nodes. Then all hyperedges containing $v$ send their
message to $v$. Thus, the contribution of $u$ to the feature update of $v$
depends on how many co-incidences $u$ and $v$ share, and to what extent the
feature of $u$ is retained after the aggregations. In \gls{bamp}, $u$ and $v$
exchange messages if and only if they share an edge. The message from $u$ to
$v$ contains the features of $u,v$, and the feature of their shared edge $e$.
These features are then transformed by $\psi_{\uparrow}$. Finally, the messages
from all upper adjacencies of $v$ are aggregated, forming the collective
message $m^{\mathcal{N}_{\uparrow}}_c$, which is fed into the final update.
There are also variants of \gls{bamp} which use nested
aggregations~\cite{hajij_cell_2021}.



There are numerous examples of MP based on SCs; example models include Message passing simplicial networks (MPSNs)~\cite{bodnar_weisfeiler_2021}, Simplicial 2-complex convolutional neural networks (S2CNNs)~\cite{bunch_simplicial_2020}, Simplicial attention networks~\cite{goh_simplicial_2022} (SATs), 
or SCCNN~\cite{yang2023convolutional} and their specialized variant~\cite{yang_simplicial_2022}. These are all BAMP-based models.

\subsection{Neural MP on Cell Complexes (CCs)}\label{subsec:hypergraph:cell_complex}

Neural MP on CCs uses, as the primary adjacency scheme, the BAMP adjacency. However, unlike in SCs, it usually uses a restricted version of BAMP.
For example, CW Networks~\cite{bodnar_weisfeiler_2022} employ MP on cell complexes by sending messages from boundary-adjacent and upper-adjacent cells. The representation of a cell $c$ is updated according to

\begin{align}\label{eqn:cell_message_passing}
    m_{\mathcal{B}}^{(l+1)}(c) &= \bigoplus_{b\in
    \mathcal{B}(c)}\psi_{\mathcal{B}}\left( \textbf{H}_c^{(l)},\textbf{H}_b^{(l)}
    \right) \\
    m_{\mathcal{N}_{\uparrow}}^{(l+1)}(c) &= \bigoplus_{d\in
    \mathcal{N}_{\uparrow}(c), \delta\in \mathcal{C}(c)\cap \mathcal{C}(d)
    }\psi_{{\uparrow}}\left(
    \textbf{H}_c^{
        (l)},
    \textbf{H}_d^{(l)}, \textbf{H}_{\delta}^{(l)}\right) \\
    \textbf{H}_c^{(l+1)} &= \phi\left( \textbf{H}_c^{(l)}, m_{\mathcal{B}}^{(l+1)}
    (c), m_{\mathcal{N}_{\uparrow}}^{(l+1)}(c)\right)
\end{align}
for learnable functions $\phi,\psi_{\mathcal{B}},\psi_{{\uparrow}}$ and
permutation-invariant aggregators $\bigoplus$. An embedding on the CC 
$\mathcal{K}$ can be obtained by pooling multisets of cell representations
\begin{align*}
    \textbf{H}_{K} = \textsc{Pool}\left( \{\{\textbf{H}_c^{(L)}\}\}_{\dim=j}: j=0,
    \dots,\dim_{\max}  \right).
\end{align*}
Interestingly, this model considers only boundary and upper adjacency, while
omitting co-boundary and lower adjacency. The authors show that this does not harm
the expressivity of the model in terms of its ability to distinguish
non-isomorphic CCs.

There are other models for MP on CCs, including
Cell complex neural networks (CCNNs)~\cite{hajij_cell_2021}
or
Cell Attention Networks (CANs)~\cite{giusti2022cell}.


\subsection{Neural MP on NT-Col-Graphs}
\label{sec:hognns-ntcol}

Node-tuples act as a tool for capturing information across different parts of the graph beyond that available when using plain graphs; they have been used to strengthen the expressive power of GNNs. Several works target this class of models~\cite{maron2019provably, morris_weisfeiler_2021, morris2020weisfeiler, zhou2023distance}. The proposed wiring pattern by Morris et al.~\cite{morris_weisfeiler_2021, morris2020weisfeiler}, called the \textbf{Down-Adjacency Message Passing (DAMP)}, is based on down adjacencies (see Section~\ref{sec:graph_with_node_tuples}).

\begin{definition}[Down-adjacency \acrlong{mp} for node-tuple
collections]\label{def:mp_down_adjacency}
    Given a node-tuple-collection $\mathcal{C}=(G,\mathfrak{T},
    \textbf{x})$, and a neighbourhood notion given by down- or local down-adjacency,
    $\mathcal{N}\in \{\mathcal{N}_{\downarrow},
    \mathcal{N}_{\text{loc},\downarrow}\}$, we update the
    features of node-tuples
    in one \acrlong{mp} step by aggregating messages from a node-tuple's
    neighbours
    \begin{align*}
        \textbf{x}^{\text{new}}_\textbf{v} = \phi\left( \textbf{x}_\textbf{v},
        \bigoplus_{\textbf{u}\in
        \mathcal{N}(\textbf{v})}\underbrace{\psi\left
        (\textbf{x}_\textbf{v},\textbf{x}_\textbf{u} \right)}_{\text{message
        }\textbf{u}\to \textbf{v}}\right).
    \end{align*}
\end{definition}

\Gls{damp} is similar to \gls{bamp} when using only lower adjacency. However,
in \gls{damp} $\mathcal{N}_{\downarrow}, \mathcal{N}_{\text{loc},\downarrow}$
refer to down and local down adjacency respectively (see
Definition~\ref{def:node_tuple:down_local_down}), while in \gls{bamp}
$\mathcal{N}_{\downarrow}$ denotes lower adjacency (see
Definition~\ref{def:hypergraph:incidence_boundary_adjacency}).


\subsection{Neural MP on SCol-Graphs \& ST-Col-Graphs}\label{sec:subgraph_repL}

Finally, we turn to architectures for subgraph-collections. There is a large body of research on this topic and numerous models have been proposed~\cite{pappDropGNNRandomDropouts2021, cotta_reconstruction_2021, frasca2022understanding, bevilacqua_equivariant_2022, zhao_stars_2022, zhang_nested_2021,  sandfelder_ego-gnns_2021, alsentzer_subgraph_2020, bouritsas_improving_2021, thiede_autobahn_2022, zhao2022practical, zhou2023relational, zhang2023complete, huangboosting}. In general, the architectures based on subgraph-collections usually proceed as follows:

\begin{enumerate}[leftmargin=1.0em]
    \item \label{step:construct_subgraphs} Construct a collection of subgraphs from
    an input graph.
    \item \label{step:subgraphs_wiring} Specify the wiring scheme for subgraphs.
    \item \label{step:learn_subgraph_representations} Learn representations for the
    subgraphs, harnessing the specified wiring pattern.
    \item  \label{step:aggregate_subgraphs} Aggregate subgraph representations
    into a graph representation.
\end{enumerate}
\normalsize

We identify three classes of methods for constructing subgraph collections and learning their representations. Based on this, we further classify these HOGNN architectures into three types: \textbf{reconstruction-based}, \textbf{ego-net}, and \textbf{general subgraph wiring} approaches.
In reconstruction-based schemes, one removes nodes or edges and learns representations for the resulting induced subgraphs~\cite{rongDropEdgeDeepGraph2020, pappDropGNNRandomDropouts2021, cotta_reconstruction_2021, bevilacqua_equivariant_2022}.
In ego-net schemes, subgraphs are constructed by selecting vertices according to some scheme, and then building subgraphs using multi-hop neighborhoods of these vertices~\cite{zhao_stars_2022, zhang_nested_2021, sandfelder_ego-gnns_2021, bevilacqua_equivariant_2022}.
In general subgraph wiring, the approach is to pick specific subgraphs of interest~\cite{alsentzer_subgraph_2020}, for example subgraphs
that are isomorphic to chosen template graphs~\cite{thiede_autobahn_2022}.

The subgraph wiring scheme usually simply follows the general subgraph adjacency, and is architecture-specific. Schemes vary; for example, Bar-Shalom et al.~propose to associate subgraphs with node clusters rather than with individual nodes~\cite{bar2024flexible}.
Some schemes do not consider adjacency between subgraphs at all and instead simply pool subgraph representations into a graph representation~\cite{pappDropGNNRandomDropouts2021, cotta_reconstruction_2021}, thereby treating them as bags of subgraphs similar to DeepSets~\cite{zaheer_deep_2018}.

Finally, Qian et al.~\cite{qian2022ordered} have proposed a scheme that imposes an ordering between subgraphs, in addition to its subgraph adjacency used for wiring.

\subsection{Neural MP on Motif-Graphs \& SCnt-Graphs}
\label{sec:hognns-motifs}

Neural MP implemented in HOGNNs based on Motif-Graphs follows the motif-adjacency defined in the specific work.
Interestingly, it is usually defined and implemented using ``global'' formulations that explicitly use adjacency matrices~\cite{rossiHigherorderNetworkRepresentation2018, lee2019graph}. As such, these HOGNNs do not explicitly harness messages between motifs. Instead, they implicitly provide this mechanism in their matrix-based formulations for constructing embeddings.
Instead, models that are based on SCnt-Graphs often explicitly use wiring based on the input graph structure~\cite{bouritsas_improving_2021}.

\subsection{Discussion}

We briefly discuss different aspects related to all the above-summarized HOGNNs.

\subsubsection{Wiring Flavors}
\label{sec:flavors}

The choice of wiring flavor is important for performance (latency, throughput), expressiveness, and overall robustness.
The convolutional flavor is simplest to implement and use, and it is usually advantageous for high performance, as one can precompute the local formulation coefficients for a given input graph. However, it was shown to have less competitive expressiveness than other flavors. \florian{Reference?} On the other hand, attentional and general MP flavors are usually more complicated and harder to implement efficiently, because any of the functional building blocks ($\psi$, $\phi$, $\oplus$) can return scalars or vectors that must be learnt.

We observe that all three wiring flavors (convolutional, attentional, and general MP) are widely used in the neural architectures based on a hypergraph or its variants. Specifically, there have been explicit HOGNNs with all these flavors for HOGNNs based on general HGs, SCs, and CCs; see Table~\ref{tab:overview_higher_order_methods}.
However, HOGNNs based on other GDMs, do not consider all these flavors. For example, architectures based on NT-Col-Graphs have only considered the convolutional flavor so far. Exploring these flavors for such model categories would be an interesting direction for future work.

\iftr
\subsubsection{Local vs.~Global Formulations}
\label{sec:formulations}

Both types of formulations come with tradeoffs.
Global formulations can harness methods from different domains (e.g., linear algebra and matrix computations), for example communication avoidance. They may also be easier to vectorize, because they deal with whole feature and adjacency matrices (instead of individual vectors as in local formulations).
On the other hand, local formulations can be programmed more effectively because one focuses on a ``local'' view from a single vertex, which is often easier to grasp. Moreover, such formulations may also be easier to schedule more flexibly on low-end compute resources such as serverless functions because functions in question operate on single vertices/edges instead of whole matrices.

Many of the considered models across all the GDMs are formulated using the local approach.
This is especially visible with models based on HGs, SCs, CCs and NT-Col-Graphs, as illustrated in Table~\ref{tab:overview_higher_order_methods}. These models directly follow the IMP (Section~\ref{sec:hognns-hgs}), BAMP (Section~\ref{subsec:hypergraph:simplicial_complex}), and DAMP-based (Section~\ref{sec:hognns-ntcol}) wiring patterns, which are intrinsically locally formulated.
However, some models use a mixture of the local and global formulations, for example, S2CNNs or CCNNs.
Finally, few models are globally formulated. Two notable exceptions are the Hypergraph Convolution or HONE.
\fi

\subsubsection{Multi-Hop Channels}
\label{sec:multihops}

Multi-hop channels are often referred to as channels introducing ``higher-order neighborhoods''~\cite{frasca2020sign, abu-el-haija_mixhop_2019}. For example, in MixHop~\cite{abu-el-haija_mixhop_2019} messages are passed multiple hops in every \acrlong{mp} step. This way, a node can see nodes beyond its immediate neighbourhood. If every node saw its $2$-hop neighbourhood in every step, this would suffice to distinguish the two graphs:\ one graph has a $2$-hop neighbourhood which contains a triangle, while the other does not.
There exist several such architectures~\cite{abu-el-haija_mixhop_2019, frasca2020sign, yang2023convolutional, liu2019higher, rossi_hone_2018, feng2022powerful, brossard2020graph}.

Many multi-hop architectures are based on PGs, and the higher order in these architectures solely involves harnessing multi-hop neighborhoods. This includes MixHop~\cite{abu-el-haija_mixhop_2019} and SIGN~\cite{frasca2020sign}.
Some multi-hop based architectures are related to GDMs beyond PGs, for example SCCNN~\cite{yang2023convolutional}, which combines MixHop with SCs.

Interestingly, multi-hop architectures are often formulated using the global approach. This is because there is a correspondence between obtaining the information about $k$-hop neighbors and computing the $k$-th power of the adjacency matrix. This has been harnessed in, for example, SIGN~\cite{frasca2020sign}.

\subsubsection{Nesting}
\label{sec:hognns-nested}

There exist several models that use nesting explicitly. Models from this category can be referred to with different words (besides ``nested''), for example ``hierarchical'' or ``recursive''.
These models mostly harness PGs as the underlying GDM. This includes Hierarchical GNNs~\cite{chen2021hierarchical}, GoGNNs~\cite{wang_gognn_2020}, and others~\cite{fey2020hierarchical, liu2022graph, yadati2020neural}. However, one architecture introduces the concept of Recursive Hypergraphs~\cite{luo2022shine}.
The wiring in these models usually follows a multi-level approach, where messages are consecutively exchanged among vertices at different levels of nesting.

\subsubsection{Existing Blueprints for HOGNNs}
\label{sec:blueprints}

There exist a few related blueprints for HOGNNs.
Importantly, Hajij et al.~\cite{hajijtopological, papillon2023architectures} offer a blueprint for MP focusing on topological GNNs (i.e., based on combinatorial complexes, cellular complexes, simplicial complexes, and hypergraphs). It comes with a comprehensive overview of numerous models expressed with the proposed formulation. The main difference to our work is that our blueprint, by covering classes of models such as subgraph, multi-hop, or node-tuple GNNs, abstracts away and does not focus on certain details offered in~\cite{hajijtopological, papillon2023architectures} (such as the explicit consideration of intra- and inter-neighborhoods), while simultaneously covering aspects related to these non-topological models.
There are also other related blueprints that focus on subgraph GNNs~\cite{frasca2022understanding} or equivariant GNNs~\cite{bevilacqua_equivariant_2022}.
Finally, PyTorch Geometric High Order, a library for HO-GNNs, has been recently developed~\cite{wang2023pytorch}.

\ifuncompressed
\section{Example HOGNN Architectures}

We also provide more details about selected representative HOGNN architectures.

\maciej{This will only be in the arxiv version, will not be included in the TPAMI submission?}

\subsection{HOGNNs on Hypergraphs}

\textbf{Hypergraph Convolution~\cite{bai_hypergraph_2021}} is
the first convolutional hypergraph architecture, designed for learning
node representations based on common hyperedges. 

In \textbf{hypergraph neural networks~\cite{feng_hypergraph_2019}}, node features are
updated according to the following hyperedge convolutional operation:

\begin{align}\label{eqn:hypergraph_convolutional_network}
    \textbf{H}_V^{(l+1)} = \sigma\left( D_V^{-1/2}BWD_{\mathcal{E}}\inv B\transpose
    D_V^{-1/2}\textbf{H}_V^{(l)}\Theta^{(l)} \right)
\end{align}

\noindent
where 
\begin{itemize}[leftmargin=0.75em]
    \item an HG $\mathcal{H}=(V,\mathcal{E})$ has an incidence matrix $B$ and hyperedge weights
$w (e)$ stored in a diagonal matrix $W\in \RE^{m\times m}$ with $W_{jj}=w(e_j)$
for $j=1,\dots,m$; 
\item $D_V\in\RE^{n\times n}$ denote the diagonal node degree
  matrix with $(D_V)_{ii}=\sum_{j=1}^{m}w(e_j)B_{ij}=(BW)_{i,:}$ equal to the
  weighted sum of hyperedges the node $v_i$ is incident on,
  \item $D_{\mathcal{E}}\in \RE^{m\times m}$ is the diagonal hyperedge degree matrix
  with $(D_{\mathcal{E}})_{jj}=\sum_{i=1}^{n}B_{ij}$ counting the number of
  nodes that are incident on hyperedge $e_j\in \mathcal{E}$,
\item $\Theta^{(l)}$ is a learnable matrix, 
\item The convolution operator
$D_V^{-1/2}BWD_{\mathcal{E}}\inv B\transpose D_V^{-1/2}$ is a normalised
version of the hyperedge relatedness $BB\transpose$ which in entry
$(BB\transpose)_{ij}$ counts the number of hyperedges a pair of nodes
$v_i, v_j$ are simultaneously incident on. 
\end{itemize}

Applying this model to unweighted
graphs, one recovers graph convolutional networks
(GCNs)~\cite{kipf_semi-supervised_2017}. The diagonal edge
degree matrix simplifies to $D_{\mathcal{E}} = 2I$, $W=I$, moreover, we have
$BB\transpose=A+D_V$. Thus
Equation~\ref{eqn:hypergraph_convolutional_network} transforms to

\begin{align*}
    \textbf{H}_V^{(l+1)} &= \sigma\left( D_V^{-1/2}BWD_{\mathcal{E}}\inv B\transpose
    D_V^{-1/2}\textbf{H}_V^{(l)}\Theta^{(l)} \right)\\
    &= \sigma\left( \tfrac{1}{2}(I+(D_V)^{-1/2}A(D_V)^{-1/2}) \textbf{H}_V^{(l)
    }\Theta^{(l)}\right)
\end{align*}

\noindent
which corresponds to the GCN node feature update up to
normalization~\cite{kipf_semi-supervised_2017}.

\textbf{Hypergraph Attention (HAT)~\cite{bai_hypergraph_2021}} builds on this convolutional architecture and further proposes an attention module for learning a weighted incidence matrix. The attentional module in layer $l$ is computed according to 

\begin{align*}
    B^{\text{att},(l)}_{ij} = \frac{\exp\left( \sigma\left( \text{sim}
    (\textbf{x}_{v_i}\Theta^{(l)},\textbf{x}_{e_j}\Theta^{(l)})
    \right)
    \right)}{\sum_{k:\{v_k,v_i\}\sub e_j}\exp\left( \sigma \left( \text{sim}(
        (\textbf{x}_{v_k}\Theta^{(l)},\textbf{x}_{e_j}\Theta^{(l)})
        \right) \right)}
\end{align*}

for a similarity function sim$^{(l)}$ of two vectors and learnable parameter matrices $\Theta^{(l)}$. 
This similarity function can be given by, for example, an inner product with a learnable vector $a^{(l)}\in \RE^{2d_\mathsf{l}}$, i.e., $\text{sim}^{(l)}(x,y) = (a^{(l)})\transpose(x \parallel y)$.

\noindent
Note that it is essential for hyperedges and nodes to have a representation in the same domain, so their inner product can be computed (more specifically, since the similarity function takes two elements from the same domain as input).

Although closely related, HAT does not directly generalize GAT~\cite{velickovic_graph_2018}. While HAT uses an attention module to learn the incidence matrix, GAT uses it to learn adjacency between nodes. However, an HG $\mathcal{H}=(V,\mathcal{E})$ can be expressed as a bipartite graph $G_{\mathcal{H}}=(V_{G_{\mathcal{H}}},E_{G_{\mathcal{H}}})$ with node sets $V_{G_{\mathcal{H}}} = V \sqcup \mathcal{E}$ and $\{v_i, e_j\} \in E_{G_{\mathcal{H}}}$ if $v_i \sub e_j$. Hypergraph attention then corresponds to GAT on the bipartite graph $G_{\mathcal{H}}$.

\subsection{HOGNNs on Simplicial Complexes}

\textbf{Message passing simplicial networks (MPSNs)~\cite{bodnar_weisfeiler_2021}}
introduce the MP framework for SCs. Here, messages are sent from upper and
boundary adjacencies and, in addition, from from lower and co-boundary
adjacencies. The MP communication channels used are BAMP only. 

\textbf{Simplicial 2-complex convolutional neural networks
(S2CNNs)~\cite{bunch_simplicial_2020}} focus on SCs of dimension
at most $2$. In addition to upper and lower adjacencies, boundary adjacencies are
considered for the feature updates. They additionally
normalise the adjacency and boundary-adjacency
matrices~\cite{schaubRandomWalksSimplicial2020}. For better readibility we omit the details of the
normalisation and collectively denote the normalisation
functions by $\Phi(\cdot)$. The feature updates are then given by
\begin{align*}
    \textbf{H}_{\mathcal{K}_0}^{(l+1)} = \sigma\Big( \Big[ &\Phi(B_1B_1^T)
    \textbf{H}_{\mathcal{K}_0}^{(l)
    }\Theta^{
        (l)}_{0,0} \parallel \Phi(B_1) \textbf{H}_{\mathcal{K}_1}^{(l)} \Theta^{
        (l)}_{1,0} \Big] \Big),\\
    \textbf{H}_{\mathcal{K}_1}^{(l+1)} = \sigma\Big( \Big[ &\Phi(B_1^T)
    \textbf{H}_{\mathcal{K}_0}^{(l)} \Theta^{
        (l)}_{0,1} \\ \parallel
    &\Phi(B_1^TB_1,B_2B_2^T)
    \textbf{H}_{\mathcal{K}_1}^{(l)} \Theta^{
        (l)}_{1,1} \\ \parallel &\Phi(B_2) \textbf{H}_{\mathcal{K}_2}^{(l)} \Theta^{
        (l)}_{2,1} \Big] \Big),\\
    \textbf{H}_{\mathcal{K}_2}^{(l+1)} = \sigma\Big( \Big[ &\Phi(B_2^T)
    \textbf{H}_{\mathcal{K}_1}^{(l)} \Theta^{
        (l)}_{1,2} \parallel \Phi(B_2\transpose B_2)
    \textbf{H}_{\mathcal{K}_2}^{(l)} \Theta^{
        (l)}_{2,2}  \Big] \Big)
\end{align*}
for learnable parameter matrices $\Theta^{(l)}_{i,j}$ in every layer $l$.
This framework can be generalised to any order $p$ of simplices, by considering
functions of $B_p\transpose, B_{p}\transpose B_p, B_{p+1}B_{p+1}\transpose$
and $B_{p+1}$ for computing updates based on boundary adjacent $(p-1)$-simplices,
lower and upper-adjacent $p$-simplices and co-boundary adjacent $(p+1)$-simplices
respectively.

In the simplicial network methods considered above, adjacencies between simplices
and the resulting feature representation updates are tied to the structure of the
SC, given by functions of the boundary operator. \textbf{Simplicial
attention networks~\cite{goh_simplicial_2022} (SATs)} generalises
GATs~\cite{velickovic_graph_2018}
and allow these adjacencies to be learned. For two $p$-simplices $c,d$ in an
SC $\mathcal{K}$, their relative orientation $o_{c,d}\in \{\pm
1\}$ is $1$ if they have equal orientation and $-1$ otherwise. SATs consider
attention coefficients $\alpha_{c,d}^{\uparrow},\alpha_{c,d}^{\downarrow}$ for upper
and lower adjacencies to update simplex features
\begin{align*}
    \textbf{H}_{c}^{(l+1)} = \phi\left( \sum_{d\in \mathcal{N}_{\uparrow}(c)}
    \alpha_{c,d}^{\uparrow, (l)}\Theta_{\mathsf{U}}^{(l)} \textbf{H}_{d}^{(l)},
    \sum_{d\in \mathcal{N}_{\downarrow}(c)}
    \alpha_{c,d}^{\downarrow, (l)}\Theta_{\mathsf{L}}^{(l)} \textbf{H}_{d}^{(l)}
    \right)
\end{align*}
for an update function $\phi$ which aggregates its two inputs and performs
further computations, for example given by an MLP. The attention coefficients are
computed layerwise and updated according to
\begin{align*}
    \alpha_{c,d}^{\uparrow, (l)} &= o_{c,d}\cdot \text{softmax}(\textsc{Att}
    (\Theta_{\mathsf{U}}^{(l)}\textbf{H}_{c}^{(l)},\Theta_{\mathsf{U}}^{(l)}\textbf{H}_{d}^{(l)}))\\
    \alpha_{c,d}^{\downarrow, (l)} &= o_{c,d}\cdot \text{softmax}(\textsc{Att}
    (\Theta_{\mathsf{L}}^{(l)}\textbf{H}_{c}^{(l)},\Theta_{\mathsf{L}}^{(l)
    }\textbf{H}_{d}^{(l)}))
\end{align*}
where $\textsc{Att}$ is an attention function, for example an inner product. Note
that the learnable matrices $\Theta_{\mathsf{U}}^{(l)},\Theta_{\mathsf{L}}^{(l)}$
used to compute the attention coefficients coincide with those in the feature
updates.

\textbf{SCCNN}~\cite{yang2023convolutional} and their specialized variant~\cite{yang_simplicial_2022} are also BAMP - they use Hodge Laplacians as the basis of their simplicial convolutions (therefore boundary-adjacency in simplicial complexes) and combine this with MixHop.

\subsection{HOGNNs on Cell Complexes}

\textbf{Cell complex neural networks (CCNNs)~\cite{hajij_cell_2021}} are designed
similarly to CW Networks~\cite{bodnar_weisfeiler_2022},
but differ in two main aspects. Firstly, they restrict boundary-adjacency to cells
with contiguous dimensions. This also transfers to the derivative notions of
adjacency, such that only cells of equal dimension can be lower or upper-adjacent.
Secondly, in CCNNs messages are only sent from only upper-adjacent cells. In
particular, the highest-dimensional cells do not receive any messages are their
representations are not updated. Similarly to hypergraph message passing
networks~\cite{heydari_message_2022} the aggregation is nested. Feature
updates for a cell $c$ are given by:

\small
\begin{align*}
    \textbf{H}^{(l+1)}_c &= \phi\left( \textbf{H}^{(l)}_c, \bigoplus^{
        (1)}_{d\in\mathcal{N}_{\uparrow}}\underbrace{\psi_1\left
    (\textbf{H}^{(l)}_c,\textbf{H}^{(l)}_d,
    \bigoplus^{(2)}_{e\in \mathcal{C}(c)\cap \mathcal{C}(d)}\underbrace{\psi_2\left
    (\textbf{H}^{(l)}_{e}\right)}_{M_{e\to d}}
    \right)}_{M_{d\to c}}
    \right)
\end{align*}
\normalsize

for learnable functions $\phi,\psi_1,\psi_2$ and permutation-invariant aggregators
$\bigoplus^{(1)},\bigoplus^{(2)}$. The authors also propose a convolutional
architecture based on a cellular adjacency matrix. Letting
$\hat{n}=|\mathcal{K}-\mathcal{K}_{\dim(\mathcal{K})}|$ denote
the number of cells in a CC $\mathcal{K}$ omitting cells of the highest
dimension, the
cell adjacency matrix $A^{\mathcal{K}}\in \ZE^{\hat{n}\times \hat{n}}$ tracks the
number of common coboundaries of two cells. For $i,j\in [1:\hat{n}]$,
$A^{\mathcal{K}}_{ij}=|\mathcal{C}(c_i)\cap \mathcal{C}(c_j)|$. The diagonal cell
degree matrix is defined by $D^{\mathcal{K}}_{ii} =
\sum_{j=1}^{\hat{n}}A^{\mathcal{K}}_{ij}$ for $i=1,\dots,\hat{n}$. Letting
$\hat{A}^{\mathcal{K}}=I_{\hat{n}}+(D^{\mathcal{K}})^{-1/2}A^{\mathcal{K}}
(D^{\mathcal{K}})
^{-1/2}$, the convolutional cell complex network (CCXN) update reads
\begin{align*}
    \textbf{H}^{(l+1)}_{\mathcal{K}} = \sigma\left( \hat{A}^{\mathcal{K}}
    \textbf{H}^{(l+1)}_{\mathcal{K}} \Theta^{(l)}\right)
\end{align*}
for a learnable parameter matrix $\Theta^{(l)}$, resembling the update in
GCN~\cite{kipf_semi-supervised_2017}.

\subsection{HOGNNs on NT-Col-Graphs}

Two special cases of the neural architectures based on DAMP and local DAMP are
the \textbf{$k$-GNN} and the \textbf{local $k$-GNN
architecture}~\cite{morris_weisfeiler_2021}.
Here, a convolution-type
architecture is proven to have the same expressive power as the general
message-passing model.
In $k$-GNNs, node-tuple representations are learned
according to messages from lower-adjacencies:

\begin{align}\label{eqn_k_GNN}
\textbf{H}_\textbf{v}^{(l+1)} = \sigma\left( \Theta_1^{(l)
}\textbf{H}_{\textbf{v}}^{(l)} +
\sum_{\textbf{u}\in
\mathcal{N}_{\downarrow}(\textbf{v})}\Theta_2^{(l)}\textbf{H}_{\textbf{u}}^{(l)}
\right)
\end{align}

for learnable parameter matrices $\Theta_1^{(l)},\Theta_2^{(l)}$.
In local $k$-GNNs, the feature update is as in Equation~\ref{eqn_k_GNN}, except
that the summation is over local neighbourhoods.

In \textbf{$k$-invariant-graph-networks
($k$-IGNs)~\cite{maron_invariant_2019}}, the collective feature
representation $\textbf{H}_{V^k}^{\text{out}}$ for
node-$k$-tuples is computed by feeding feature matrices through
node-permutation-equivariant linear layers $L_i$ followed by non-linearities
$\sigma$. The output is then transformed by a node-permutation-invariant layer $h$
followed by an MLP:
\begin{align*}
    \textbf{H}_{V^k}^{\text{out}} = \textsc{MLP}\circ h\circ L_d\circ \sigma \circ
    \dots \circ \sigma \circ L_1 (\textbf{X}_{V^k})
\end{align*}
They authors have proven that $k$-IGNs are at least as powerful as $k$-WL at
distinguishing non-isomorphic graphs~\cite{maronProvablyPowerfulGraph2020}. Later
it was shown that they actually have the same
power~\cite{geerts_expressive_2020}.
Moreover, the form of linear equivariant and
invariant layers is well understood.
The authors have provided bases for these
linear spaces and proven that their dimensions are independent of the number of
nodes and instead only depend on $k$.

Other models based on the NT-Col-Graph are k-WL-GNN, delta-k-GNN, and delta-k-LGNN+~\cite{morris2020weisfeiler}, or the architecture by Maron et al.~\cite{maronProvablyPowerfulGraph2020}.

\subsection{HOGNNs on SCol-Graphs \& ST-Col-Graphs}

\subsubsection{Ego-Net Architectures}

Ego-GNNs~\cite{sandfelder_ego-gnns_2021} build
a subgraph collection from the one-hop neighbourhoods $\hat{N}_v = \{v\}\cup
\mathcal{N}_v$ for every node $v\in V$. For each neighbourhood, $\hat{N}_v$, an
individual \gls{gnn} is trained, yielding for every node $v\in V$ a
representation $\textbf{x}^{\hat{N}_u}_v$ in the ego-net of its neighbours
$u\in \mathcal{N}_v$.  A new node representation is obtained by averaging
$\textbf{x}_v^{\text{new}}=\textsc{MEAN} (\{\textbf{x}^{\hat{N}_u}_v: u\in
\hat{\mathcal{N}}_v\})$. Finally, a \gls{gnn} takes these new node features as
input and learns a graph representation. \Glspl{ngnn}~\cite{zhang_nested_2021}
uses a similar approach. Initially, they build a subgraph-collection from the
$r$-hop neighbourhood $B_r(v)$ of every node $v\in V$ for some $r\in \nat$.
Next, they learn subgraph representations $\textbf{x}_{B_r(v)}$ for $B_r(v)$
using one \gls{gnn} per subgraph. In the next step, nodes $v$ inherit the
feature $\textbf{x}_{B_r(v)}$ of the subgraph centred at $v$. A \gls{gnn} is
then trained with these features to build a graph representation. Another way
to see this is that the \gls{gnn} performs \acrlong{mp} between subgraphs,
where the \acrlong{mp} channels between the subgraphs are given by the
connectivity of the nodes they are centred at. Further
architectures~\cite{ranjan_asap_2020,zhao_stars_2022} with a similar approach
have been proposed.

The \textbf{Ego-GNN~\cite{sandfelder_ego-gnns_2021}} model learns node
representations based on message-passing within $1$-ego-nets. Conceptually, node
features are updated in two steps. First, a predefined number $p$ of
message-passing steps are
performed within each $1$-ego-net. We refer to this learnable function as
the ego-net encoder and denote the representation of nodes in the
ego-net for vertex $v_i$ by in step $l+1$ by $\phi^p_{\text{ego},i}(\textbf{H}_V^{(l)
})$. Then the information is aggregated across
ego-nets by computing a node $v$'s new representation $\textbf{H}_v^{(l+1)}$ as
the average over representations of $v$ in its neighbours' $1$-ego-nets. Thus the
update is given by

\begin{align*}
    \textbf{H}_{v_i}^{(l+1)} = \frac{\sum_{j\in B_1(v_i)}\phi^p_{\text{ego},j}
        (\textbf{H}^{(l)}_{v_i})}{\deg(v_i)+1}.
\end{align*}

The authors show that
in contrast to standard GNN methods, Ego-GNNs can identify closed triangles in
a graph. Moreover, they show that their architecture is strictly more powerful
than the WL-test at distinguishing non-isomorphic graphs.

The \textbf{GNN-as-kernel (GNN-AK) method~\cite{zhao_stars_2022}} is based
on similar ideas as Ego-GNNs and NGNNs, allowing the incorporation of the node
representations in different subgraphs and using subgraph
embeddings based on pooling the representations of a base GNN applied to an
$r$-ego-net. Let $\phi^{(l)}_u\left(B_r(v)\right)$ denote the feature
representation of node $u$ computed by the base GNN applied to $B_r(v)$
in layer $l$. Their basic architecture GNN-AK updates node representations based
on their ego-net embedding and the node's representation in its own ego-net:
\begin{align*}
    \textbf{H}_v^{(l+1)|\text{centroid}} &=  \phi_v^{(l)}\left(B_r(v)\right)\\
    \textbf{H}_v^{(l+1)|\text{subgraph}} &= \bigoplus_{u\in B_r(v)}\left(
    \phi^{(l)
    }_u\left(B_r(v)\right) \right)\\
    \textbf{H}_v^{(l+1)} &= \textsc{Fuse}\left( \textbf{H}_v^{(l+1)|\text{centroid}},
    \textbf{H}_v^{(l+1)|\text{subgraph}}  \right)
\end{align*}
where $\bigoplus$ refers to a permutation-invariant aggregator and $\textsc{Fuse}$
to concatenation or sum. They show that using any MPNN as base GNN, this model is
strictly more powerful than the WL-test. Moreover, GNN-AK can distinguish certain
graphs that $3$-WL cannot. They propose a second variant, GNN-AK$^+$, which
incorporates node representations from different ego-nets and 
the distance $d_{u,v}$ between nodes into the node representation updates.
\begin{align*}
    \textbf{H}_v^{(l+1)|\text{centroid}} &=  \phi_v^{(l)}\left(B_r(v)\right)\\
    \textbf{H}_{v,\text{gated}}^{(l+1)|\text{subgraph}} &= \bigoplus_{u\in
    B_r(v)
    }\sigma(d_{u,v})\phi^{(l)
    }_u\left(B_r(v)\right) \\
    \textbf{H}_{v,\text{gated}}^{(l+1)|\text{context}} &= \bigoplus'_{u\in B_r(v)
    }\sigma(d_{u,v})\phi^{(l)
    }_v\left(B_r(u)\right) \\
    \textbf{H}_v^{(l+1)} = \textsc{Fuse}&\left(
    d_{u,v}, \textbf{H}_v^{(l+1)
    |\text{centroid}}, \textbf{H}_{v,\text{gated}}^{(l+1)|\text{subgraph}},
    \textbf{H}_{v,\text{gated}}^{(l+1)|\text{context}}
    \right)
\end{align*}
for a sigmoid function $\sigma$ and possibly two distinct permutation-invariant
aggregators $\bigoplus,\bigoplus'$. They show that GNN-AK$^{+}$ has the same power
as GNN-AK and slightly improves on GNN-AK's performance. 
%

\subsubsection{Reconstruction-Based Architectures}

In DropGNN~\cite{pappDropGNNRandomDropouts2021}, a subgraph is constructed by deleting nodes
of the input graph independently and uniformly at random. In this fashion, one
obtains a collection of node-deleted subgraphs. To build features for these
subgraphs, they are fed to a GNN, which yields subgraph representations.
Finally, these representations are aggregated into a single graph
representation with a set encoder similar to Deep Sets~\cite{zaheer_deep_2018}.
Once the subgraph representations have been computed, no other structural information
is used to build the graph representation. Drop Edge~\cite{rongDropEdgeDeepGraph2020}
uses a related approach: it acts as a regular GNN, except that, during
the MP steps, single edges are removed uniformly at random. Thus node
feature updates take place in edge-removed subgraphs. This method does not
explicitly learn representations for subgraphs - it simply uses the node
representations obtained at the end of the training to build a graph
representation. ReconstructionGNNs~\cite{cotta_reconstruction_2021} learn
representations for a graph from representations of
fixed-size subgraphs of $G$. In the first step, one picks $k\in
\{1,\dots,|V|-1\}$ and builds a subgraph-collection by sampling subgraphs of
$G$ with $k$ nodes. Then a GNN is applied to these node-induced
subgraphs, yielding subgraph representations. Finally, these are transformed
and aggregated into a graph representation, for example, using a set
encoder~\cite{zaheer_deep_2018}. A variant of ReconstructionGNNs uses all
size-$k$ subgraphs. However, this is only feasible for small $k\in \{1,2\}$ or
large $k\in \{n-2,n-1\}$. Like DropGNN, ReconstructionGNNs omit the
interrelational structure of subgraphs.

\subsubsection{General Subgraph-Adjacency Based Schemes}

The third class of methods focuses on more specific subgraphs. \textbf{Subgraph neural
networks~\cite{alsentzer_subgraph_2020}} start with a given subgraph collection.
The goal is to learn a representation for each subgraph which can be used to make
predictions. They devise a \acrlong{mp} scheme based on a sophisticated
subgraph adjacency structure. In \textbf{Autobahn~\cite{thiede_autobahn_2022}}, one
covers a graph with a chosen class of subgraphs, for example, paths and cycles.
The method then applies local operations based on the node intersections of
subgraphs. Subgraph representations are then aggregated
into graph representations using a permutation-invariant aggregator. Finally, the
architecture applies local convolutions to update the graph representation.

\subsection{HOGNNs on Nested GDMs}

\textbf{Nested graph neural networks (NGNNs)~\cite{zhang_nested_2021}} follow a similar approach as
Ego-GNNs, however, they use a different aggregation scheme. Their architecture
is composed of two levels:\ the base GNN and the outer GNN. The base GNN learns
a representation for the $r$-ego-nets of every vertex by applying a GNN
architecture on the $r$-ego-nets. Instead of keeping the representations of
single nodes in every ego-net $B_r(v)$, they are aggregated into a single
ego-net representation $\textbf{H}_{v,ego}$ for every $v\in V$. The outer GNN
treats the ego-net representations $\textbf{H}_{v,ego}$ as node-representations
and performs message-passing to update the ego-net features. Finally, these
updated features can be pooled into a graph representation. The authors show
that when using appropriate GNN architectures for the base and the outer GNN,
NGNNs are strictly more powerful than the WL-test. Moreover, they discuss how
higher-order methods operating on node-tuples can be incorporated into this
nested approach. Since $k$-GNNs consider $\mathcal{O}(n^k)$ node tuples,
applying the method to subgraphs of size $c$ reduces the number of tuples to
$\mathcal{O}(nc^k)$.

In \textbf{GoGNN~\cite{wang_gognn_2020}}, L-graph representations are learned
and then fed to a graph-attention
network~\cite{velickovic_graph_2018} which learns
representations based on neighbourhoods in the H-graph. They apply their work
to predicting chemical-chemical and drug-drug interactions.

The \textbf{SEAL-AI/CI framework~\cite{liSemiSupervisedGraphClassification2019}} has
been devised to solve L-graph classification by learning two classifiers
corresponding to the two levels. In every update step, the information learned
by one classifier is fed to the other. This approach has been applied to social
networks in which L-graphs corresponded to social sub-groups that were
connected by common members and the objective was to distinguish between gaming
and non-gaming groups. 
\fi

\section{Expressiveness in HOGNNs}
\label{chapter:HoGRL_expressiveness}

We also overview approaches for analyzing the expressiveness of HOGNNs.

\subsection{Approaches for Expressiveness Analysis}

There are several ways to formally compare the power of different HOGNNs~\cite{sato2020survey}. One approach is to consider which pairs of input graphs can be distinguished by a given GNN (Section~\ref{sec:HoGRL_expressiveness_isomorphism}). Another approach investigates whether a GNN can count selected subgraphs (Section~\ref{sec:HoGRL_expressiveness_substructure}). Finally, one can also analyze which function classes respective GNNs can approximate (Section~\ref{sec:func-approx}).

\subsubsection{Isomorphism Based Expressiveness} \label{sec:HoGRL_expressiveness_isomorphism}

The idea underlying isomorphism-based expressiveness is to consider which non-isomorphic graphs a given HOGNN can distinguish. The \gls{gi} problem is the computational task of discerning pairs of non-isomorphic simple graphs. Dating back to the 1960s, classifying its complexity remains unsolved. The state-of-the-art algorithm by Babai~\cite{babai_graph_2016,helfgott_graph_2017} runs in quasipolynomial time. However, for many subclasses of graphs, polynomial-time algorithms are known, for example, for planar graphs~\cite{hopcroft_linear_1974}, trees~\cite{kelly_congruence_1957}, circulant graphs~\cite{muzychuk_solution_2004} and permutation graphs~\cite{colbourn_testing_1981}. In the expressiveness analysis of HOGNNs~\cite{zhang2023complete, zhou2023relational, wang2024empirical}, one usually compares an architecture's ability to distinguish non-isomorphic graphs to a heuristic for GI-testing. 

The \gls{wltest}~\cite{weisfeiler_reduction_1968} is an efficient \gls{gi} heuristic based on iterative colour refinement.

\begin{definition}[\acrlong{wltest}]\label{def:1-WL}
    Given a \emph{vertex-coloured} graph $G=(\mathcal{V},E,c^{(0)})$ with colour
    function $c^{(0)}\colon \mathcal{V}\to \Sigma$, vertex colours for $v\in \mathcal{V}$ are updated
    according to
    \begin{align*}
        c^{(l+1)}_v = \textsc{Hash}\left( c^{(l)}_v, \{\{c^{(l)}_u: u\in
        \mathcal{N}_v\}\}
        \right)
    \end{align*}
    for a bijective function $\textsc{Hash}$ to $\Sigma$. In every step $l$, the
    \emph{histogram} of nodes colours $\{\{c^{(l)}_v: v\in \mathcal{V}\}\}$ defines the colour of
    the graph $c_G^{(l)}$. When the graph colour $c_G^{(l)}=c_G^{(l+1)}$ remains equal
    in two consecutive steps, the algorithm terminates. Two graphs are
non-isomorphic if their final graph colours are distinct. Otherwise, the
isomorphism test is inconclusive.
\end{definition}

The \gls{wltest} terminates after at most $\mathcal{O}((|m|+|n|)
\log|\mathcal{V}|)$ iterations~\cite{noauthor_power_nodate}. Recent work has characterised the
types of non-isomorphic graphs that can be distinguished by the
\gls{wltest}~\cite{arvind_power_2015}. For an overview of the power and limitations
of the \gls{wltest}, we refer to~\cite{noauthor_power_nodate}.

As has been noted by several authors, one iteration in the \gls{wltest} closely
resembles \acrlong{mp} when replacing colours with vertex features:
\begin{align*}
    \textbf{x}^{\text{new}}_v = \phi\left( \textbf{x}_v, \bigoplus_{u\in
    \mathcal{N}_v}\underbrace{\psi\left
    (\textbf{x}_v,\textbf{x}_u\right)}_{\text{message
    }u\to v
    }\right).
\end{align*}
The main difference between MP-GNNs and the \gls{wltest} is that feature values are relevant in GNNs, while only the distribution or histogram or colours matters in the \gls{wltest}. Moreover, information may be lost by the transformations $\psi,\phi$ and the aggregation $\bigoplus$. Xu et al.~\cite{xu2018powerful} have shown that MP-GNNs are at most as powerful as the \gls{wltest} at distinguishing non-isomorphic simple graphs. 
%

Multiple variants of the \acrfull{wltest} have been
introduced, often forming the basis for deep learning
architectures~\cite{morris_weisfeiler_2021, morris2021weisfeiler}. This
includes tests for simple graphs~\cite{wangTwinWeisfeilerLehmanHigh2022}, simplicial
complexes, cell complexes~\cite{bodnar_weisfeiler_2022},
node-tuple-collections~\cite{cai_optimal_1992}, and others~\cite{batatia2022mace}. The tests follow a similar
approach:\ starting with a collection of objects with colours and channels
connecting them, every object sends its colour along all outgoing channels to its
adjacencies.
Once all colours have been received, the new colour is defined by a function of
the current colour and multisets of received colours. The algorithm terminates once the graph colours are equal in two
consecutive steps. The most prominent higher-order variants are the two
k-\glspl{wltest}~\cite{cai_optimal_1992},
which apply colour refinement to node-tuple-collections derived from the
isomorphism-type lifting of simple graphs.
%

\begin{definition}[k-\Acrlongpl{wltest}]
    Let $G=(\mathcal{V},\mathcal{E},c^{(0)})$ be a \emph{vertex-coloured} graph with colour
    function $c^{(0)}\colon \mathcal{V}\to \Sigma$ and $k\in \nat$.
    %
    \begin{enumerate}
        \item In the $k$-\gls{wltest}, colours are updated according to
        {\small
        \begin{align*}
            c_{\textbf{v}}^{(l+1)} = \textsc{Hash}\left( c_{\textbf{v}}^{(l)}, \left(
            \{\{c_{\textbf{u}}^{(l)}: u\in \mathcal{N}_j^{k\text{-WL}}(\textbf{v})
            \}\}:
            j\in [k] \right)\right)
        \end{align*}
        }
        \item In the $k$-\gls{fwl}-test, colours are updated according to
        {\small
        \begin{align*}
            c_{\textbf{v}}^{(l+1)} &= \textsc{Hash}\left( c_{\textbf{v}}^{(l)},
            \left\{\left\{
            \left( c_{\textbf{u}}^{
                (l)}: u\in \mathcal{N}_w^{k\text{-FWL}}(\textbf{v}) \right), w\in
            \mathcal{V}  \right\}\right\}
            \right)
        \end{align*}
        }
    \end{enumerate}
    where
    {\small
    \begin{align*}
        \mathcal{N}_j^{k\text{-WL}}(\textbf{v}) &= \left\{ (v_{i_1},\dots,
        v_{i_{j-1}},w,
        v_{i_{j+1}},\dots,v_{i_{k}}) : w\in \mathcal{V} \right\} \\
        \mathcal{N}_w^{k\text{-FWL}}(\textbf{v}) &= \left( (v_{i_1},\dots,v_{i_{j-1}},w,
        v_{i_{j+1}},\dots,v_{i_{k}}): j\in [1:k] \right)
    \end{align*}
    }
    The graph colour is defined as the histogram of node-tuple colours. The
    algorithms terminate when the graph colour of two consecutive steps remains
    unchanged. As before, graphs are non-isomorphic if their final colours
    differ. Otherwise, the test is inconclusive.
\end{definition}

The two variants of the \gls{wltest} on $k$-node tuples differ in the employed neighbourhoods.
The $k$-WL-test works with tuples of neighbourhood colour multisets, in which
multisets are collections of down adjacencies indexed by nodes and tuples are
indexed by coordinates $j\in \{1,\dots,k\}$. The $k$-FWL test uses multisets of
neighbourhood colour tuples in which the tuples are indexed by nodes $w\in \mathcal{V}$ and
the entries of the tuples are down adjacencies indexed by coordinates $j\in
\{1,\dots,k\}$. This subtle difference makes one method more powerful than the other at distinguishing non-isomorphic graphs, namely $k$-FWL has the same discriminative power as $(k+1)$-WL~\cite{cai_optimal_1992} for all $k\geqslant 2$. For $k=1$, both algorithms
simplify to the 1-WL-test (\ref{def:1-WL}). In particular, $1$-WL and $2$-WL have the same distinguishing power. The $k$-FWL
test is known to terminate after $\mathcal{O}(k^2n^{k+1}\log n)$
steps~\cite{cai_optimal_1992}, while we
are not aware of such results for $k$-WL. The $(k+1)$-WL-test has strictly higher
discriminative power than $k$-WL for every $k\geqslant 2$~\cite{cai_optimal_1992}.
Thus the $k$-WL-test for $k\in \nat$ form a family of graph isomorphism heuristics with strictly increasing distinguishing powers for the increasing $k\geqslant 3$ increases. We call this the WL \emph{hierarchy}.

%
Xu et al.~\cite{xu2018powerful} show that MP-GNNs have the same discriminative power as the 1-WL-test if
$f,\phi$ and the graph-pooling function are injective~\cite{xu2018powerful}.
Morris et al.~\cite{morris_weisfeiler_2021, morris2021weisfeiler} show a similar upper bound for DAMP
architectures on isomorphism-type-lifted node-tuple-collections (see
Definition~\ref{def:mp_down_adjacency}), for example $k$-\glspl{gnn}. They found
that the discriminative power of DAMP methods on
$k$-node-tuple-collections is upper bounded by the discriminative power of the
$k$-\gls{wltest}. Moreover, there exist DAMP architectures on
$k$-node-tuple-collections which attain the same power. Some other \gls{gnn}
methods on node-$k$-tuple-collections, such as
$k$-\glspl{ign}~\cite{maron_invariant_2019} have the same discriminative power
as $k$-WL~\cite{geerts_expressive_2020}, while $\delta$-$k$-\glspl{gnn}
have lower expressive power.

In the $\delta$-k-WL-test~\cite{morris2020weisfeiler}, the colour
function additionally stores whether the neighbour is a local lower adjacency, i.e.,
$C_{\textbf{u}}^{ (l)}$ is replaced by
$(C_\textbf{u}^{(l)},\mathbbm{1}_{\textbf{u}\in
\mathcal{N}_{\downarrow,\text{loc}}(\textbf{v})})$ in the equation above. The
authors prove that this variation of the $k$-WL-test is strictly more powerful
at distinguishing non-isomorphic graphs than $k$-WL. To reduce the
computational cost of this algorithm, which scales as $\Omega(n^k)$, the
authors propose the local $k$-WL test, denoted $k$-LWL. In $k$-LWL, only
colours from lower local adjacencies are sent. While this reduces the
computational cost for some graphs, one also loses the expressiveness guarantees
of $\delta$-$k$-WL and $k$-WL. To remedy this, they propose an enhanced
$k$-LWL$^{+}$ test. It is based on $k$-LWL, but in the $i$-th iteration, the
colour message from the local neighbour $\textbf{x}$ which differs in the
$j$-th coordinate has an additional argument

\begin{align*}
    \#_i^j(\textbf{v},\textbf{x})=|\{\textbf{w}: \textbf{w}\in
    \mathcal{N}_{\downarrow}(\textbf{v}): C^{(i)}_{\textbf{w}}=C^{(i)
    }_{\textbf{x}}\}|,
\end{align*}

\noindent
that is, the number of lower adjacencies of $\textbf{v}$ that have the
same colour as $\textbf{x}$. The authors show that in connected graphs,
$k$-LWL$^+$ has the same power as $\delta$-$k$-WL and that the connectedness
condition can be lifted by adding an auxiliary vertex. Although $k$-LWL$^{+}$ has
a better runtime than $\delta$-$k$-WL, its run-time still scales as $\Omega(n^k)$.
The authors further propose sampling techniques to address it. 

The WL hierarchy is a widely used framework for measuring the
expressive power of \glspl{gnn}. However, some \gls{gnn} architectures do not
align with the hierarchy, for example,
\glspl{gsn}~\cite{bouritsas_improving_2021},
\glspl{mpsn}~\cite{bodnar_weisfeiler_2021},
\glspl{cwn}~\cite{bodnar_weisfeiler_2022} or
Autobahn~\cite{thiede_autobahn_2022}. The \gls{wl} hierarchy is built around
plain graphs and NT-Col-graphs. Next, we will
see an alternative approach to measuring the expressiveness of \glspl{gnn},
based on counting isomorphic subgraphs.

\begin{figure*}[t]
    \centering
    \includegraphics[width=1\textwidth]{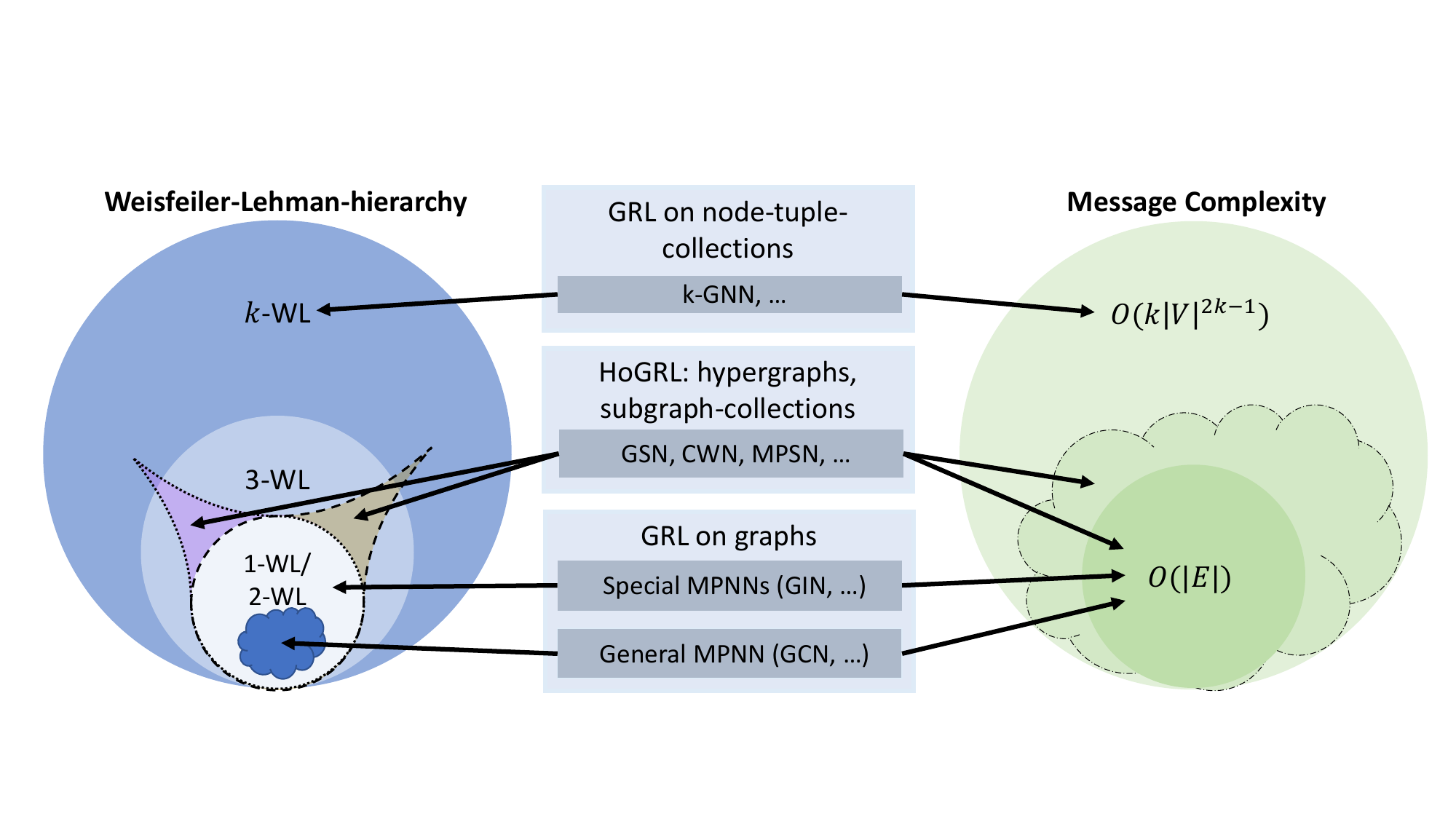}
    \vspace{-6em}
    \caption{Overview of expressive power and message complexity (\#exchanged messages) in HOGRL models. The shapes on the left represent subclasses of graphs, which are distinguishable by a \gls{gi} heuristic or a \gls{grl} method. The circles of increasing radius represent the \gls{wl}-hierarchy, while purple and yellow shapes indicate subclasses of graphs that methods can distinguish, which do not align with the \gls{wl}-hierarchy. The blue cloud is a strict subset of $1$-\gls{wl}-distinguishable graphs, which stands for different subsets of graphs which standard \glspl{mpnn} can distinguish. The right column depicts message complexity. The green cloud indicates a message complexity greater than $\mathcal{O}(|\mathcal{E}|)$, but lower than $\mathcal{O}(k|\mathcal{V}|^{2k-1})$ for large $k$. The central column shows \gls{grl} methods for graphs and \glspl{hogdm}, with edges indicating the expressive power, respectively, and the message complexity of the methods.}
    \label{fig:hogrl_expressiveness}
\end{figure*}

\subsubsection{Substructure Count Based Expressiveness} \label{sec:HoGRL_expressiveness_substructure}

Some approaches
aim at harnessing substructure counts as a measure of GNN expressiveness~\cite{bouritsas_improving_2021, zhang2024beyond, huangboosting}. Specifically, a given GNN is more expressive than another GNN, if it can count -- for a specified subgraph $S$ and for each vertex $v$ or edge $e$ -- what is the count of $S$ that $v$ or $e$ belong to.
%
%
Recent work has improved the understanding of substructure-based expressiveness and
led to new architectures. Chen et al.~\cite{chen2020can} analysed the ability
of established methods to perform subgraph counting for various
classes of subgraphs.
For example, they found that MP-GNNs and
$2$-IGNs can perform subgraph counting of star-shaped patterns, while they fail at
subgraph-counting for connected patterns with $3$ or more nodes.
Higher-order $k$-IGNs can perform induced subgraph-count for patterns
consisting of at most $k$ nodes.
GSNs~\cite{bouritsas_improving_2021} employ
MP on plain graphs with augmented node or edge features encoding the
occurrence of a node, respectively edge in subgraph orbits (see
Definition~\ref{def:orbit}). With particular choices of subgraphs, the authors
demonstrate that GSNs are strictly more powerful than the WL test and no less
powerful than the $2$-FWL test at distinguishing non-isomorphic graphs.

\subsubsection{Function Approximation Based Expressiveness}
\label{sec:func-approx}

While discriminatory frameworks are commonly used, \gls{grl} methods have also been
analysed for their ability to approximate function
classes~\cite{puny2023equivariant}. Maron et al.~\cite{maron_universality_2019} have explored this
question for continuous permutation-invariant functions $\mathcal{F}$ from the class of graphs $\mathcal{G}_n$ with $n$ nodes to $\RE$ and identified neural architectures ($k$-\glspl{ign}) which are universal approximators of $\mathcal{F}$ under certain conditions. Interestingly, the universal approximation of
permutation-invariant continuous functions and the ability to distinguish
non-isomorphic graphs are equivalent~\cite{chen_equivalence_2019}. That is, a
class of permutation-invariant continuous functions $\mathcal{F}$ from $\mathcal{G}_n$ to $\RE$ can distinguish all non-isomorphic
graphs in $\mathcal{G}_n$ if and only if $\mathcal{F}$ is a universal approximator
of permutation invariant functions from $\mathcal{G}_n$ to $\RE$. For a more in-depth discussion of expressiveness in graph representation learning, we
point to the literature~\cite{geerts_expressiveness_2022,geerts_expressive_2020,sato2020survey}.

\subsection{Expressiveness vs.~Computational Complexity}

We also discuss the relation between expressiveness and computational complexity of HOGNNs. We focus on the WL hierarchy related expressiveness since most published results work with it. In our analysis, we only consider bounds on the \emph{number} of messages passed, neglecting the complexity of the transformations involved in MP.

In general, MP architectures on graphs, such as
GCN~\cite{kipf_semi-supervised_2017}, GraphSAGE~\cite{hamilton_inductive_2017}, or
GAT~\cite{velickovic_graph_2018} are efficient, one \acrlong{mp} step
involving $\Theta(|\mathcal{E}|)$ messages. However, their expressiveness is upper-bounded by
the $1$-\gls{wl}-test. A subclass of \acrlong{mp} architectures such as
GIN~\cite{xu2018powerful} or $1$-\glspl{ign}~\cite{maron_invariant_2019} attain the same
expressiveness as $1$-\gls{wl}, while also sending only $\Theta(|\mathcal{E}|)$ messages. Methods with
augmented node and edge features, e.g., \gls{gsn}~\cite{bouritsas_improving_2021}
maintain a low \acrlong{mp} complexity, while being strictly more expressive
than $1$-\gls{wl} and not less powerful than $3$-\gls{wl}. Note that this gain in expressive power requires additional pre-processing. The efficiency of hypergraph-based methods, including methods on simplicial complexes and cell complexes, heavily
depends on the chosen lifting. With certain liftings, it has been shown
that \gls{mpsn}~\cite{bodnar_weisfeiler_2021} and \gls{cwn}~\cite{bodnar_weisfeiler_2022}
attain a strictly higher expressiveness than $1$-\gls{wl} and are provably not less
powerful than $3$-\gls{wl}. Similarly, the \acrlong{mp} complexity of subgraph-based methods depends on the chosen subgraph-collection and certain subgraph-based
models provably beat $1$-\gls{wl} and are no less powerful than $3$-\gls{wl}. A subcollection
of models on $k$-node-tuple-collections, for example,
$k$-\gls{gnn}~\cite{morris_weisfeiler_2021} have been proven to have the same
expressiveness as $k$-\gls{wl} for any $k\geqslant 2$. In $k$-\gls{gnn}s all node-tuples
$\textbf{v}=(v_{i_1},\dots v_{i_k}) \in \mathcal{V}^k$, are considered giving $|\mathcal{V}|^k$ node-tuples in total.
Each node-tuple sends its feature to all its down-adjacencies
{\small
\begin{align*}
    \mathcal{N}_{\downarrow}(\textbf{v}) = \left\{ (v_{i_1},\dots,
    v_{i_{j-1}},w,
    v_{i_{j+1}},\dots,v_{i_{k}}) : w\in \mathcal{V}, j\in [1:k] \right\}.
\end{align*}
}
Since $|\mathcal{N}_{\downarrow}(\textbf{v})|=k\cdot |\mathcal{V}|^{k-1}$, assuming that
messages are sent bidirectionally, we obtain a total of $k\cdot |\mathcal{V}|^{2k-1}$
messages sent in every iteration. Figure~\ref{fig:hogrl_expressiveness} displays the expressiveness and
\acrlong{mp} complexity for a selection of simple and \gls{hogrl} methods.
Note that the \acrlong{mp} complexity scales polynomially in the number of
nodes and for $k$-\gls{gnn} exponentially in the parameter $k$.

\ifuncompressed
\section{Works Related to Higher-Order GNNs}

We also summarize several aspects that are not the focus of this work, but are still related to HOGNNs.

There have been efforts into combining HO with \textbf{temporal} graph data models~\cite{eliassi2021higher}.
This includes graph data models such as $d$-dimensional De Bruijn graphs~\cite{scholtes2014causality, larock2020hypa}, memory networks~\cite{rosvall2014memory}, HO Markov models for temporal networks~\cite{benson2015tensor, peixoto2017modelling, xu2016representing}, motif-based process representations~\cite{schwarze2021motifs}, multi-layer and multiplex networks~\cite{kivela2014multilayer}, hypergraphs~\cite{chen2023hypergraph, yi2020hypergraph, wang2024dynamic, yang2020lbsn2vec++}, and others~\cite{zhu2022high}.

Another line of works is related to \textbf{random walks} on HOGDMs as a way of harnessing HO for learning graph representations.
This includes works dedicated to hypergraphs~\cite{schnake2021higher, zhang2019hyper, carletti2020random, chitra2019random, lu2013high, hein2013total, hayashi2020hypergraph, cooper2013cover, aksoy2020hypernetwork}, hyper-networks~\cite{tu2018structural}, and simplicial complexes~\cite{billings2019simplex2vec, zhan2023simplex2vec, hacker2020k}.

Finally, a line of recent works into topological deep learning~\cite{hajij2021topological} extends HO graph RL into various directions beyond standard GNNs.
Topo-MLP~\cite{ramamurthy2023topo} illustrates representation learning for simplicials without having to harness explicit MP.
Others works study various relationships between GDMs and topological structures, for example investigating topological invariance within vertex embeddings by harnessing tools from persistent homology~\cite{hajij2021persistent}.
Many associated methods have been implemented within TopoX~\cite{hajij2024topox}, a suite of Python packages for machine learning on topological domains.

Finally, there have been efforts into harnessing HO in the context of \textbf{heterogeneous graphs}~\cite{sun2021heterogeneous, fan2021heterogeneous}. In such graphs, nodes and edges may belong to different \textit{classes}~\cite{ma2023single, sun2012mining, zhang2019heterogeneous}.

\fi

\section{Limitations \& Research Opportunities}

We now review future research directions in HOGNNs.

\textbf{Expressivity vs. Efficiency Trade-offs.} There is an inherent trade-off between the expressive power of a HOGNN and its computational complexity. Models that achieve higher discrimination power (e.g. matching $k$-WL for large $k$) often do so at the cost of combinatorial explosion in runtime or memory~\cite{morris_weisfeiler_2021}. For example, a $k$-tuple based GNN that enumerates all $k$-node combinations has $O(n^k)$ nodes in its lifted graph, leading to an explosion in messages passed and parameters as $k$ increases.
\iftr
Even hypergraph or complex-based networks can incur high costs: a single hyperedge connecting $m$ nodes might be represented in a standard GPU structure by $O(m^2)$ pairwise links (for clique expansion) or by introducing extra mediator nodes (for star expansion), either of which increases computation and storage. Similarly, message passing on all faces of a simplicial complex can be much heavier than on edges of a graph when the complex has many high-dimensional simplices. Scalability is thus a concern: HOGNNs may struggle on very large graphs or dense higher-order interactions. 
\fi
\iftr
In practice, designers restrict the order of interactions (e.g., do not enumerate all $k$-node combinations, and use only small hyperedges or bounded-size cell complexes) or employ sampling heuristics, but this can cap the theoretical expressivity. 
\else
In practice, designers restrict the order of interactions (e.g., employ sampling heuristics), but this can cap the theoretical expressivity.
\fi
The runtime and memory overhead of HOGNN layers is a key limitation when moving to industrial-scale data. In summary, the very features that make HOGNNs powerful can render them inefficient for large problems – finding the right balance or developing more scalable higher-order algorithms remains an active research challenge.

\textbf{Ambiguity and Information Loss.} HO graph representations are not unique, and choosing a particular HOGDM can introduce ambiguity or omit information. When “lifting” a plain graph to a higher-order structure, there are often multiple ways to do so. For instance, given a simple graph, one could model it as a hypergraph by treating every triangle as a 3-node hyperedge, but one could also include larger hyperedges or none at all; one could form a simplicial complex by taking all cliques as simplices (clique complex) or only certain ones based on domain knowledge. There is no canonical higher-order encoding for a general graph – different choices (clique expansion vs.~star expansion of hypergraphs, or different subgraph selections in nested models) yield different inductive biases. This can lead to inconsistent results: the performance of a HOGNN may depend heavily on how the higher-order structure was constructed, which is often problem-specific and requires careful tuning.


\textbf{Lack of Benchmarks.} There is a lack of standardized benchmarks for many HOGNN scenarios. While graph ML has a suite of benchmarks (OGB, etc.), hypergraph and simplicial complex learning benchmarks are only now emerging, and they often focus on small-scale or specific domains (e.g. hypergraph classification on a few biological networks).

\iftr
\textbf{Exploring New HO Graph Data Models.}
One avenue of future works lies in developing new HOGDMs beyond those in Table~\ref{tab:gdms}, which could better encapsulate HO interactions between entities in a way that is both computationally efficient and representative of the underlying complexities of the data. For example, one could consider novel substructures for Motif-Graphs, such as different forms of dense subgraphs~\cite{gibson2005discovering, sariyuce2015finding, ma2017fast, lee2010survey, besta2021graphminesuite, gianinazzi2021parallel, besta2022probgraph}.
\fi

\textbf{Exploring New HO Neural Architectures \& Models.}
Similarly to exploring new HOGDMs, one can also study new HOGNN models and model classes. One direction in this avenue would be to explore novel forms of communication channels beyond those described in Section~\ref{chap:from_gdm_to_repL}. 
Here, one specific idea would be to harness the attentional and message-passing flavors for models based on more complex GDMs such as Nested-Graphs, Nested-Hypergraphs, and others.
Another direction would be to harness mechanisms beyond the HOGNN blueprint, for example new activation functions~\cite{sharma2017activation}. Finally, sparsification (e.g., of HO neighbors) based on randomization could prove fruitful in order to alleviate compute costs while offering a tunable accuracy tradeoff.

%
\textbf{Developing Processing Frameworks for HOGNNs.}
There exists a plethora of frameworks for GNNs~\cite{fey2019fast, li2020pytorch, wang2019deep, wan2022pipegcn, zhang2022pasca, waleffe2022marius++, zheng2021distributed, hoang1050efficient, md2021distgnn, cai2021dgcl, wu2021seastar, balin2021mg, thorpe2021dorylus, wang2020gnnadvisor, zhu2019aligraph, wang2021flexgraph, zhang2020agl, jia2020improving, lin2020pagraph, bai2021efficient, zhang20212pgraph, zhang2021gmlp, chen2020fusegnn, tian2020pcgcn, hu2020featgraph, liu2020g3, ma2019neugraph}.
A crucial direction would be to construct such frameworks for HOGNNs, focusing on research into system design and architecture, programmability, productivity, and others.
%
%

\ifuncompressed
\textbf{Integrating HOGNNs into Graph Databases and Graph Frameworks.}
A related research direction is to augment the capabilities of modern graph databases~\cite{angles2018introduction, davoudian2018survey, han2011survey, gajendran2012survey, gdb_survey_paper_Kaliyar, kumar2015domain, gdb_survey_paper_Angles, besta2023gdi, besta2023thegdi, besta2023demystifying} and dynamic graph frameworks~\cite{besta2021practice, sakr2021future, choudhury2017nous} with HOGNNs. While initial designs for such systems combined with GNNs exist~\cite{besta2022neural, ren2023neural}, HOGNNs have still not being considered for such a setting. 
This integration would facilitate the storage, management, and analysis of HO graph data, providing researchers and practitioners with the tools needed to deploy HOGNNs in real-world applications. Research opportunities include the development of standardized APIs, query languages, and optimization techniques for efficient data retrieval and manipulation, ensuring that these advanced models can be effectively utilized in a variety of computational environments.
\else
\textbf{Integrating HOGNNs into Graph Databases and Graph Frameworks.}
A related research direction is to augment the capabilities of modern graph databases and dynamic graph frameworks with HOGNNs. While initial designs for such systems combined with GNNs exist~\cite{besta2022neural}, HOGNNs have still not being considered for such a setting. 
This integration would facilitate the storage, management, and analysis of HO graph data, providing researchers and practitioners with the tools needed to deploy HOGNNs in real-world applications. Research opportunities include the development of standardized APIs, query languages, and optimization techniques for efficient data retrieval and manipulation, ensuring that these advanced models can be effectively utilized in a variety of computational environments.
\fi

\textbf{Parallel and Distributed HO Neural Architectures.}
Parallel and distributed computing offers a pathway to tackle the scalability challenges inherent in processing large-scale, complex graph data. Constructing algorithms that can efficiently distribute the workload of HO neural networks across multiple processors or nodes can significantly reduce training and inference times. Research could explore novel parallelization strategies, communication protocols, and synchronization mechanisms tailored to the unique demands of HO graph processing, aiming to optimize computational efficiency and resource utilization.
\ifuncompressed
One could also investigate effective integration of prompting with distributed-memory infrastructure and paradigms, such as remote direct memory access (RDMA)~\cite{besta2015active, di2022building, di2019network, gerstenberger2013enabling, besta2014fault, besta2015accelerating} or serverless processing~\cite{mao2022ermer, baldini2017serverless, mcgrath2017serverless, jonas2019cloud, copik2021sebs}.
\else
One could also investigate effective integration of prompting with distributed-memory infrastructure and paradigms, such as remote direct memory access (RDMA)~\cite{gerstenberger2013enabling} or serverless processing~\cite{baldini2017serverless}.
\fi
%

\ifuncompressed
\textbf{Exploring Hardware-Acceleration for HO Neural Architectures.}
Understanding energy and performance bottlenecks and mitigating them with specialized techniques such as processing-in-memory~\cite{ahn2015pim, besta2021sisa, ghose2019processing_pim, mutlu2022modern, mutlu2019, seshadri2017ambit}, dedicated NoCs~\cite{besta2018slim, besta2019graph, gianinazzi2022spatial, iff2023hexamesh}, FPGAs~\cite{besta2019graph, de2020transformations, mittal2020survey}, or even quantum devices~\cite{steiger2018projectq, preskill2018quantum} could enable much more scalable models and model execution under stringent conditions.
\else
Understanding energy and performance bottlenecks and mitigating them with specialized techniques such as processing-in-memory~\cite{ahn2015pim}, dedicated NoCs, FPGAs, or even quantum devices could enable much more scalable models and model execution under stringent conditions.

\fi

\ifuncompressed
\textbf{Global Formulations for HOGNNs.}
There have been efforts into building global formulations for GNNs~\cite{besta2023high, besta2023parallel, tripathy2020reducing, bazinska2023cached} and general graph computing~\cite{georganas2014parallel, kepner2015graphs, buluc2010linear, kepner2016mathematical, buluc2008challenges, bulucc2011parallel, bulucc2010solving, yang2022graphblast, bulucc2011combinatorial, solomonik2017scaling, kwasniewski2021pebbles, besta2017push, besta2017slimsell}. This facilitates applying mechanisms such as communication avoidance or vectorization~\cite{solomonik2013cyclops, solomonik2014tradeoffs, kwasniewski2019red, kwasniewski2021parallelmm}. Constructing and implementing such formulations for HOGNNs would be an interesting direction.
\else
\textbf{Global Formulations for HOGNNs.}
There have been efforts into building global formulations for GNNs~\cite{besta2023high} and general graph computing~\cite{buluc2010linear}. This facilitates applying mechanisms such as communication avoidance or vectorization. Constructing and implementing such formulations for HOGNNs would be an interesting direction.
\fi

\textbf{Temporal HOGNNs.}
An important venue of research is to investigate how to combine HO and temporal graphs~\cite{ma2017fast, besta2023hot}. This could involve using HO for more accurate temporal-related predictions.

\ifuncompressed
\textbf{Summarization \& Compression.}
Graph compression and summarization is an important topic that received widespread attention~\cite{liu2018graph, tian2008efficient, riondato2017graph, zhang2010discovery, navlakha2008graph, besta2018log, besta2018survey, besta2019slim}.
GNNs have been used to summarize and compress graphs~\cite{liu2022compact, shabani2024comprehensive, blasi2022graph}. HOGNNs impose hierarchical compute patterns over the graph structure, potentially offering novel capabilities for more effective graph compression and summarization.
\else
\textbf{Summarization \& Compression.}
Graph compression and summarization is an important topic that received widespread attention~\cite{besta2018survey}.
GNNs have been used to summarize and compress graphs~\cite{liu2022compact}. HOGNNs impose hierarchical compute patterns over the graph structure, potentially offering novel capabilities for more effective graph compression and summarization.
\fi

\ifuncompressed

\textbf{Integrating HOGNNs into LLM Pipelines.}
GNNs have also been used together with LLMs and broad generative AI pipelines~\cite{chen2024exploring, fan2024graph, huang2024can, liu2024can, xu2024llm, ren2024survey, qiao2024login}. An exciting research direction is to explore how HOGNNs, and more broadly, how HGDMs could be harness to enhance, for example, the structured reasoning LLM schemes~\cite{besta2023graph, besta2024topologies, yao2024tree, wei2022chain}.

\else

\textbf{Integrating HOGNNs into LLM Pipelines.}
GNNs have also been used together with LLMs and broad generative AI pipelines~\cite{chen2024exploring}. An exciting research direction is to explore how HOGNNs, and more broadly, how HGDMs could be harness to enhance, for example, the structured reasoning LLM schemes~\cite{besta2023graph}.

\fi

\maciej{add some notes / example defs from past sections? For example, this new def of an adjacency in NT-col-graphs?}

\section{Conclusion}

Graph Neural Networks (GNNs) have enabled graph-driven breakthroughts in areas such as drug design, mathematical reasoning, and transportation. Higher Order GNNs (HOGNNs) have emerged as a crucial class of GNN models, offering higher expressive power and accuracy of predictions. However, as a plethora of HOGNN models have been introduced, together with a variety of different graph data models (GDMs), architectures, and mechanisms employed, it makes it very challenging to appropriately analyze and compare HOGNN models.

This paper addresses these challenges by introducing a blueprint and an accompanying taxonomy of HOGNNs, focusing on formal foundations. We model a general HOGNN scheme as a composition of an HOGDM, the specification of message-passing (MP) channels imposed onto that HOGDM (which prescribes how the feature vectors are transformed between GNN layers), and the specifics of lifting and lowering transformations that are used to convert the input plain graph into, and back from, an HO form. 
The taxonomy is then used to survey and analyze existing designs, dissecting them into fundamental aspects. Our blueprint facilitates developing new HOGNN architectures.

We also conduct an analysis of HOGNN methods in terms of their expressiveness and computational costs. Our investigation results in different insights into the tradeoffs between them.
We also provide insights into open challenges and potential research directions, navigating the path for
future research avenues into more effective HOGNNs.

\section*{Acknowledgements}

We thank Hussein Harake, Colin McMurtrie, Mark Klein, Angelo Mangili, and the whole CSCS team granting access to the Ault and Alps machines, and for their excellent technical support. We thank Timo Schneider for immense help with infrastructure at SPCL. This project received funding from the European Research Council (Project PSAP, No.~101002047), and the European High-Performance Computing Joint Undertaking (JU) under grant agreement No.~955513 (MAELSTROM). This project was supported by the ETH Future Computing Laboratory (EFCL), financed by a donation from Huawei Technologies. This project received funding from the European Union's HE research and innovation programme under the grant agreement No. 101070141 (Project GLACIATION). We gratefully acknowledge Polish high-performance computing infrastructure PLGrid (HPC Center: ACK Cyfronet AGH) for providing computer facilities and support within computational grant no. PLG/2024/017103, and the Swiss AI Initiative for the computational grant.

{
\bibliographystyle{abbrv}
\bibliography{references, refs-new, own}
}

\end{document}